%% file: arxiv.tex
\documentclass[10pt, journal, compsoc]{IEEEtran}

\usepackage{amsmath,amsfonts}
\usepackage{amsthm,amssymb}
\usepackage{mathrsfs}
\usepackage{algorithm}
\usepackage{algorithmic}
\usepackage{array}
\usepackage{amsfonts}
\usepackage{algorithm}
\usepackage{tikz}

\usepackage{stfloats}
\usepackage{url}
\usepackage{verbatim}
\usepackage{graphicx}

\usepackage{cite}

\usepackage{bm}
\usepackage{booktabs}
\usepackage{booktabs}
\usepackage{multirow}
\usepackage{caption3}
\usepackage{caption}
\usepackage{xcolor}
\usepackage{float}
\usepackage{diagbox}
\usepackage{color}
\usepackage{pifont}
\usepackage{threeparttable}
\usepackage{float}

\usepackage{hyperref}
\hypersetup{colorlinks=true}

\usepackage{colortbl}

\usepackage{ragged2e}

\newfloat{codefloat}{tbp}{lop}
\floatname{codefloat}{Algorithm}
\usepackage{algorithm}
\usepackage{algorithmic}
\usepackage{xcolor}

\begin{document}

%
\title{\includegraphics[width=1.0cm]{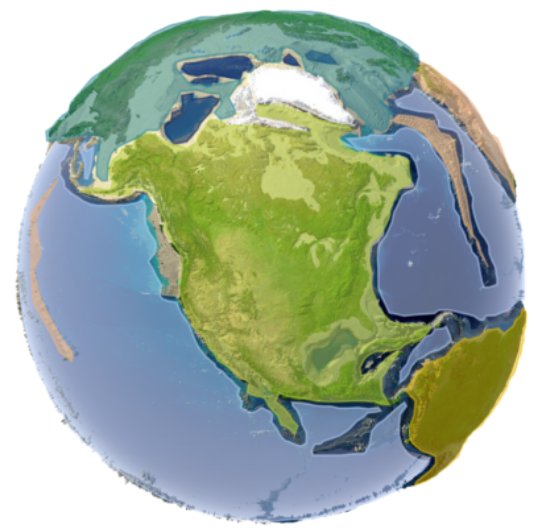} CrossEarth: Geospatial Vision Foundation Model for Domain Generalizable Remote Sensing 
Semantic Segmentation}

%
%
%
%

\author{
Ziyang Gong$^*$,
Zhixiang Wei$^*$,
Di Wang$^*$,
Xiaoxing Hu$^*$,
Xianzheng Ma, 
Hongruixuan Chen, 
Yuru Jia, 
Yupeng Deng, 
Zhenming Ji$^\dagger$,
Xiangwei Zhu$^\dagger$, ~\IEEEmembership{Member,~IEEE,} 
Xue Yang$^\dagger$, ~\IEEEmembership{Member,~IEEE,} \\
Naoto Yokoya, ~\IEEEmembership{Member,~IEEE,}
Jing Zhang,~\IEEEmembership{Senior Member,~IEEE,} \\
Bo Du,~\IEEEmembership{Senior Member,~IEEE,} 
Junchi Yan,~\IEEEmembership{Senior Member,~IEEE,} \\
Liangpei Zhang,~\IEEEmembership{Fellow,~IEEE}
\thanks{Ziyang Gong is with the School of Computer Science, Shanghai Jiao Tong University, China (e-mail: gongziyang@sjtu.edu.cn);\\
Zhenming Ji is with the Key Laboratory of Cryospheric Science and Frozen Soil Engineering, Northwest Institute of Eco-Environment and Resources, Chinese Academy of Sciences, China (e-mail: jizhenming@nieer.ac.cn); \\
Xue Yang is with the School of Automation and Intelligent Sensing, Shanghai Jiao Tong University, China (e-mail: yangxue-2019-sjtu@sjtu.edu.cn);\\
Junchi Yan is with the School of Artificial Intelligence, Shanghai Jiao Tong University, China (e-mail: yanjunchi@sjtu.edu.cn);\\
Yupeng Deng and Xiangwei Zhu are with the School of Electronics and Communication Engineering, Sun Yat-sen University, China (e-mail: dengyp8@mail2.sysu.edu.cn; zhuxw666@mail.sysu.edu.cn); \\
Di Wang, Xianzheng Ma, Jing Zhang, and Bo Du are with the School of Computer Science, Wuhan University,  China; Di Wang, Jing Zhang, and Bo Du are also with the Zhongguancun Academy, China (e-mail: d\_wang@whu.edu.cn; maxianzheng@whu.edu.cn; jingzhang.cv@gmail.com; dubo@whu.edu.cn);  \\
Liangpei Zhang is with the State Key Laboratory of Information Engineering in Surveying, Mapping and Remote Sensing, Wuhan University, China. (e-mail: zlp62@whu.edu.cn); \\
Zhixiang Wei is  with the School of Engineering Science, University of Science and Technology of China, China (e-mail: zhixiangwei@mail.ustc.edu.cn);\\
Xiaoxing Hu is with the School of Information and Electronics, Beijing Institute of Technology, China (e-mail: 3120230695@bit.edu.cn); \\
Hongruixuan Chen and Naoto Yokoya are with the University of Tokyo and also with the RIKEN Center for Advanced Intelligence Project, Japan (e-mail.gschrx@gmail.com; yokoya@k.u-tokyo.ac.ip). \\
Yuru Jia is with the KU Leuven, Belgium, and is also with the KTH Royal Institute of Technology, Sweden (e-mail: yuruj@kth.se).\\
This work started when Ziyang Gong and Zhenming Ji were affiliated with the School of Atmospheric Sciences, Sun Yat-sen University, China.\\
Correspondence e-mail:  jizhenming@nieer.ac.cn
\\ $^*$: Equal contribution; $^\dagger$: Corresponding author. }}

\markboth{Journal of \LaTeX\ Class Files,~Vol.~XX, No.~XX, XXX~XXXX}%
{Shell \MakeLowercase{\textit{et al.}}: Bare Demo of IEEEtran.cls for Computer Society Journals}
%



\IEEEtitleabstractindextext{%
\begin{abstract}
\justifying
Due to the substantial domain gaps in Remote Sensing (RS) images that are characterized by variabilities such as location, wavelength, and sensor type, Remote Sensing Domain Generalization (RSDG) has emerged as a critical and valuable research frontier, focusing on developing models that generalize effectively across diverse scenarios. However, research in this area remains underexplored: (1) Current cross-domain methods primarily focus on Domain Adaptation (DA), which adapts models to predefined domains rather than to unseen ones; (2) Few studies target the RSDG issue, especially for semantic segmentation tasks. Existing related models are developed for specific unknown domains, struggling with issues of underfitting on other unseen scenarios; (3) Existing RS foundation models tend to prioritize in-domain performance over cross-domain generalization. To this end, we introduce the first vision foundation model for RSDG semantic segmentation, \textbf{CrossEarth}. CrossEarth demonstrates strong cross-domain generalization through a specially designed data-level Earth-Style Injection pipeline and a model-level Multi-Task Training pipeline. In addition, for the semantic segmentation task, we have curated an RSDG benchmark comprising 32 semantic segmentation scenarios across various regions, spectral bands, platforms, and climates, providing comprehensive evaluations of the generalizability of future RSDG models. Extensive experiments on this collection demonstrate the superiority of CrossEarth over existing state-of-the-art methods. Our codes and models will be available at \url{https://github.com/VisionXLab/CrossEarth}.

\end{abstract}

}

\maketitle

\IEEEdisplaynontitleabstractindextext

\IEEEpeerreviewmaketitle

\IEEEraisesectionheading{\section{Introduction}\label{sec:introduction}}


\IEEEPARstart{T}{he} advancement of data-driven earth system science is promoting human’s understanding of our homeland \cite{earth_data_sys_sci}. As a significant data source, Remote Sensing (RS) images provide in-depth records for featuring the spatial geometric properties of geospatial objects on ground surfaces and have been applied to extensive fields, including precision agriculture \cite{prec_agri}, urban planning \cite{urban_plan, rui2020survey, abdollahi2021multi}, environment monitoring \cite{environ_survey}, disaster assessment \cite{disa_ass, song2024synrs3d}, etc.

Among these fields, with the advantage of pixel-level comprehension, RS semantic segmentation has been regarded as a fundamental task of accurately and flexibly identifying the category of land uses and covers. In the early times, the RS semantic segmentation community tended to directly utilize existing classical segmentation networks in the computer vision field, such as PSPNet \cite{pspnet}, DeeplabV3+ \cite{deeplabv3_p}, and DANet \cite{danet}. However, since these models are mainly designed for natural images, it can be foreseen that they are hard to cope with RS images. Compared to natural scenes, RS images possess many specialized challenges: \textbf{(1) Significant size variation of foreground objects. (2) Tiny objects and complex backgrounds. (3) Serious foreground-background imbalance. \cite{pointflow}}. For these issues, numerous RS-related segmentation networks are then developed \cite{pointflow, farseg, farseg_pp, factseg, unetformer}, improving the interpretation efficacy.

However, the scenario where both training and testing images come from similar data sources under an independent and identically distributed setting is idealized. In reality, due to the diversity in data acquisition, RS images exhibit considerable variations across multiple dimensions, including wavelength ranges (e.g., RGB vs. IR-R-G), ground sampling distances (high and low resolutions), and platform differences (such as satellites and drones). Additionally, the variability in surface coverage distribution further increases the complexity of RS scenes. Beyond location differences, RS scenes can also display significant visual distinctions in various geographical landscapes, such as urban and rural areas. These discrepancies pose significant challenges for existing intra-domain RS segmentation methods.

\begin{figure*}[!t]
    \centering
    \includegraphics[width=1.0\linewidth]{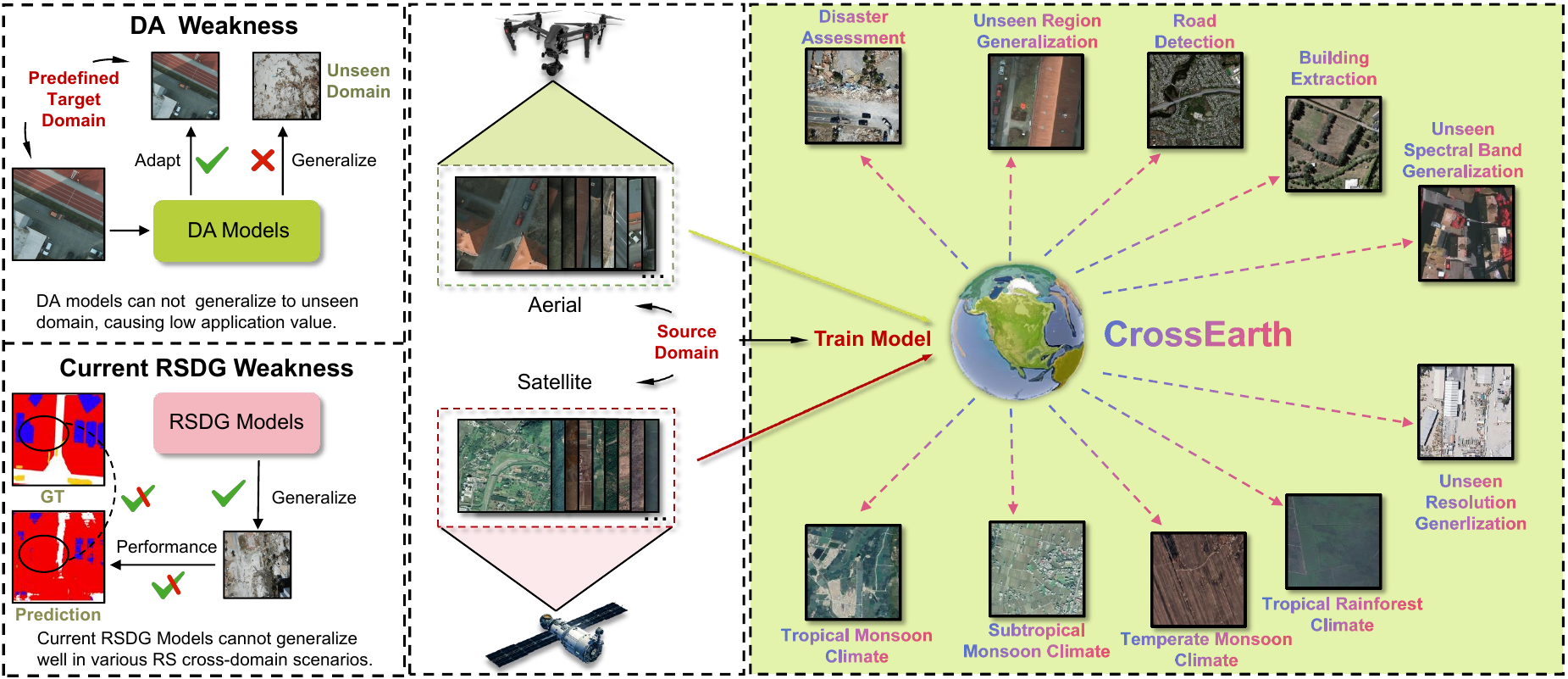}
    \caption{Teaser for \textbf{CrossEarth}. Existing cross-domain RS models mainly focus on DA \cite{   ben2006analysis, ganin2015unsupervised, zou2018unsupervised, li2019bidirectional,ma2022both, gong2024coda, gong2023train, li2024parsing, xiao20233d, bi2024dgss, bi2024generalized, song2024synrs3d}, which only adapt to predefined target domains rather than to diverse unseen domains. On the other hand, existing RSDG methods cannot generalize well across cross-domain scenarios. To address these issues, we propose CrossEarth, the first VFM specifically designed for RSDG, capable of bridging diverse domain gaps and  handling multiple semantic segmentation tasks. } 
    \label{teaser}
\end{figure*}

Under such circumstances, cross-domain semantic segmentation has emerged as a pivotal area of interest within RS in past years. Most excellent works \cite{hong2023cross, cai2022iterdanet, bai2022domain, chen2022mutual, schenkel2019domain} greatly promoted the development of this area by leveraging Domain Adaptation (DA) techniques \cite{   ben2006analysis, ganin2015unsupervised, zou2018unsupervised, li2019bidirectional,ma2022both, gong2024coda, gong2023train, li2024parsing, xiao20233d, bi2024dgss, bi2024generalized, song2024synrs3d}. These DA approaches have notably reduced the dependency on labeled data by leveraging transfer learning to bridge the gap between labeled source domains and unlabeled target domains. However, DA setting forces models to adapt from a source domain to predefined target domains, limiting their ability to generalize well to diverse unseen domains---one of the key challenges in real-world scenarios, as illustrated in the upper left of Figure \ref{teaser}.

In contrast, Domain Generalization (DG) \cite{wang2022generalizing, li2018domain, lambert2020mseg, zhou2022domain} effectively addresses this issue. DG enables models trained on source domain data to generalize to diverse, unseen domains without requiring target domain data. This makes DG a more valuable strategy than DA for RS semantic segmentation, where acquiring target domain data is both time-consuming and labor-intensive. However, as illustrated in the bottom left of Figure \ref{teaser}, there are still very few works in RS semantic segmentation that focus on developing DG models \cite{liang2024single, luo2023diverse, iizuka2023frequency}. These methods often focus on specific scenes, resulting in underfitting when applied to other unseen domain data.

Intuitively, the concept of DG aligns with that of foundation models, which aim to learn universal knowledge through pretraining on massive datasets and can be transferred to a wide range of downstream tasks across different scenarios in a zero-shot manner \cite{fm_definition}. This naturally suggests the potential for combining DG research with foundation models. So far, many advanced Vision Foundation Models (VFMs) \cite{bommasani2021opportunities} have been developed for the RS field \cite{rsp, ringmo, rvsa, bfm, rsprompter, skysense, spectralgpt, saratr-x, hypersigma, wang2024mtp, li2024new, mendieta2023GFM, dong2024upetu, samrs, reed2022self, cong2022satmae}. Compared to traditional scene-specific approaches, these VFMs have demonstrated superior performance across various RS tasks. However, our literature review indicates that existing RS VFMs primarily focus on in-domain segmentation, where training and testing images are drawn from the same data source. When it comes to cross-domain generalization, their understanding capabilities remain limited and largely unexplored.

For this purpose, there is a clear demand for developing a strong foundation model to advance the research of Remote Sensing Domain Generalization (RSDG). To address this, we introduce \textbf{CrossEarth}, the first VFM specifically designed for RS semantic segmentation under the condition of cross-domain generalization. As shown in the right of Figure \ref{teaser}, CrossEarth is equipped with robust domain generalizability, capable of bridging diverse domain gaps, including region, resolution, spectral bands, climate, and even the combination of these factors. Moreover, CrossEarth is highly versatility, it can be applied to extensive segmentation scenarios, ranging from standard RS land cover classification to disaster assessment, road detection, and building extraction, demonstrating its adaptability and effectiveness across various RS applications.

Technically, CrossEarth’s generalizability is achieved through Data Manipulation and Representation Learning \cite{dgsurvery}, comprising two complementary pipelines: Earth-Style Injection and Multi-Task Training. The Earth-Style Injection pipeline augments source domain data by incorporating styles related to Earth-domain data, broadening the training domain distribution to encompass potentially unknown domains. This approach strengthens the model's generalization ability at the data level. The Multi-Task Training pipeline integrates both semantic segmentation and Masked Image Modeling (MIM) by utilizing a shared DINOv2 \cite{oquab2023dinov2} backbone to extract common features. In this way, the model simultaneously performs high-level and low-level tasks, thereby learning robust semantic features essential for cross-domain generalization. Additionally, by integrating detailed and global geospatial semantics extracted from the original image, we further enhance backbone features to deepen the understanding of RS concepts. In summary, these design elements collectively equip CrossEarth with superior cross-domain generalizability, achieved through innovations in both data and model architecture.

\begin{figure}[]
\centering
\includegraphics[width=0.98\linewidth]{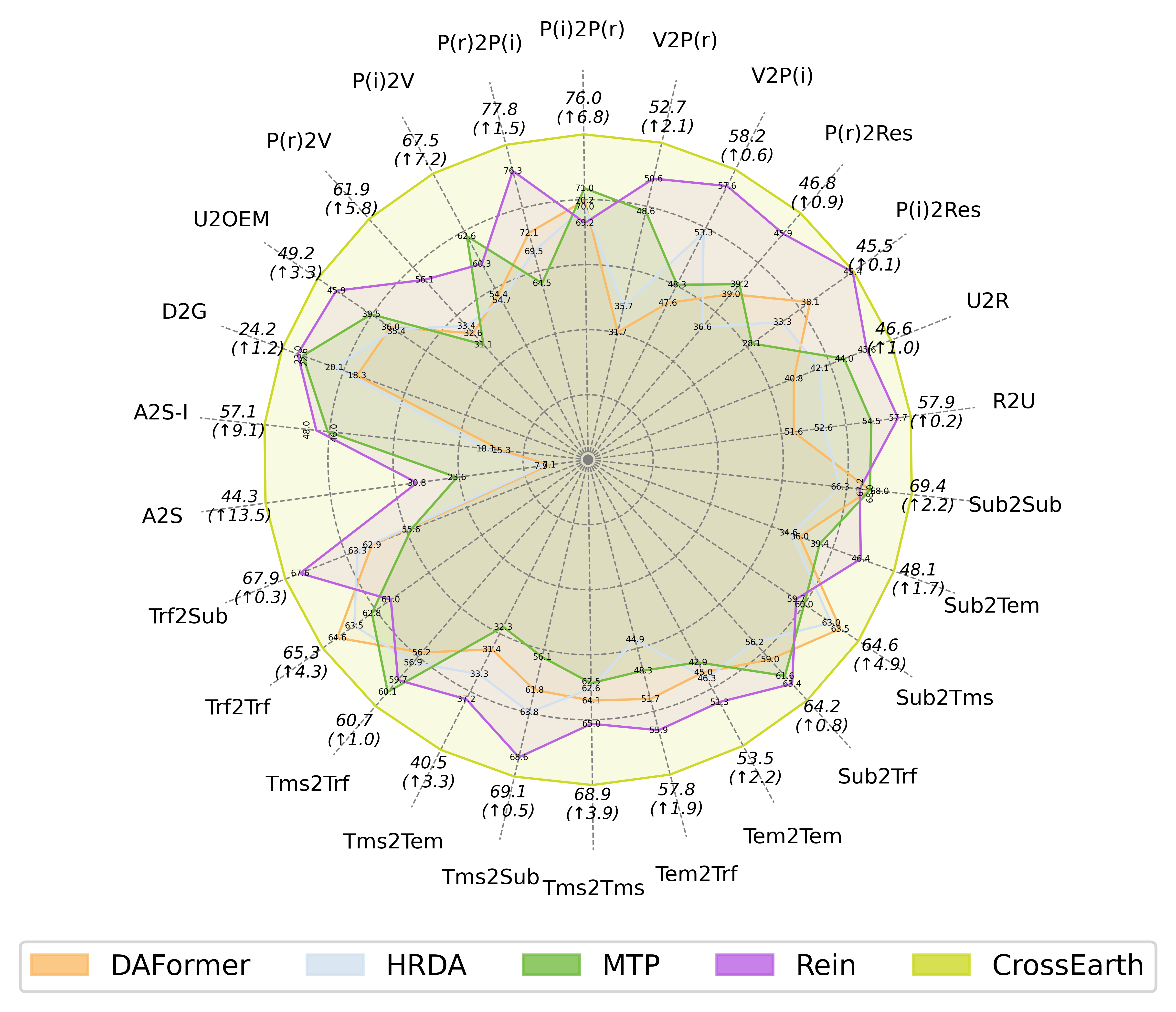}
\caption{We evaluate representative models on 32 evaluation benchmarks, where CrossEarth achieves state-of-the-art performances on 26 settings across various segmentation scenes, demonstrating strong generalizability. All results are reported as mIoU scores.}
\label{radar}
\end{figure}

Additionally, due to the limited availability of benchmarks for testing model generalizability in the RS field, we have collected extensive RS semantic segmentation datasets (details provided later) and extended them to DG settings. Under these settings, we conducted numerous experiments, with results in Figure \ref{radar} demonstrating CrossEarth's outstanding generalizability compared to both specialized DA models and VFMs. The main contributions of this paper are summarized as follows:

(1) We introduce CrossEarth, the first VFM designed for RSDG semantic segmentation. To this end, we propose effective Earth-Style Injection and Multi-Task Training pipelines to improve the cross-domain generalizability from both data and model architecture levels. As a result, CrossEarth is able to cope with diverse domain gaps.

(2) We build a benchmark encompassing 32 semantic segmentation scenarios across 5 domain gap settings to test the cross-domain generalizability of RSDG methods. To our knowledge, it is so far the most comprehensive DG evaluation benchmark. The complete benchmark and preprocessing scripts will be made publicly available.

(3) We conduct extensive experiments on the constructed benchmarks, and the results indicate that CrossEarth achieves state-of-the-art (SOTA) performance, outperforming existing open-source VFMs and DA models. This highlights CrossEarth's superior ability to effectively overcome diverse domain gaps, including variations in regions, spectral bands, platforms, and so on.

\begin{figure*}[t]
    \centering
    \includegraphics[width=1\textwidth]{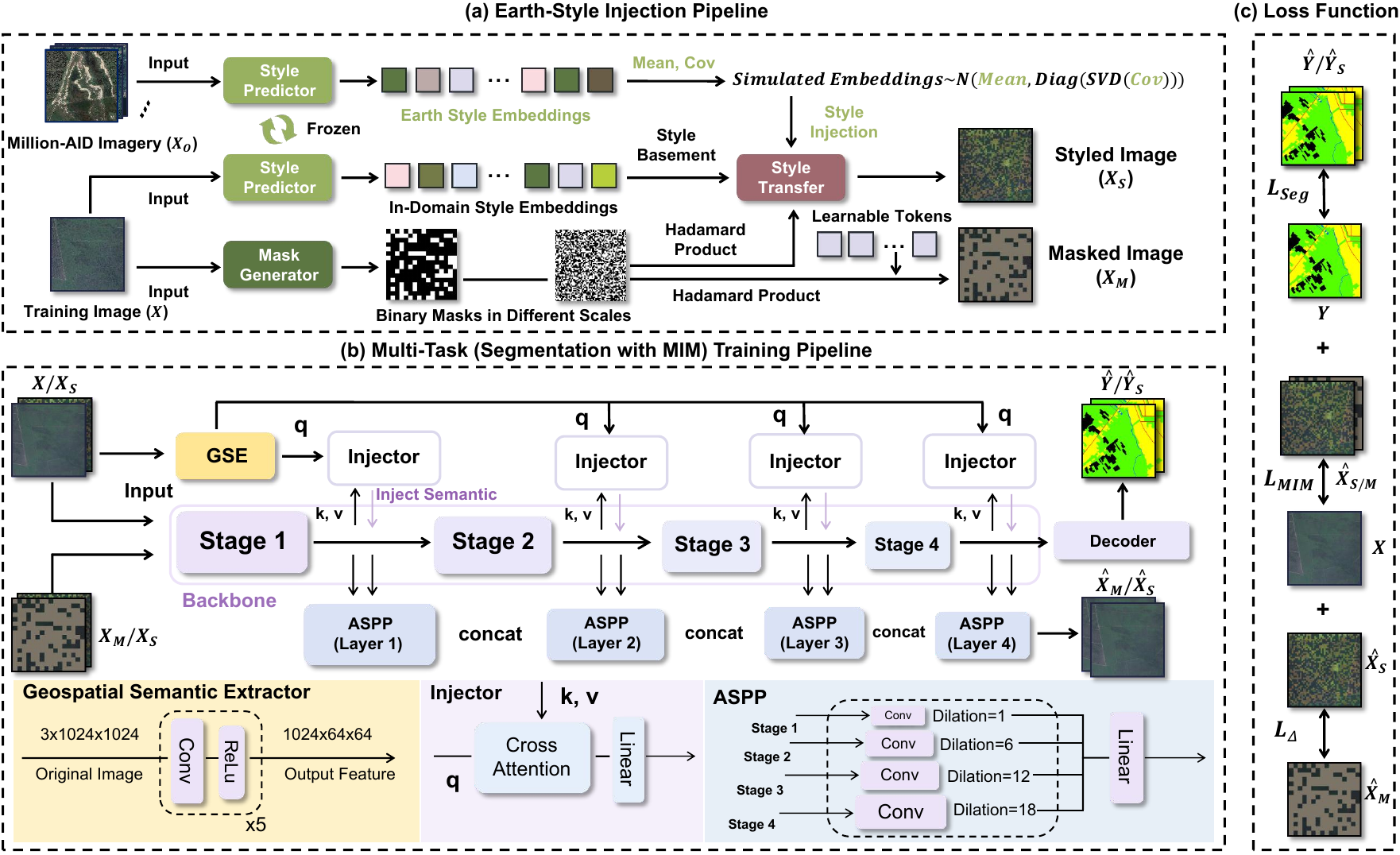}
    \caption{
    Framework of CrossEarth. The model integrates an Earth-Style Injection pipeline with a Multi-Task Training pipeline. At each iteration, a sample gate selects either the original image $X$ or its styled variant $X_S$ for the segmentation flow (backbone $f_\phi$, GSE $f_G$, Injectors $f_I$, and decoder $f_\theta$). Meanwhile, the MIM flow reconstructs masked inputs $X_M$ (together with $X_S$ if selected) via the ASPP decoder $f_A$. The backbone $f_\phi$ is shared but frozen, while GSE and Injectors are applied only to the segmentation branch. Training jointly optimizes $L_{Seg}$, $L_{MIM}$, and $L_\Delta$.
    }
    \label{pipeline}
\end{figure*}


\section{Related Work}\label{sec:related works}

Driven by the explosive impact of Large Language Models (LLMs) \cite{brown2020language, chowdhery2023palm, touvron2023llama}, a major paradigm shift has occurred in the computer vision field, moving from domain-specific models to generalist Foundation Models (FMs) \cite{wang2023images, bar2022visual, wang2023seggpt, bai2024sequential, wang2023internimage, he2022masked, liu2024visual, wang2024visionllm, chen2024internvl, achiam2023gpt}. Among the VFMs \cite{rsp, ringmo, rvsa, bfm, rsprompter, skysense, spectralgpt, saratr-x, hypersigma, wang2024mtp, li2024new, mendieta2023GFM, dong2024upetu, samrs, cong2022satmae, scheibenreif2024parameter,scheibenreif2024parameter} have also seen substantial innovation, where Cha et al. \cite{bfm} introducing the first billion-scale focused on RS detection and segmentation. HyperSIGMA \cite{hypersigma} represents the first billion-scale VFM for multispectral RS imagery, supporting multiple tasks like classification, detection, unmixing, denoising, and super-resolution. Skysense \cite{skysense} captures spatiotemporal information and performs well across diverse scenarios. While MTP \cite{wang2024mtp} employs multi-task pretraining, is currently recognized as the SOTA open-source VFM for RS semantic segmentation. Beyond these advances, recent work\cite{pirvu2023multi} addresses learning and prediction in the Earth System from a multi-task and multi-layer perspective, where tasks act as mutual teachers within a hypergraph structure. By exploiting consensus among pathways, the model achieves significantly coherent interpretations of Earth processes and improves generalization under scarce labels. In addition, many studies adapt the Segment Anything Model (SAM) to RS, such as text-based PerSAM \cite{osco2023segment} exploring its zero- and one-shot segmentation capabilities and showing the promise of promptable segmentation. Nevertheless, due to the limited exploration of DG in the RS field, the performance of existing RS foundation models in cross-domain tasks is still underexplored. Recognizing that the objective of cross-domain generalization aligns with the zero-shot capabilities of foundation models, we propose CrossEarth, the first VFM for RSDG semantic segmentation, to inspire future research in this critical direction.

\section{Method}\label{sec:method pipeline}

In this section, we first introduce the overall framework of CrossEarth in \ref{overview}. The details of the Earth-Style Injection pipeline and the Multi-Task Training pipeline will be presented in \ref{earth-style-injection}, and \ref{mim-training}, respectively.

\subsection{Overview of CrossEarth}
\label{overview}

CrossEarth consists of two pipelines: an Earth-Style Injection pipeline and a Multi-Task Training pipeline. The former includes an Augmentor module, which is composed of a Style Predictor $p_s$ and a Style Transfer module $p_t$, together with a Mask Generator $p_m$. For the Multi-Task Training pipeline, there are two task flows: semantic segmentation and Masked Image Modeling (MIM). In this pipeline, we adopt a common backbone DINOv2 $f_\phi$ to extract features. Except for the $f_\phi$, semantic segmentation flow consists of a Mask2Former Decoder $f_\theta$, a Geospatial Semantic Extractor (GSE) $f_G$, and an Injector $f_I$. MIM flow consists of an ASPP \cite{chen2018encoder} decoder $f_{A}$.

At each training iteration, the in-domain images $X\in \mathbb{R}^{H\times W\times 3}$ are first processed by the Earth-Style Injection pipeline. Specifically, the Augmentor produces a styled variant $X_S$, while the Mask Generator $p_m$ outputs a masked version $X_M$. A sample gate $u \sim \text{Bernoulli}(p)$ then determines whether the segmentation flow receives the original image $X$ ($u=0$) or the styled image $X_S$ ($u=1$).  The selected input is passed through the backbone $f_\phi$, refined by the GSE and Injectors, and finally decoded by $f_\theta$ to produce the segmentation prediction $\hat{Y}$ (for $X$) or $\hat{Y}_S$ (for $X_S$).
In parallel, the MIM branch takes either $X_M$ alone ($u=0$) or both $X_M$ and $X_S$ ($u=1$) as input. For $X_M$ and $X_S$, their backbone features are processed by the ASPP decoder $f_A$, and the outputs are concatenated and projected to $\hat{X}_M$ or $\hat{X}_S$, thereby reconstructing the original and styled images, respectively. This process has been shown in Algorithm \ref{algorithm1}.

Notably, the Earth-Style Injection pipeline only focuses on data manipulation without parameter updating. During training, only the parameters of GSE, Injector, Mask2Former decoder, and ASPP decoder are updated. In the following text, we will illustrate the technique details and design motivations of each component in CrossEarth.

\subsection{Earth-Style Injection Pipeline}
\label{earth-style-injection}
As we mentioned in the Introduction, this pipeline aims to enlarge the coverage of the training domain distribution. Thus, in this section, we first discuss the distribution issues existing in cross-domain generalization. Following \cite{koh2021wilds, yaoimproving}, we define that the training domain distributions can be denoted as $D_{train}$ and the test domain distributions are $D_{test}$. Typically, domain gaps occur when training model on $D_{train}$ but testing on $D_{test}$, due to the discrepancies between distributions, i.e., $D_{train} \cap D_{test} \approx \varnothing$. In such cases, expanding the coverage of $D_{train}$ \cite{zhao2024shade} is considered a reasonable way to enhance the model's generalizability.

To this end, we introduce a data augmentation paradigm to inject style embeddings of field-related data $X_{O}$ into in-domain training data. We suppose that the distribution $D_{field}$ of employed $X_{O}$ partially overlaps with the distribution $D_{test}$ of unknown scenes. Thus, augmented training distributions are designed as a union: $D^\prime_{train}=D_{train} \cup D_{field}$, ensuring that $D^\prime_{train} \cap D_{test} \neq \varnothing$. In our paper, $X_O$ refers to the Million-AID dataset \cite{millionaid}, which comprises over one million globally distributed scene instances and has become a standard pretraining resource for building remote sensing foundation models \cite{rvsa,wang2024mtp,bfm}. This pipeline effectively reduces domain distribution gaps at the data level, enhancing the model's ability to generalize across diverse unseen domains.

We next illustrate technical details. As shown in Figure \ref{pipeline}, before generating a styled image $X_S$, three important components are needed: Earth-Style Embeddings $\varepsilon_o$ extracted by $p_s$ for style injection, In-domain Style Embeddings $\varepsilon$ extracted by $p_s$ as the style basis, and Binary Mask $M$ generated by $p_m$ to combine the former two embeddings. 
Concretely, $p_s$ first accepts $X$ and $X_O$ as input images to generate Earth-Style Embeddings $\varepsilon_o$ and In-Domain style Embeddings $\varepsilon$, respectively.
\begin{equation}
\begin{split}
    \varepsilon_o  &= p_s(X_O)\\
    \varepsilon & =  p_s(X).
\end{split}
    \label{eq:epsilon}
\end{equation} 
Here, we extract the mean value and covariance matrix of $\varepsilon_o$ in an off-line manner to initialize a simulated embedding $\varepsilon^\prime_o$. The reason for this step is to avoid extracting $\varepsilon_o$ in every iteration to improve training efficiency.
\begin{equation}
    {\varepsilon_o}^\prime \sim N(Mean,Diag\thinspace(SVD\thinspace(Covariance))),
\end{equation}
where $Diag$ means the diagonal elements of a matrix and $SVD$ means the singular value decomposition to the covariance matrix. Then, Mask Generator $p_m$ accepts $X$, mask ratio $\tau_m$, and mask patch size $B$. Here, the patch size $B$ determines how many patches in $X$ will be divided into, and then $\tau_m$ decides which patches need to be masked. The related analyses and selections of these two hyperparameters have been presented in the supplementary material. The equations below show the process of how $p_m$ works:
\begin{equation}
    M \sim U(0,\thinspace1)^{\lceil \frac{H}{B} \rceil\thinspace\times\thinspace \lceil \frac{W}{B}\rceil}.
\label{eq: M1}
\end{equation}
\begin{equation}
    M_i = \left\{\begin{array}{cl} 
         1, \thinspace  & if \thinspace \thinspace M_i > \tau_m \quad i=1,\cdots,\lceil \frac{H}{B} \rceil\thinspace\times\thinspace \lceil \frac{W}{B}\rceil \\
         0, &otherwise
    \end{array}\right..
\label{eq: M2}
\end{equation}
For generating the mask $M$, we refer to the MIC \cite{hoyer2023mic}. Specifically, we evaluate whether the pixel value in $M$ exceeds $\tau_m$. If yes, the pixel will be set as 1, otherwise, it will be set to 0. Then we resize $M$ from ($\lceil \frac{H}{B} \rceil\thinspace, \lceil \frac{W}{B}\rceil$) to  ($H, W$) for later image generation. 

Finally, we leverage $p_t$ to inject simulated Earth styles into $X$ and apply a Hadamard product with $M$ to obtain $X_S$: 
\begin{equation}
    X_S = p_t\thinspace({\varepsilon_o}^\prime,\thinspace\varepsilon,\thinspace X) \odot M + X \odot (1-M).
\label{eq: XS}
\end{equation}

Besides generating $X_S$, this pipeline also generates an extra masked image $X_M$ for subsequent MIM tasks. This process introduces learnable visual prompts $v \in \mathbb{R}^{H\times W\times 3}$, which are initialized with zero and will be integrated with $M$. The generation of $X_M$ can be formulated as follows: 
\begin{equation}
    X_M = X \odot M \odot v.
\label{eq: XM}
\end{equation}

Notably, the structures of $p_s$ and $p_t$ in this pipeline are lightweight networks from \cite{jackson2019style}, which are kept frozen in our framework to keep the style augmentation module efficient and scalable for large-scale training.

\input{table2}

\subsection{Multi-Task Training Pipeline}
\label{mim-training}
Multi-task learning has been validated to be beneficial in improving models' representation ability \cite{wang2024mtp}. Inspired by this, we simultaneously consider two tasks: semantic segmentation and MIM, as shown in Figure \ref{pipeline} (b) and (c). 

\textbf{Semantic Segmentation Flow} This is the main training flow of CrossEarth, comprising a DINOv2 backbone network  \cite{oquab2023dinov2} $f_\phi$, a GSE $f_G$, an Injector $f_I$, and a Mask2Former Decoder \cite{cheng2022masked} $f_\theta$. Notably, the original structure of our baseline Rein \cite{Wei_2024_CVPR} includes only $f_\phi$ and $f_\theta$. However, the backbone $f_\phi$ and decoder $f_\theta$ in Rein was not designed for RS imagery, as they lack the capacity to learn geospatial semantics. To this end, we introduce a GSE to obtain geospatial queries and an Injector for capturing related knowledge from the features extracted by the backbone network.

\input{algorithm1}

In designing the GSE, we assert that the generated geospatial queries should be both original and global. Therefore, we aim to keep the GSE structure as simple as possible. Technically, GSE is composed of five convolutional layers \cite{lecun1989backpropagation} with ReLU activation \cite{glorot2011deep}, as shown in Figure \ref{pipeline}. Similar to the backbone, GSE accepts $X$ and $X_S$ as the inputs. Then these queries will be used by an Injector to extract geospatial features, and this operation is achieved by a simple cross-attention \cite{huang2019ccnet}, where the backbone features serve as key and value $k,v$, and geospatial features act as the query $q$. 
Specifically, the queries $q$ adaptively select and weight the most relevant $k,v$ pairs during the attention process, producing enhanced features that are fed back into the backbone. By applying this operation across multiple backbone layers, geospatial semantics are not only preserved but also progressively reinforced throughout the representation learning process.

The whole Semantic Segmentation flow can be given:

\begin{equation}
\hat{Y}_{seg} =
\begin{cases}
\hat{Y} \leftarrow f_\theta\!\left(f_I\!\left(f_G(X),\ f_\phi(X)\right)\right), & u=0 \\
\hat{Y}_S \leftarrow f_\theta\!\left(f_I\!\left(f_G(X_S),\ f_\phi(X_S)\right)\right), & u=1
\end{cases}
\label{seg_predict}
\end{equation}
Here, to ensure the generalization ability of the model, we adopt both original images and styled images during the training. Following our baseline model Rein \cite{Wei_2024_CVPR}, the segmentation loss is defined as $L_{seg}(Y, \hat{Y})$ or $L_{Seg}(Y,\hat{Y}_S)$, where $Y$ is the labeled segmentation map, and $L_{seg}$ involves both cross entropy loss and dice loss \cite{milletari2016v}.

\textbf{Masked Image Modeling Flow} This flow aims to collaborate with the segmentation pipeline for robust geospatial feature learning. 
In our view, incorporating the MIM flow offers two key advantages: (1) Accomplish segmentation and MIM tasks simultaneously ensure that the features extracted by the backbone are generalizable and domain-invariant; (2) As a low-level vision task, MIM encourages models to capture fine-grained global information, aligning with the design objectives of the GSE.

Therefore, we still leverage the backbone network $f_\phi$ to extract features from $X_S$ and $X_M$. Then, we introduce an ASPP decoder $f_A$ for image restoration. Unlike traditional ASPP techniques that only process the final feature, our MIM flow feeds all features from different stages of the backbone into the ASPP decoder $f_A$ \cite{chen2018encoder}, as shown in Figure \ref{pipeline}. In this process, each stage feature is refined through convolution layers with varying dilation rates, allowing $f_A$ to capture more diverse information than the standard ASPP approach. Notably, we deactivate the GSE and injectors in this process to prevent additional assistance to the MIM flow, thereby encouraging the backbone to realize its full potential. Finally, feature processed by $f_A$ will pass a vanilla linear layer to generate image predictions. We also show this process in Figure \ref{pipeline}, which can be formulated as:
\begin{equation}
\hat{X} =
\begin{cases}
\hat{X}_{M} = \text{Linear}(f_A(f_\phi(X_M))), & u=0 \\
\hat{X}_{M}, \hat{X}_{S} = \text{Linear}(f_A(f_\phi(X_M, X_S))), & u=1
\end{cases}
\label{eq: mim}
\end{equation}

In the MIM flow, the reconstruction loss $L_{\text{MIM}}$ is formulated as the discrepancy between the predicted outputs and their corresponding ground-truth images, measured by either the $\ell_1$ or $\ell_2$ norm. Specifically, when $u=0$, the loss is computed solely between $\hat{X}_M$ and $X_M$. When $u=1$, it is defined as the sum of two terms, namely the distances between $\hat{X}_M$ and $X_M$, and between $\hat{X}_S$ and $X_S$. The choice of loss function depends on the dataset, and related discussions are shown in Sec.~\ref{final ablation}. Meanwhile, to better constrain image restoration, we introduce a metric loss $L_\Delta$ to the MIM flow, where the styled and masked image predictions $\hat{X}_S$ and $\hat{X}_M$ calculate an L1 loss. The metric loss can be expressed as $L_\Delta(\hat{X}_S, \hat{X}_M)$ and the overall training loss of the Multi-Task Training pipeline is as follows:

\begin{equation}
L =
\begin{cases}
L_{Seg}+ L_{MIM}, & u=0, \\[6pt]
L_{Seg} + L_{MIM} + L_\Delta, & u=1.
\end{cases}
\label{loss}
\end{equation}

Notably, $L_\Delta$ is only activated when $u=1$. In contrast, $X_M$ is not randomly sampled, and it is always used for image restoration.

\section{RSDG Semantic Segmentation Benchmark}

\begin{figure}
    \centering
        \includegraphics[width=1\linewidth]{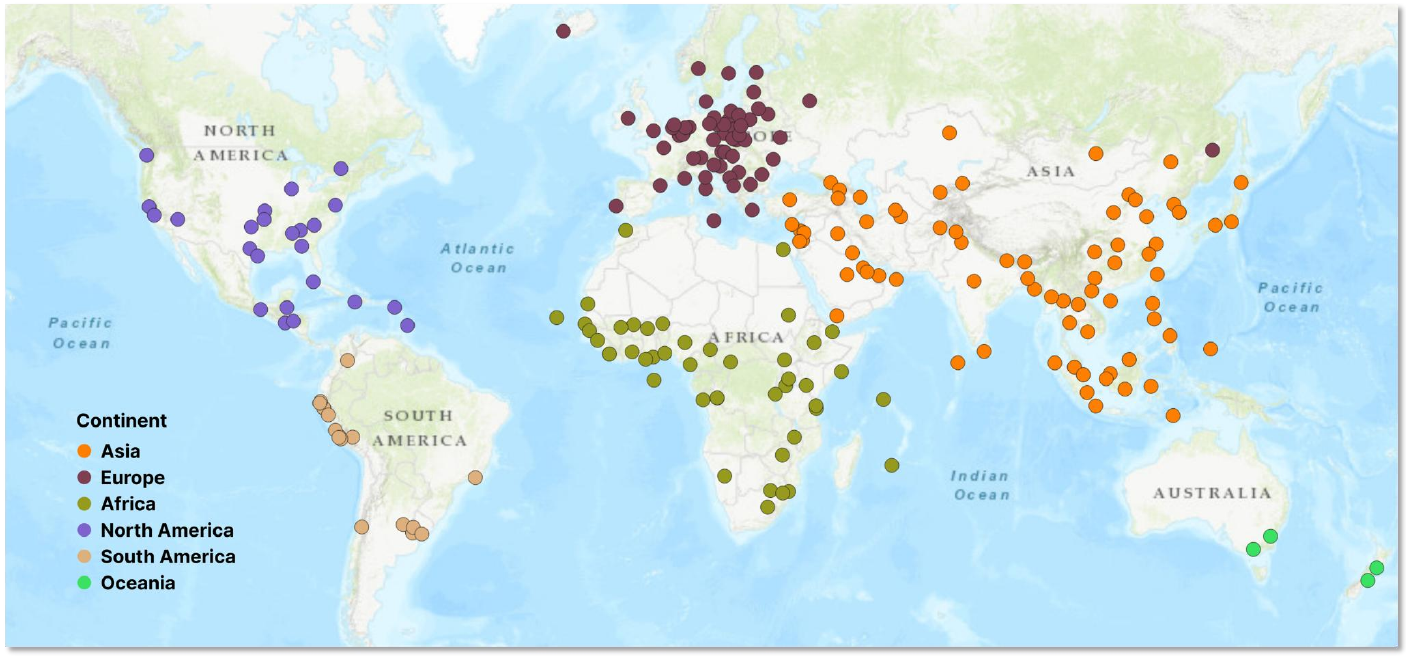}
    \caption{Geographic distribution of the CrossEarth benchmark, demonstrating its comprehensive coverage across hundreds of cities on six continents.}
    \label{fig:global map}
\end{figure}

As noted above, the current RS community lacks unified benchmarks for evaluating model generalizability. To support the advancement of the RSDG field, we have compiled widely-used RS semantic segmentation datasets and extended them to DG settings, as shown in Table \ref{benchmark collection}. Our benchmark comprises 32 semantic segmentation task settings including three specific application scenarios: disaster assessment, building extraction, and road detection, across five compositional domain gaps: (1) Unseen Region; (2) Unseen Spectral Band; (3) Unseen Region and Spectral Band; (4) Unseen Region and Platform; (5) Unseen Region and Climate. Specifically, we have collected and organized a diverse set of widely-used RS semantic segmentation datasets, including ISPRS Potsdam and Vaihingen \cite{zhang2023pseudo}, RescueNet \cite{rahnemoonfar2022rescuenet}, LoveDA \cite{wang2021loveda}, WHU Building \cite{liang2023multilevel, liang2023unsupervised}, DeepGlobe \cite{demir2018deepglobe}, Massachusetts \cite{MnihThesis}, CASID \cite{liu2023large}, OpenEarthMap \cite{openearthmap}, and GlobalRoadNet \cite{globalroadnet}.  The geographical location distribution of this benchmark is visualized in Figure \ref{fig:global map}. For more details on the datasets used and the construction of benchmarks, please refer to the supplementary material.

\input{table3}
\begin{figure*}[t]
    \centering
    \includegraphics[width=1\textwidth]{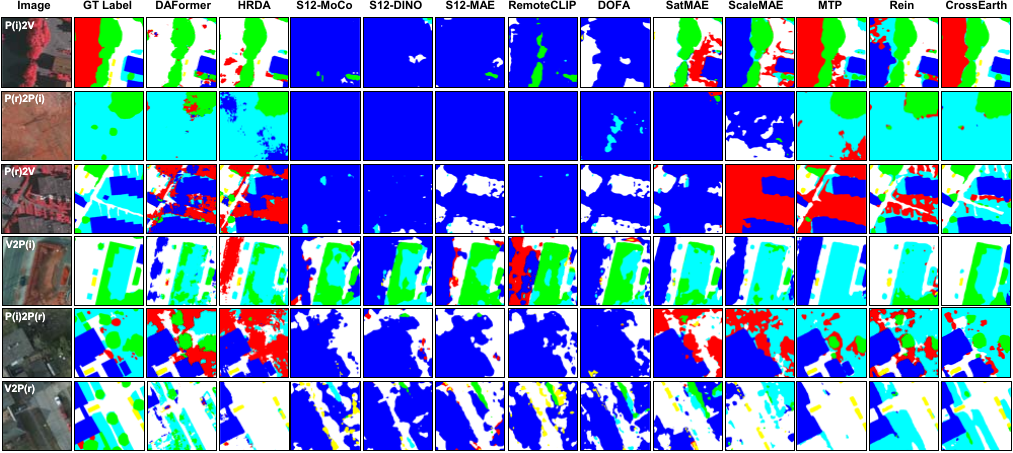}
    \caption{Visualizations of predicted segmentation maps on Potsdam and Vaihingen benchmarks. For the color map, white is the Impervious surface class, \textcolor[RGB]{255,0,0}{red} is the clutter class, \textcolor[RGB]{0, 0, 255}{blue} is the building class, \textcolor[RGB]{0, 255, 255} {cyan} is the low vegetation class, \textcolor[RGB]{0, 255, 0}{green} is the tree class, and \textcolor[RGB]{255, 255, 0}{yellow} is the car class. }
    \label{p2v_seg}
\end{figure*}


\section{Experiments}
\subsection{Preliminary}
To thoroughly investigate various domain gaps and demonstrate the generalizability of CrossEarth, we conduct extensive experiments by comparing it with a series of representative models on the constructed RSDG benchmark.

Specifically, we choose DAFormer \cite{hoyer2022daformer}, HRDA \cite{hoyer2022hrda}, SSL4EO-MoCo (S12-MoCo)~\cite{SSL4EO}, SSL4EO-DINO (S12-DINO)~\cite{SSL4EO}, SSL4EO-MAE (S12-MAE)~\cite{SSL4EO}, RemoteCLIP~\cite{liu2024remoteclip}, DOFA~\cite{xiong2024neural}, SateMAE~\cite{cong2022satmae}, ScaleMAE~\cite{reed2023scale}, MTP \cite{wang2024mtp}, and Rein \cite{Wei_2024_CVPR} for comparison. Here, we adopt the extended DG version of DAFormer and HRDA \cite{hoyer2023domain}. Notably, CrossEarth is the first VFM for RSDG.

For all experiments, we set the iteration number to 30K with a batch size of 1, optimizing the models using the AdamW \cite{adamw} with a learning rate of 1e-4. Additionally, when training the comparison models, we adhere to their original settings, including their optimizers, learning rates, and other specialized parameters. For the MIM loss, we use the L1 loss for CASID experiments, while other datasets employ Mean Squared Error (MSE) loss. All experiments are conducted with Pytorch on NVIDIA V100 GPUs.

\subsection{Generalize to Unseen Region}

In this section, the \textit{region} factor is the primary discrepancy between source and unseen domains. For example, in the P(i)2V experiments, the source domain Potsdam images and the unseen Vaihingen images share the same IR-R-G bands, rendering the regional difference as the main gap. Consequently, models trained on the source domain Potsdam are expected to perform well on the unseen region Vaihingen. The datasets used in this setting include Potsdam, Vaihingen, LoveDA, DeepGlobe, Massachusetts, etc.

\textbf{Potsdam and Vaihingen} Using the ISPRS Potsdam and Vaihingen datasets, we conduct experiments on two DG benchmarks: P(i)2V and V2P(i), and the results have been shown in Table \ref{p2v}. In the P(i)2V setting, RS VFMs exhibit substantial variability. While MTP achieves the strongest result with 62.6\% mIoU, others such as ScaleMAE (35.5\%), SatMAE (30.6\%), DOFA (25.3\%), RemoteCLIP (14.3\%), and the S12 variants (18–28\%) perform considerably worse, especially on small-object categories like \textit{Car}, where scores often approach zero. Rein achieves 60.3\%, slightly below MTP, likely due to its natural-image origin. In contrast, CrossEarth delivers the best overall performance with 67.5\%, surpassing Rein by +7.2\% and MTP by +4.9\%, and showing consistent advantages on broad categories (\textit{Impervious surfaces}, \textit{Low vegetation}) as well as challenging ones such as \textit{Clutter}.

On the V2P(i) benchmark, similar trends are observed. ScaleMAE (35.2\% mIoU), SatMAE (31.3\%), and DOFA (27.9\%) remain limited, while MTP achieves the best among RS VFMs with 48.3\%. Rein is a strong baseline at 57.6\%, particularly excelling on the \textit{Building} class. CrossEarth establishes a new SOTA with 58.2\% mIoU, exceeding Rein by +0.6\% and maintaining stable gains across both large-area and small-object categories, demonstrating greater robustness and generalization.


\textbf{LoveDA (Rural and Urban)} In addition to different cities, we further consider another regional cross-domain gap characterized by urban and rural landscapes. For this purpose, we use the LoveDA dataset and conduct experiments on two benchmarks: Rural to Urban (R2U) and Urban to Rural (U2R), with results shown in Table \ref{loveda}. 

In the R2U setting, specialized DA models such as DAFormer and HRDA achieve moderate performance (around 52–53\% mIoU). Among RS VFMs, most approaches including S12-MoCo/DINO/MAE and RemoteCLIP struggle to adapt, often failing on classes such as \textit{Road} and \textit{Barren}. SatMAE and ScaleMAE perform relatively better, particularly on dominant categories (e.g., ScaleMAE reaches 71.6\% on \textit{Water}), but their overall results remain limited. Within the RS VFM family, MTP achieves the strongest performance, excelling on the \textit{Road} class (58.3\%), while Rein, though initialized from natural-image pretraining, achieves competitive overall results with notable strengths in \textit{Building} (64.0\%) and \textit{Agriculture} (61.1\%). On the U2R benchmark, a similar trend emerges. Most RS VFMs again show limited effectiveness, particularly on agriculture-dominated categories. Although ScaleMAE and SatMAE remain more competitive, they still fall short of MTP and Rein. CrossEarth achieves SOTA performances on both benchmarks, with mIoU improvements of 0.2\% and 1.0\%, respectively, compared to the baseline Rein. Notably, in the U2R task, the largest gains appear in \textit{Building} (+5.2\%) and \textit{Agriculture} (+4.7\%), despite the significant visual differences these classes present across urban and rural landscapes (see pictures in the supplementary material), indicating CrossEarth's effectiveness in bridging such gaps.

\input{table4}

\textbf{Road Detection (DeepGlobe and Massachusetts)} RS road detection is fundamental for advancing transportation infrastructure and maintaining up-to-date map data. However, the environmental complexity of roads across different geographical locations makes it challenging to apply a single model effectively across diverse regions, leading to high data collection and model training costs. To evaluate cross-regional road detection performance, we employ DeepGlobe and Massachusetts datasets, where we train on the DeepGlobe dataset and test on the Massachusetts test set (D2M), and the Results have been shown in Table \ref{road_building_table}. 

In the D2M benchmark, DA models such as DAFormer and HRDA achieve moderate results (41.9–50.3\% mIoU), with HRDA performing better. RS-specific VFMs including S12-MoCo, S12-DINO, S12-MAE, DOFA, and RemoteCLIP perform very poorly ($<$10\%), indicating that pretraining alone does not transfer effectively to road extraction. SatMAE (23.3\%) and ScaleMAE (32.1\%) alleviate this issue to some extent, but still fall short of stronger baselines. Among RS VFMs, MTP attains the best performance (54.3\%), demonstrating strong adaptation to the thin and elongated structures of roads. By comparison, the general-purpose VFM Rein also performs competitively (49.7\%), surpassing most RS-specific VFMs. Our CrossEarth reaches 50.5\%, slightly exceeding Rein (+0.8\%) and HRDA, though still below MTP. We argue that this gap may be related to the structure of the backbone network, and provide further discussion in the supplementary material.


\input{table11}

\input{table5}
\begin{figure*}[t]
    \centering
    \includegraphics[width=1\textwidth]{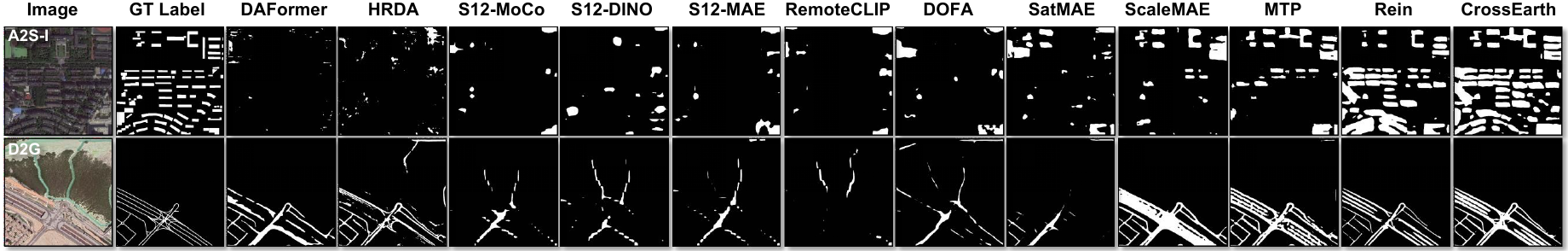}
    \caption{Visualizations of segmentation maps on road detection, building extraction under global generalization.}
    \label{road_building}
\end{figure*}

\textbf{Disaster Assessment (Potsdam and RescueNet)} Given the uncertainty of disasters, it is crucial to adopt a model that can be rapidly deployed across different regions without additional training time or data costs, highlighting the importance of RSDG. This emergency capability is essential for post-disaster tasks, such as estimating damage levels and planning rescue routes, especially in years with frequent disasters. To evaluate this capability, we construct two related benchmarks: P(r)2Res and P(i)2Res. For the gap of unseen regions, we first discuss P(r)2Res, as shown in Table \ref{Rescue}. 

In the P(r)2Res benchmark, CrossEarth stands out particularly on the Building class, achieving 59.7\% mIoU and surpassing Rein by a large margin of +10.2\%. Compared to other VFMs, the gap is even more pronounced: RS-specific models such as S12-MoCo, DOFA, and RemoteCLIP fail to generalize to this category, often staying below 10\% mIoU. More advanced RS VFMs like SatMAE, ScaleMAE, and MTP offer improvements (13.8\%, 32.9\%, and 40.9\%, respectively), yet still fall far short of reliable disaster-relevant building segmentation. 


We attribute this to: (1) Earth-Style Injection with multi-task learning enables stronger cross-domain representation extraction, allowing the model to capture structural and semantic variations of buildings under post-disaster conditions; (2) dataset characteristics—RescueNet emphasizes building damage assessment, with many images sampled such that buildings dominate the scene, making this class particularly critical; (3) the object-centric bias of DINOv2, which aligns with the building-focused nature of RescueNet. Together, these factors explain why CrossEarth achieves such strong gains on buildings, highlighting its practical value for post-disaster assessment where accurate building segmentation is essential.

\textbf{Global Generalization} As shown in Table~\ref{loveda}, in the original LoveDA experiments, the improvement of CrossEarth over Rein was relatively modest. To further assess generalization, we mapped the OpenEarthMap labels to the LoveDA dataset and directly evaluated the Urban2Rural-trained weights on the OpenEarthMap validation set. On globally distributed scenes, CrossEarth demonstrates much stronger generalization, achieving 49.2\% mIoU, which surpasses Rein (45.9\%, +3.3\%) and outperforms MTP by nearly 10\%. In contrast, most other models collapse in performance, suggesting that the moderate gains on LoveDA are largely tied to dataset constraints, while CrossEarth reveals its true capability when evaluated globally. For road extraction on the GlobalRoadNet dataset (D2G), as shown in Table~\ref{road_building_table}, CrossEarth also achieves the best result, surpassing Rein by +1.2\%. This aligns with the trend in Table~\ref{loveda}, where CrossEarth exhibits stronger advantages when transferred to global benchmarks. Additional qualitative results are provided in Figure \ref{road_building}.

These results collectively validate the effectiveness of CrossEarth and clearly show its superiority in cross-domain generalization in globally distributed scenarios.

\subsection{Generalization to Unseen Spectral Band}
Similar to the challenge of unseen regions, experiments in this section mainly evaluate the model's capacity to generalize to unseen spectral bands. Here, we still utilize ISPRS Potsdam and Vaihingen datasets and construct the P(i)2P(r) and P(r)2P(i) benchmarks across different channels.

\textbf{Potsdam and Vaihingen}  
As shown in Table~\ref{p2v}, in the P(r)2P(i) benchmark, CrossEarth achieves 77.8\% mIoU, surpassing Rein (+1.5\%) and MTP (+13.3\%), while other RS-specific VFMs, such as SatMAE (35.2\%) and ScaleMAE (46.6\%), perform far worse under spectral shifts. Rein, whose backbone is pretrained on natural RGB images, suffers a performance drop trained on IR-R-G inputs, reaching only 69.2\% mIoU. In the reverse P(i)2P(r) setting, CrossEarth again leads with 76.0\% mIoU, exceeding Rein by +6.8\% and remaining well above all other VFMs. These results indicate that conventional RS VFMs struggle with unseen spectral bands, whereas CrossEarth effectively bridges such gaps and maintains top-tier performance in both directions.

\subsection{Generalization to Unseen Region and Spectral Band}
\label{region and band}
This setting aims to examine the models' generalizability across regions and spectral bands simultaneously.  Compared to the above experiments, the task is more difficult and poses a higher requirement for model capabilities.

\textbf{Potsdam and Vaihingen} We also use these datasets to construct two additional benchmarks: P(r)2V and V2P(r). As shown in Table \ref{p2v}, model performance shows a noticeable decline compared to the accuracies on the settings of P(i)2V and V2P(i). DAFormer, HRDA, and MTP all suffer severe performance drops, with mIoU decreasing by more than 20\% when switching from P(i)2V to P(r)2V. Likewise, RS VFMs such as SatMAE and ScaleMAE remain well below 40\%, highlighting the particular challenge of adapting to joint regional and spectral-band shifts. In contrast, CrossEarth’s design integrates both RS knowledge learning and cross-domain generalization. Consequently, CrossEarth achieves the best performance with 61.9\% mIoU on P(r)2V and 52.7\% mIoU on V2P(r), outperforming Rein by 5.8\% mIoU and 2.1\% mIoU, respectively.

We also present qualitative results in Figure \ref{p2v_seg}. Due to the difficulty of this setting, model predictions show noticeable differences from the ground-truth labels. In the P(r)2V benchmark, almost all models tend to be affected by the clutter class. For DAFormer and HRDA, this influence is particularly severe, impacting the building and low vegetation categories. When using VFMs, such as MTP and Rein, this disturbance is somewhat alleviated. Notably, CrossEarth significantly reduces this misclassification, as evidenced by the marked decrease in misclassified red regions.

\textbf{Disaster Assessment (Potsdam and RescueNet)} In the P(i)2Res benchmark (Table \ref{Rescue}), CrossEarth achieves the best overall performance with 45.5\% mIoU, slightly surpassing Rein (+0.1\%). The most notable improvement is observed in the Building class, where CrossEarth attains 60.6\%, yielding a remarkable gain of +13.3\% over Rein. By contrast, RS-specific VFMs such as S12-MoCo and RemoteCLIP collapse below 11\%, while more advanced models like SatMAE (13.7\%) and ScaleMAE (35.0\%) also fall far short. Although MTP improves to 36.8\%, it still lags considerably behind CrossEarth. These results demonstrate CrossEarth’s robustness in disaster-related building segmentation, even under the dual challenge of unseen regions and spectral shifts.

\input{table6-1} 

\subsection{Generalization to Unseen Region and Platform}
We further consider the variations introduced by different data acquisition platforms. RS imagery can generally be divided into two types: satellite and aerial. The images from different sources have significant domain gaps in resolution, object scale, and cover range.  Satellite images typically have lower resolutions compared to aerial images, making them ideal for large-scale monitoring, while aerial imagery is suitable for capturing detailed land cover information.

\textbf{Building Extraction (WHU Building)}
To explore the issues outlined above, we use the widely known WHU Building dataset and conduct experiments between WHU-Aerial and WHU-Satellite\uppercase\expandafter{\romannumeral2} images, resulting in two settings: Aerial to Satellite (A2S) and Satellite to Aerial (S2A). As shown in Table \ref{road_building_table}, the A2S setting is extremely challenging. DA methods such as DAFormer and HRDA collapse below 8\% mIoU, while RS VFMs including SatMAE, ScaleMAE, and RemoteCLIP also fail to adapt, mostly staying under 15\%. Even stronger RS VFMs like MTP reach only 23.6\%, and Rein, the best among baselines, yields 30.8\%. These results indicate that models trained on high-resolution aerial imagery struggle to generalize to satellite images, underscoring the asymmetric nature of cross-platform transfer. In contrast, the reverse setting (S2A) proves more successful, with most models attaining considerably higher scores. This asymmetry may stem from the fact that satellite images contain smaller and more abstract object information, which makes adaptation difficult. Nevertheless, CrossEarth effectively bridges the platform gap, achieving a substantial gain of +13.5\% mIoU over Rein (from 30.8\% to 44.3\%). We attribute this improvement to the geospatial design of CrossEarth, which enables it to capture building distributions and structures in RS scenarios, thus facilitating generalization even in lower-resolution domains with more abstract semantics. The sub-optimal performance of CrossEarth on the S2A benchmark will be further analyzed in the supplementary material.

\textbf{Global Generalization} Table~\ref{road_building_table} further reports results on building extraction. On the globally distributed WHU Building Aerial-to-Satellite-I (A2S-I) benchmark, CrossEarth achieves a +9.1\% gain over the baseline, highlighting its strength to generalize at the global scale. We also provide qualitative visualizations of model predictions in Figure \ref{road_building}. These results further illustrate that, compared to other models, CrossEarth delivers clearer visual outcomes for building recognition, exhibiting superior cross-domain adaptability.

\begin{figure*}[t]
    \centering
    \includegraphics[width=1\textwidth]{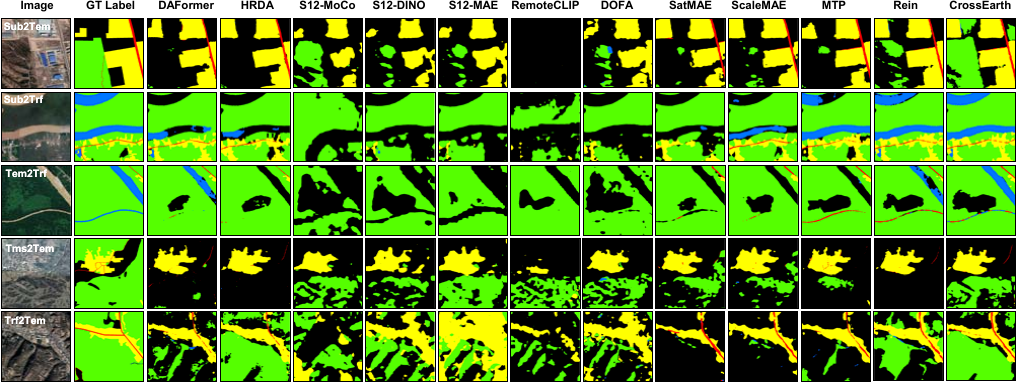}
    \caption{Visualizations of predicted segmentation maps of DA models and VFMs on CASID benchmarks. Here, \textcolor[RGB]{255, 0, 0}{red} is the road class, \textcolor[RGB]{255, 255, 0}{yellow} is the building class, \textcolor[RGB]{0, 112, 255}{blue} is the water class, \textcolor[RGB]{85, 255, 0}{green} is the forest class, and black is the background class. }
    \label{casid_visualization}
\end{figure*}

\input{main_ablation_table}

\subsection{Generalization Unseen Region and Climate}

Last but not least, we consider a factor often overlooked by the RS image interpretation community: climate. In our view, both DA and DG aim to train a unified model with robust generalizability across various domains. If we aim to develop a model capable of generalizing to any area on Earth, climate is an essential issue. Different climates bring significant variations, such as (1) various building densities influenced by cultural and living conditions, (2) distinct plant distributions due to temperature and humidity differences, and (3) variations in water body areas related to drought levels. Thus, climate presents a valuable direction for researching cross-domain semantic segmentation, though few existing methods currently focus on it.

Considering these issues, we construct DG benchmarks using  CASID dataset \cite{liu2023large}. We conduct 12 out-of-domain and 4 in-domain experiments, as shown in Table \ref{casid_exp}. We first focus on the in-domain experiments, where training and testing are conducted within the same domain. Both DAFormer and HRDA achieve strong results under the supervised setting. However, in most out-of-domain cases, including Sub2Tem, Sub2Trf, and Tms2Trf, their performance drops considerably, indicating limited generalizability. Under DG tasks, the gap among RS VFMs becomes even more pronounced: many collapse below 30\% mIoU, and even SatMAE and ScaleMAE, though stronger (33.1–62.5\%), still fall short of well-performing baselines such as Rein. Although MTP and Rein exhibit more stable generalization, they do not consistently outperform DA models, underscoring the challenges posed by substantial climate and regional discrepancies.

In contrast, CrossEarth achieves SOTA performance across most benchmarks. It surpasses DA models and existing VFMs on in-domain benchmarks and maintains competitive accuracy under domain shifts, outperforming Rein by +4.9\% mIoU in Sub2Tms and by +3.3\% in Tms2Tem. These results highlight CrossEarth’s strong segmentation capability and superior generalizability.

In addition to the quantitative experiments, we present predicted segmentation maps in Figure \ref{casid_visualization}. In the Sub2Tem setting, CrossEarth demonstrates superior recognition of forest areas compared to other models, consistent with the results in Table \ref{casid_exp}. Overall, these visual prediction maps provide an intuitive representation of the qualitative findings. Based on this comprehensive set of experiments, we conclude that while CrossEarth is more effective in bridging climate gaps than current other models, there are still many challenges to be addressed in cross-climate research.

\subsection{Ablation Studies}
\label{final ablation}

We conduct ablation studies on CASID, as shown in Table \ref{all_ablation}, with the following observations.

(1) Effectiveness of Main Components (R1–R8): Each module contributes distinct benefits. GSE (R2) brings notable gains on the in-domain dataset (Subms), though its effect on cross-domain benchmarks is limited. In contrast, MIM (R3) and Style Injection (R4) yield more improvements on cross-domain tasks, supporting better generalization. Their combinations (R5–R7) further enhance performance, and the full CrossEarth (R8) achieves the best overall results, confirming the complementarity among the modules.

(2) Cross Attention vs FFN in Injector (R9): We compare the Injector implemented with a simple linear projection (FFN) against our cross-attention design. The results show that both variants outperform the baseline, but cross-attention achieves overall better performance, indicating stronger feature interaction and more effective fusion of geospatial queries with backbone features. This confirms that cross-attention is a more suitable choice for the Injector.

(3) Effect of Sample Gate Probability $p$ (R11–R14): We examine how different injection probabilities $p$ affect performance. A small ratio of styled samples (e.g., $p=0.1$) achieves the best results, while larger ratios gradually reduce accuracy, and an excessive ratio ($p=0.9$) leads to clear degradation. This suggests that a limited amount of styled data is beneficial for enhancing diversity and improving generalization, whereas too many styled samples overwhelm the training distribution and may limit the model’s ability to capture the original data characteristics.

(4) Loss Design for MIM (R15–R16): We compare $\ell_1$ and $\ell_2$ losses for MIM. Although $\ell_2$ yields certain improvements, it underperforms $\ell_1$ overall. A possible explanation is the characteristics of CASID, where background or clutter pixels dominate (often over 85\%). In such imbalanced settings, $\ell_1$ may provide a stronger penalty on foreground errors, which helps reconstruction consistency. Therefore, our final design (R8) adopts $\ell_1$ for both losses.

\subsection{Visualization}

\begin{figure}[]
    \centering
    \includegraphics[width=1\linewidth]{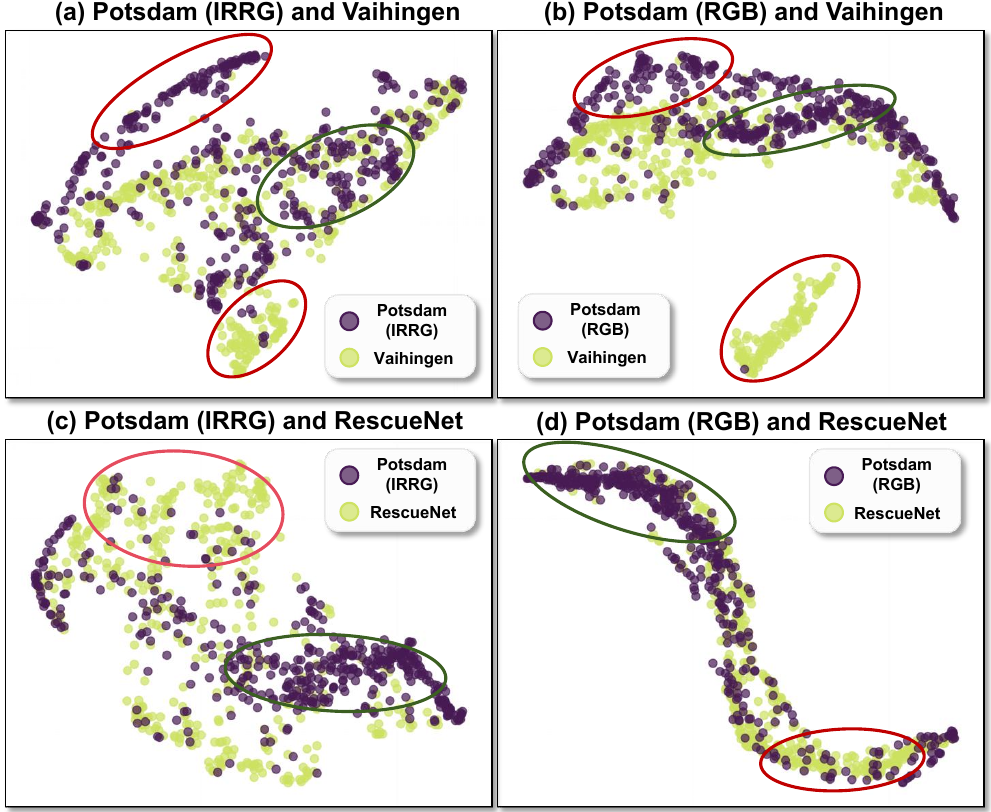}
    \caption{The UMAP visualization of the region and spectral band gaps. Dots in different colors represent the feature map pixel distributions from different domains. The green circle highlights areas with closely aligned distributions, indicating smaller domain gaps. Conversely, the red circle marks more distant areas, representing larger domain gaps.}
    \label{umap_domain_gap}
\end{figure}

\begin{figure}[t]
    \centering
    \includegraphics[width=1\linewidth]{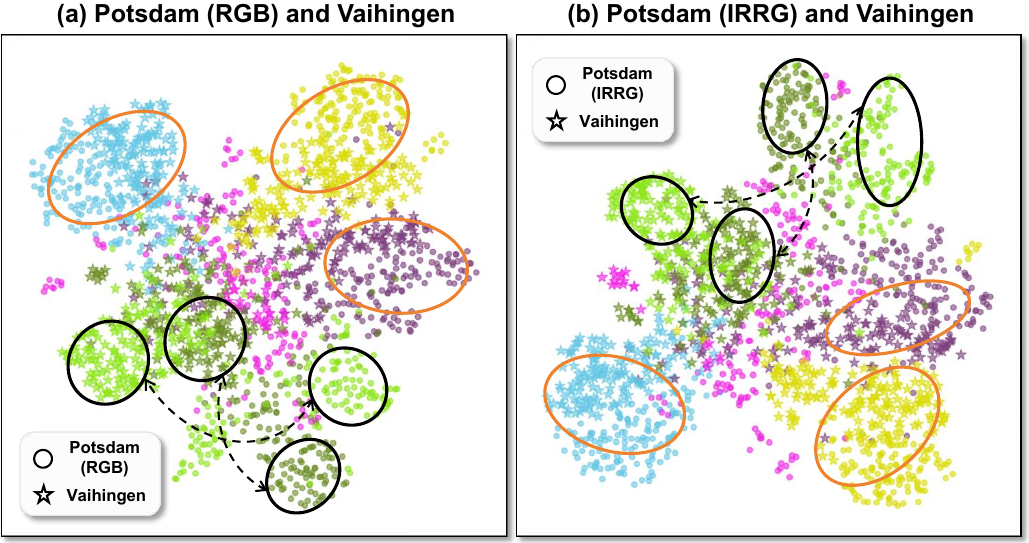}
    \caption{The UMAP of inter-domain feature clustering. Dots and stars represent two different domains. For color-label mapping: \textcolor[RGB]{126, 63, 126}{purple} represents the background class, \textcolor[RGB]{99, 199, 230}{cyan} represents the building class, \textcolor[RGB]{107, 141, 34}{dark green} represents the low vegetation class, \textcolor[RGB]{139, 230, 19}{light green} represents the tree class, \textcolor[RGB]{220,220,0}{yellow} represents the car class, and \textcolor[RGB]{244,34,232}{red} represents the clutter class. The orange circle highlights areas where features from both domains are highly similar, while the black circle indicates areas where low vegetation and tree classes are often misinterpreted.} 
    \label{umap_domain_invariant}
\end{figure}

\textbf{Domain Gaps} In Sec. \ref{region and band}, we compare performance in P(r)2V with P(i)2V, attributing the performance degradation to combined domain gaps. To support this assumption, we use the UMAP technique \cite{umap} to visualize domain gaps more intuitively, as shown in Figure \ref{umap_domain_gap}, where features are extracted from the last layer of the backbone network. The figure shows larger gaps between Potsdam (RGB) and Vaihingen, as well as between Potsdam (IRRG) and RescueNet, suggesting that the large domain distribution gaps are likely related to spectral band discrepancies. These observations align well with the performance variations in Table \ref{p2v} and Table \ref{Rescue}, where smaller domain gaps facilitate easier generalization and yield higher accuracies, while larger domain gaps present greater challenges for model adaptation. Additionally, the analyses in this section further validate the rationale behind benchmark settings.

\textbf{CrossEarth’s Representative Ability} In Figure \ref{umap_domain_invariant}, we also employ the UMAP technique to reduce the dimension of features extracted from two different domains, which are represented as dots and stars, respectively. From the orange circles, we observe that features extracted by CrossEarth for the same class across different domains cluster closely together, forming well-defined groups in feature space. Despite domain differences, features for each class maintain strong cohesion, demonstrating CrossEarth’s ability to learn robust, domain-invariant features. Beyond domain invariance, the features show excellent separation between classes, with each class forming a distinct cluster. This high inter-class separability highlights CrossEarth’s strong representational capability, effectively capturing clear boundaries between categories. Another noticeable aspect is the slight overlap between the low vegetation and tree classes, as shown by the black circles. Given the natural similarity between these classes, this minor overlap is understandable and suggests that the model captures realistic characteristics during representation learning. While CrossEarth demonstrates a strong ability to distinguish different categories, this small area of overlap may also indicate the inherent challenge of fully separating highly similar classes, particularly in RSDG scenarios. Additional visualizations of other datasets are provided in the supplementary material.

\begin{figure*}[t]
    \centering
    \includegraphics[width=1\linewidth]{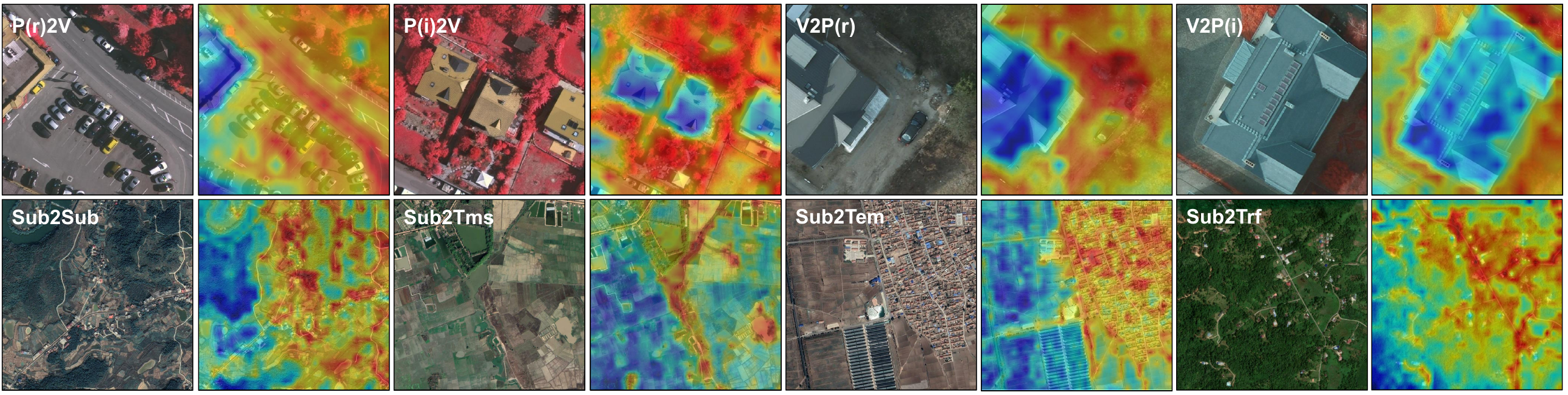}
    \caption{Visualization of the features extracted by CrossEarth across different benchmark experiments.}
    \label{feat_map}
\end{figure*}

\textbf{Geospatial Features} In Figure \ref{feat_map}, we present feature maps of CrossEarth on benchmarks based on ISPRS and CASID datasets. The features are obtained from the last layer of the backbone network. It is important to note that these visualizations represent feature maps, not heatmaps. Therefore, we mainly focus on the semantic boundaries within the features rather than on color intensities.

It can be seen that, in P(i)2V, despite buildings being surrounded by “red” trees in IR-R-G bands while the model is trained on RGB images, the features reveal clear semantic distinctions between buildings and trees. In V2P(r) and V2P(i), the feature maps effectively delineate the contours and internal structures of buildings. Additionally, in the CASID experiments, although the low-resolution satellite images contain diverse classes and complex object distributions, such as intricate road networks and buildings, the feature maps still capture high-quality semantics of roads, with road trajectories clearly indicated by red regions. Furthermore, these feature maps provide a nuanced understanding of background elements, where the terrain contours are distinctly outlined, as marked by blue regions. Notably, all these feature maps are obtained by applying the model to unseen domain scenes, showcasing CrossEarth's strong cross-domain capabilities across various gaps, consistent with the analyses related to Figure \ref{umap_domain_invariant}. More feature maps will be visualized in the supplementary material.

\section{Conclusion}

In this paper, we present \textbf{CrossEarth}, the first VFM specifically designed for RSDG semantic segmentation. To address the challenge of cross-domain generalization, we develop two key components: the Earth-Style Injection pipeline, which enriches the training data with diverse domain distributions, and the Multi-Task Training pipeline, which extracts representative semantic features suited for cross-domain scenes. These components jointly enhance DG capabilities from both data and model perspectives. Additionally, to encourage future RSDG research, we meticulously collect existing RS semantic segmentation datasets to create a comprehensive cross-domain generalization benchmark, providing rigorous evaluations of model generalizability. Extensive experiments demonstrate the superior performance and versatility of CrossEarth on multiple DG task settings involving region, spectral band, platform, and climate domain gaps, surpassing current advanced DA methods and VFMs tailored for the RS field. We hope this work will attract more attention from the RS community toward DG and inspire deeper exploration in this field. We also anticipate that CrossEarth will serve as a powerful baseline model to promote future innovations in the RSDG field.



\bibliographystyle{IEEEtran}
\bibliography{reference}
\twocolumn
\newpage
\makeatletter
\newcommand{\repeatmaketitle}{%
  \if@twocolumn
    \twocolumn[{
      \begingroup
      \centering
      \vspace*{3.25\baselineskip}
      {\@IEEEcompsoc@title@font \@title \par}
      \vspace{11pt}
      {\@IEEEcompsoc@author@font
       \def\IEEEauthorhalign##1{\centering##1}
       \@author \par}
      \vspace{2\baselineskip}
      \endgroup
    }]%
  \else
    \thispagestyle{empty}
    \begingroup
    \centering
    \vspace*{3.25\baselineskip}
    {\@IEEEcompsoc@title@font \@title \par}
    \vspace{11pt}
    {\@IEEEcompsoc@author@font \@author \par}
    \vspace{2\baselineskip}
    \endgroup
  \fi
  \captionsetup{
    font=normalfont,       
    textfont=normalfont,   
    size=normalsize,       
}
}

\newcommand{\@IEEEcompsoc@title@font}{%
  \sffamily\Huge\selectfont
  \ifCLASSOPTIONconference\large\fi
}

\newcommand{\@IEEEcompsoc@author@font}{%
  \sffamily\large\selectfont
  \ifCLASSOPTIONconference\sublargesize\fi
}

\def\IEEEauthorblockN#1{%
  \vskip0.5ex
  {\bfseries #1}\par
}

\def\IEEEauthorblockA#1{%
  \vskip0.25ex
  {\itshape #1}\par
}

\makeatother
\title{\includegraphics[width=1.0cm]{figures/crossearth.png} CrossEarth: Geospatial Vision Foundation Model for Domain Generalizable Remote Sensing 
Semantic Segmentation\\
---Supplementary Material---}


\author{
Ziyang Gong$^*$,
Zhixiang Wei$^*$,
Di Wang$^*$,
Xiaoxing Hu$^*$,
Xianzheng Ma, 
Hongruixuan Chen, 
Yuru Jia, 
Yupeng Deng, 
Zhenming Ji$^\dagger$,
Xiangwei Zhu$^\dagger$, ~\IEEEmembership{Member,~IEEE,} 
Xue Yang$^\dagger$, ~\IEEEmembership{Member,~IEEE,} \\
Naoto Yokoya, ~\IEEEmembership{Member,~IEEE,}
Jing Zhang,~\IEEEmembership{Senior Member,~IEEE,} \\
Bo Du,~\IEEEmembership{Senior Member,~IEEE,} 
Junchi Yan,~\IEEEmembership{Senior Member,~IEEE,} \\
Liangpei Zhang,~\IEEEmembership{Fellow,~IEEE}

}
\makeatletter
\begingroup
\renewcommand{\@IEEEcompsoc@title@font}{\sffamily\Huge\selectfont}
\renewcommand{\@IEEEcompsoc@author@font}{\sffamily\large\selectfont}
\endgroup
\makeatother

\appendices

\section{Overview of the Supplementary Material}

This supplementary material provides an introduction of related work, additional experiments on model hyperparameters, implementation details of CrossEarth, specifics of the constructed Remote Sensing Domain Generalization (RSDG) benchmark, and further qualitative results omitted from the main paper. It is organized as follows:

In Sec.~\ref{sec:related works}, we illustrate a brief survey of current related works consisting of works in Remote Sensing Semantic Segmentation (Sec.~\ref{related work-1}) and Remote Sensing Cross Domain Semantic Segmentation (Sec.~\ref{related work-2}).

In Sec.~\ref{casid details}, we present the per-class performance of different methods on the CASID benchmark.

In Sec.~\ref{improvement directions}, we conduct more experiments on the D2M benchmark and discuss the potential future improvement directions of CrossEarth.

In Sec.~\ref{Sec:hyperparameters}, we discuss the hyperparameter settings of CrossEarth, including the in-depth analyses of the influence on CrossEarth brought by $\tau_m$ and $B$ (Sec.~\ref{sec:parameters ablation}), and the hyperparameter settings in each benchmark (Sec.~\ref{sec:detailed ablation}).

In Sec.~\ref{More details of the CrossEarth Benchmark}, we introduce more details of the CrossEarth Benchmark, including the introduction of the datasets used in constructing the CrossEarth benchmark (Sec.~\ref{sec:detailed benchmarks}) and the process of mapping labels (Sec.~\ref{sec: rescue-to-potsdam}).

In Sec.~\ref{sec: comprehensive visualization}, we demonstrate more qualitative results, including UMAP \cite{umap} visualizations, feature map visualizations, and predicted semantic segmentation maps of CrossEarth on diverse benchmarks.

\section{Related Work}
\label{sec:related works}

\subsection{Remote Sensing Semantic Segmentation}
\label{related work-1}
Semantic segmentation is a significant and challenging computer vision task that requires models to achieve pixel-level perception to accurately delineate object contours. Since the introduction of the Fully Connected Network (FCN) \cite{long2015fully}, semantic segmentation has made substantial progress across various 2-dimensional (2D) and 3-dimensional (3D) domains, including autonomous driving \cite{tsai2018learning, ma2022both, gong2024coda, gong2023train, li2024parsing, xiao20233d, bi2024dgss, bi2024generalized}, embodied AI \cite{nilsson2021embodied, ainetter2021end}, medical imaging \cite{hatamizadeh2021swin, shamshad2023transformers, hatamizadeh2022unetr, mazurowski2023segment, bi2024learning}, and natural imaging \cite{kirillov2023segment, xiao2024cat, ke2024segment}. In the RS field, semantic segmentation also has diverse applications \cite{lv2023deep}, such as urban planning \cite{rui2020survey, abdollahi2021multi}, land resource management \cite{tong2020land}, and environmental protection \cite{subudhi2021survey}. Many studies in RS semantic segmentation focus on in-domain data learning, often by developing novel techniques to enhance feature extraction. For example, \cite{zhang2020object, zheng2020parsing, zhong2022nt} propose multi-scale object optimization to strengthen representation learning; \cite{li2023synergistical} introduces novel attention mechanisms to improve critical feature recognition; \cite{xu2022feature} utilizes HRNet \cite{hrnet} to effectively learn high-resolution features, and \cite{unetformer} combines UNet \cite{ronneberger2015u} with ViT \cite{dosovitskiy2020image} to learn both global and local features. Although these approaches achieve strong performance on specific benchmarks, they do not focus on improving models' cross-domain capabilities.

\subsection{Remote Sensing Cross Domain Semantic Segmentation}
\label{related work-2}

In the field of RS, many cross-domain approaches \cite{tang2020srda, bengana2020improving, zhang2021curriculum} have demonstrated excellent performances, significantly reducing the need for labeled data. However, according to our literature review, most existing studies mainly focus on DA. For instance, works such as \cite{ayala2023diffusion, zhao2023label, ma2023unsupervised, shi2020end, liu2019unsupervised, tasar2020standardgan, benjdira2020data, li2023spgan, earth-adapter, yurudiffusion} leverage generative models, including GANs \cite{goodfellow2020generative} and diffusion models \cite{ho2020denoising, rombach2022high}, to assist RS semantic segmentation tasks. While \cite{xi2023multilevel, zhang2021stagewise, liu2021adversarial, deng2019large, zhu2023unsupervised, chen2023memory} combine contrastive and unsupervised learning to enhance both pixel-level and domain-level semantic feature learning for aerial scenes. In addition, \cite{ismael2023unsupervised, tasar2020daugnet, gao2023prototype, tasar2020semi2i} apply data augmentation techniques to improve latent representation learning of RS imagery. In contrast, the number of works concentrated on DG \cite{liang2024single, iizuka2023frequency} is much less than DA. Compared to DA, DG methods can be flexibly applied to out-of-domain RS scenarios without requiring additional training data. Regarding the significant application value of DG, we introduce a cross-domain generalization evaluation benchmark for RS semantic segmentation tasks, advocating the RS community to devote greater attention to this field.

\section{Detailed performance on CASID benchmark}
\label{casid details}

The detailed per-category performance of different methods on the CASID benchmark is reported in Table~\ref{casid}.

\section{Directions of Improvement}
\label{improvement directions}

As mentioned earlier, CrossEarth represents the first step toward developing foundational models for RSDG semantic segmentation, rather than a comprehensive solution for all benchmark scenarios. Consequently, it may not achieve optimal performance in certain cases. In this section, we conduct an in-depth analysis to explore potential improvements to enhance CrossEarth’s performance.

\textbf{Road Extraction (D2M)} In the D2M experiments, CrossEarth achieves sub-optimal performance, while MTP demonstrates SOTA results. We attribute this primarily to differences in backbone design: (1) CrossEarth: Uses DINO-V2 \cite{oquab2023dinov2} as the backbone, based on the standard ViT architecture, but lacks pre-training on RS images, resulting in limited relevant prior knowledge. (2) MTP: Employs RVSA \cite{rvsa}, a ViT-based architecture with a rotated varied-size window attention mechanism, which likely aligns well with road detection tasks, where roads appear in various orientations. Additionally, RVSA is pre-trained on RS imagery.

\begin{table}[t!]
    \centering
    \caption{Performance comparison of different methods on the D2M benchmark, shown by mIoU scores (\%).}
    
     \resizebox{0.7\linewidth}{!}{
    \begin{tabular}{ll}
    \toprule
        CrossEarth w/ other Models & D2M \\
       \midrule
          Rein \cite{Wei_2024_CVPR} &49.7\\
          CrossEarth on Rein (Ours)& 50.5 (+0.8)\\
          MTP \cite{wang2024mtp} & 54.3\\
          CrossEarth on MTP (Ours) &55.5 (+1.2)\\
        \bottomrule
    \end{tabular}}
    
    \label{d2m_mtp}
\end{table}

To validate this, we preserve the training paradigm of CrossEarth but replace the DINO-V2 backbone with the RVSA of MTP. The results in Table \ref{d2m_mtp} validate our assumption and demonstrate that CrossEarth indeed further improves the MTP’s generalizability from 54.3\% mIoU to 55.5\% mIoU. Nevertheless, it should be noted that despite the advantages of MTP, the RVSA design is not universally applicable across all RS scenarios, especially under DG settings. Therefore, exploring new attention mechanisms or innovative approaches that integrate the strengths of both CrossEarth and MTP may be promising for further RSDG research.

\textbf{Building Detection (S2A)} Unlike the significant improvement seen in the A2S experiment with CrossEarth, performance in the S2A experiment is less optimal compared to other methods. This difference may stem from CrossEarth's Earth-Style Injection pipeline, which makes adjustments to the style and content of images. In this case, the erroneous elimination of small buildings may limit CrossEarth’s comprehension of RS scenarios, particularly when trained with low-resolution satellite images. Enhancing CrossEarth’s ability to generalize from low-resolution to high-resolution images is a promising direction.

\section{Hyperparameters of CrossEarth}
\label{Sec:hyperparameters}
\subsection{Analysis of Hyperparameters}
\label{sec:parameters ablation}
Before delving into the specifics of our ablation studies, we first recap the final step in CrossEarth's process for generating styled image $X_S$ and masked image $X_M$. Before this step, the Mask Generator creates two distinct masks. Then these masks will compute the dot product with different images to produce $X_S$ and $X_M$, respectively. Consequently, we identify four hyperparameters that can be categorized into two groups guided by their usages of generating $X_S$ or $X_M$. To research the sensitivity of CrossEarth for $\tau_m$ and $B$ deeply, we arrange a series of ablation studies on these two groups of hyperparameters with $\tau_m \in \{0.1, 0.3, 0.5, 0.7, 0.9\}$ and $B \in \{16, 32, 64, 128, 256\}$.

For all experiments in this section, we choose the Subtropical Monsoon (Sub) climate of the CASID dataset \cite{liu2023large} as the source domain, and the other three climates: Temperate Monsoon (Tem), Tropical Monsoon (Tms), and Tropical Rainforest (Trf) as unseen domains.

\input{table8}
\input{table9}

\textbf{Mask Ratio and Patch Size In Styled Image} 
The ablation study results for the $X_S$ group are presented in Table \ref{x_s ablation}. Notably, a green upward arrow indicates that the result surpasses the baseline performance of Rein's method, while a red downward arrow signifies the contrary. It can be seen that an obvious trend emerges from all experiments in this group: as the values of $\tau_m$ and $B$ increase, there is a negative correlation between these values and the performance.

In the Sub2Tem benchmark, when $\tau_m$  is maintained at 0.1 and 0.3, we observe six green arrows, indicating improvements, compared to four red arrows, indicating deteriorations. However, when values of $\tau_m$ increase to 0.5, 0.7, and 0.9, the number of green arrows diminishes markedly. A similar pattern is observed for  $B$ in Sub2Tem: when $B$ is below 64, the count of green arrows exceeds that of red arrows, but this advantage lessens as $B$ reaches 128 and 256.

In the Sub2Tms benchmark, the negative correlation is less pronounced for $\tau_m$, with a notable exception when $\tau_m$ is 0.5, where red arrows outnumber green ones. However, the trend becomes more evident when focusing on the values of $B$. At $\tau_m$ =0.5, all red arrows only appear when $B \in \{64, 128, 256\}$,  suggesting poorer performance, while better results are observed when $B$ is 16 and 32. Specifically, when $B$=16, all results in this column exceed the baseline.  Additionally, when $B \in \{32, 64, 128\}$, the majority of outcomes in these columns show improvements.  However, in the column of $B$=256, most results fall below the baseline. This pattern aligns with the observations made in Sub2Tem.     

In Sub2Trf,  the trends observed in previous experiments persist, albeit with less pronounced effects. When focusing on the rows where $\tau_m$ is set to 0.1 and 0.3, we notice that the number of green arrows exceeds the number of red arrows.  However, as $\tau_m$ increases to 0.5, 0.7, and 0.9, the balance shifts, with red arrows outnumbering green ones. Examining the results from a columnar perspective reveals that while the $B=256$ column shows a considerable number of green arrows, the poorest results are concentrated in the $B=64$ and $B=128$ columns. These outcomes align with the observed negative correlation between the hyperparameters in the $X_S$ group and the cross-domain performance of CrossEarth. We maintain that this correlation is closely tied to the function of the Earth Style Embedding Augmentation Pipeline, which is designed to enhance the distribution coverage of the training datasets. If the values of $\tau_m$ and $B$ are increased excessively, the semantics and style of the original training images may be compromised, leading to improper punishments from $L_\delta$ and $L_{Seg}$. This, in turn, can result in performance degradation. Therefore, it is crucial to find an optimal balance in selecting $\tau_m$ and $B$ values to ensure that the model's performance is not adversely affected.

\textbf{Mask Ratio and Patch Size in Masked Image}
The ablation study results for the $X_M$ group, as depicted in Table \ref{x_m ablation}, reveal intriguing insights into the effects of varying mask ratio $\tau_m$ and patch size $B$ on cross-domain performances of CrossEarth. Contrary to the negative correlation observed in the first-group experiments, this group's results indicate a more nuanced relationship, suggesting that relatively larger values of $\tau_m$ are more conducive to the MIM process. 

In the Sub2Tem benchmark, the red arrows are dominant when $\tau_m$ is set to 0.1. As $\tau_m$ increases, the results become better, indicating that the values of $\tau_m$ are positively correlated to CrossEarth performances. However, we do not find a similar trend in the variation of $B$. These results probably reveal that the MIM process favors smaller and more occlusion, rather than large blocks of occlusion in generating $X_M$.

For the Sub2Tms benchmark, it is clear to see that except for bad performances when $B \in \{64, 128, 256\}$ with $\tau_m$ =0.1, almost other performances show improvements compared with baseline Rein. Also, two-thirds of red arrows appear in the $\tau_m$=1 row aligning with the above-mentioned tendency that larger $\tau_m$ is preferred. Besides, the dominant green arrows demonstrate the robustness of CrossEarth to the variation of hyperparameters in the $X_M$ group. 

In the Sub2Trf benchmark, the distribution of green and red arrows appears more balanced. Regardless of whether $\tau_m$ is less than, equal to, or greater than 0.5, the distribution of red arrows remains relatively uniform. Accounting for the baseline equivalence at 63.4, which is represented by the red arrows, the count of red arrows is evenly split between $\tau_m < 0.5$ and $\tau_m \geq 0.5$ scenarios. However, excluding the $\tau_m = 0.5$ data point, the general tendency towards larger $\tau_m$ values still holds.

The ablation study for the $X_M$ group underscores that larger mask ratios tend to be more effective, particularly when $\tau_m \geq 0.5$. This finding suggests that the masked image generation process may benefit from increased small occlusion, potentially allowing the model to focus more on the unmasked, informative regions of the input images.

\subsection{Hyperparameter Settings for Each Benchmark}
\label{sec:detailed ablation}
Following the experiments and analyses in Sec. \ref{sec:parameters ablation} of the main text, we provide the detailed settings of hyperparameters for each experiment, as shown in Table \ref{hyper_parameter}.

\begin{table}[h]
  \scriptsize
  \caption{Detailed hyper parameter settings of Mask Ratio $\tau_m$ and Patch size $B$ on different experiments.}
  \centering
  \begin{tabular}{lcc|cc}
  \toprule
  \multirow{2}{*}{Benchmark} & \multicolumn{2}{c}{Styled Image ($X_S$)} & \multicolumn{2}{c}{Masked Image ($X_M$)} \\
  \cline{2-5}
  & $\tau_m$ & $B$ & $\tau_m$ & $B$ \\
  \midrule
  \bfseries \textit{Postdam and Vaihingen} & \multicolumn{4}{c}{}\\
  \midrule
  P(i)2V & 0.7 & 64 & 0.7 &64 \\
  P(i)2P(r) & 0.7 & 64 & 0.7 &64 \\
  P(r)2V & 0.3 & 16 & 0.7 &64 \\
  P(r)2P(i) & 0.3 & 16 & 0.7 &64 \\
  V2P(i)  & 0.5 & 32 & 0.7 &64 \\
  V2P(r) & 0.5 & 32 & 0.7 &64 \\
  \midrule
  \bfseries \textit{LoveDA} & \multicolumn{4}{c}{}\\
  \midrule
   U2R & 0.1 & 64 & 0.7 &64 \\
   R2U & 0.5 & 16 & 0.7 &64 \\
  \midrule
  \bfseries \textit{DeepGloab and Massachusetts} & \multicolumn{4}{c}{}\\
  \midrule
  D2M & 0.3 & 16 & 0.5 & 16 \\
  \midrule
  \bfseries \textit{Potsdam and RescueNet} & \multicolumn{4}{c}{}\\
  \midrule
  P(i)2Res & 0.5 & 32 & 0.7 & 64 \\
  P(r)2Res & 0.3 & 16 & 0.7 & 64 \\
  \midrule
  \bfseries \textit{WHU Building} & \multicolumn{4}{c}{}\\
  \midrule
  A2S & 0.5 & 16 & 0.7 & 64 \\
  S2A & 0.5 & 16 & 0.7 & 16 \\
  \midrule
  \bfseries \textit{CASID} & \multicolumn{4}{c}{}\\
  \midrule
  Sub (Source-Only) & 0.1 & 64 & 0.7 & 64 \\
  Sub2Tem & 0.1 & 64 & 0.7 & 64 \\
  Sub2Tms & 0.1 & 64 & 0.7 & 64 \\
  Sub2Trf & 0.1 & 64 & 0.7 & 64 \\

  Tem (Source-Only) & 0.1 & 64 & 0.7 & 128 \\
  Tem2Sub & 0.1 & 64 & 0.7 & 128 \\
  Tem2Tms & 0.1 & 64 & 0.7 & 128 \\
  Tem2Trf & 0.1 & 64 & 0.7 & 128 \\
  
  Tms (Source-Only) & 0.1 & 64 & 0.7 & 64 \\
  Tms2Sub & 0.1 & 64 & 0.7 & 64 \\
  Tms2Tem & 0.1 & 64 & 0.7 & 64 \\
  Tms2Trf & 0.1 & 64 & 0.7 & 64 \\

  Trf (Source-Only) & 0.5 & 64 & 0.7 & 64 \\
  Trf2Sub & 0.5 & 64 & 0.7 & 64 \\
  Trf2Tem & 0.5 & 64 & 0.7 & 64 \\
  Trf2Tms & 0.5 & 64 & 0.7 & 64 \\
  \bottomrule
\end{tabular}
\label{hyper_parameter}

\end{table}

\section{More details of the CrossEarth Benchmark}
\label{More details of the CrossEarth Benchmark}
\subsection{Dataset
Introduction}
\label{sec:detailed benchmarks}
The created RSDG benchmark includes several widely-used RS segmentation datasets, with details provided below:

\textbf{ISPRS Potsdam and Vaihingen} are two fundamental datasets in RS semantic segmentation. They consist of aerial images captured over Potsdam and Vaihingen cities \cite{zhang2023pseudo}. The original Potsdam dataset contains three versions: R-G-B (3 channels), IR-R-G (3 channels), and R-G-B-IR (4 channels). Vaihingen images only contain IR-R-G band. In our experiments, we use RGB and IR-R-G bands of Potsdam. 
These two datasets were also adopted as the most widely-used DA benchmark in previous works \cite{xi2023multilevel, wang2023fine, liang2023multilevel, zhang2023pseudo, ni2023category, li2021learning, zhao2023residualgan, ma2023unsupervised, li2023spgan, wu2022deep, li2022unsupervised}. We follow the MMSegmentation \cite{mmseg2020} pre-process to obtain training and testing images.

\textbf{RescueNet }\cite{rahnemoonfar2022rescuenet} is a high-resolution aerial dataset captured by unmanned aerial systems (UAS) aiming to detect buildings suffering from disasters to facilitate rescue work. RescueNet contains 11 categories, including Background, Water, Building-No-Damage, Building-Medium-Damage, Building-Major-Damage, Building-Total-Destruction, Vehicle, Road-Clear, Road-Blocked, Tree, and Pool. 

\textbf{LoveDA} \cite{wang2021loveda} dataset consists of semantic segmentation and DA (Rural and Urban domains) benchmarks. LoveDA has 7 categories, Background (Bkgd), Building (Bldg), Road (Rd), Water (Wtr), Barren (Barr), Forest, and Agriculture (Agri). In our experiments, we conduct generalization with two benchmarks: Urban-to-Rural and Rural-to-Urban.

\textbf{WHU Building} dataset provides both aerial and satellite images. In the benchmark, the DG within the WHU Building dataset focuses on cross-platform and region (Aerial-to-Satellite and Satellite-to-Aerial \cite{liang2023multilevel, liang2023unsupervised}).

\textbf{DeepGlobe} \cite{demir2018deepglobe} 2018 Satellite Image Understanding Challenge includes three public competitions for segmentation, detection, and classification tasks on satellite images. Referring to previous works \cite{zhou2018d}, We adopt the DeepGlobe dataset to serve as a source domain dataset in the DeepGlobe-to-Massachusetts experiment due to the inaccessibility of test set labels.

\textbf{Massachusetts} \cite{MnihThesis} road dataset covers various urban, suburban, and rural regions with an area of over 2600 square kilometers. In our experiments, we leverage this dataset as the target domain in DeepGlobe-to-Massachusetts.

\textbf{CASID} \cite{liu2023large} is the first RS dataset developed to address DA challenges across four climates: Subtropical Monsoon (Sub), Temperate Monsoon (Tem), Tropical Monsoon (Tms), and Tropical Rainforest (Trf). CASID focuses on five semantic categories: Background (Bkdg), Building (Bldg), Forest (Frst), Road (Rd), and Water (Wtr). Since CASID has only recently been made public, the preprocessing scripts are still limited. To this end, we implemented custom scripts to divide the dataset into training and validation sets and crop images to 1024$\thinspace\times\thinspace$1024 resolutions, following the original CASID paper’s guidelines. All associated scripts will be made publicly available on our project page.

\textbf{OpenEarthMap} \cite{openearthmap} is a benchmark dataset designed for global high-resolution land cover mapping. It contains 5,000 aerial and satellite images that have been manually annotated with eight land cover classes, spanning 97 regions in 44 countries across six continents. With a ground sampling distance of 0.25–0.5 m, models trained on OpenEarthMap demonstrate strong generalization capabilities worldwide, enabling direct, off-the-shelf applications to a wide range of geospatial tasks.

\textbf{GlobalRoadNet} \cite{globalroadnet} is a large-scale training dataset for global road extraction, built from very-high-resolution satellite imagery and open-source geospatial data. It contains 47,210 samples from 121 capital cities across six continents. Experiments show that GlobalRoadNet substantially improves road extraction performance and offers strong potential for updating global road maps such as OpenStreetMap.

\subsection{Class Mapping Details}
\textbf{RescueNet to Potsdam}
\label{sec: rescue-to-potsdam}
The label mapping process in Table \ref{label change: rescue2potsdam} converts the categories in RescueNet into Potsdam to ensure consistency across datasets. Specifically, all building-related categories in RescueNet are merged into Bldg (Building), vehicle categories are mapped to Car, water and pool categories are assigned to Bkdg (Background), road-related categories are mapped to Surf (Surface), and background categories are mapped to Clut (Clutter). In addition, the original Potsdam vegetation category (Vegt) is excluded from the final set. After this conversion, the unified label categories are standardized as Surf, Bldg, Tree, Car, and Clut.
\input{rescue-to-potsdam}
\textbf{OpenEarthMap to LoveDA}
The label mapping process in Table \ref{label_change_oem} converts the categories in OpenEarthMap into LoveDA. In this mapping, bareland is assigned to Barr (Barren), rangeland and developed space are merged into Bkgd (Background), road is mapped to Rd (Road), tree is mapped to Frst (Forest), water is mapped to Wtr (Water), agriculture land is mapped to Agri (Agriculture), and building is mapped to Bldg (Building). After these conversions, the final standardized categories are defined as Bkgd, Bldg, Rd, Wtr, Barr, Frst, and Agri.
\input{oem-to-loveda}

\section{More Qualitative Results}

\label{sec: comprehensive visualization}
\begin{figure*}[t]
    \centering
    \includegraphics[width=0.95\linewidth]{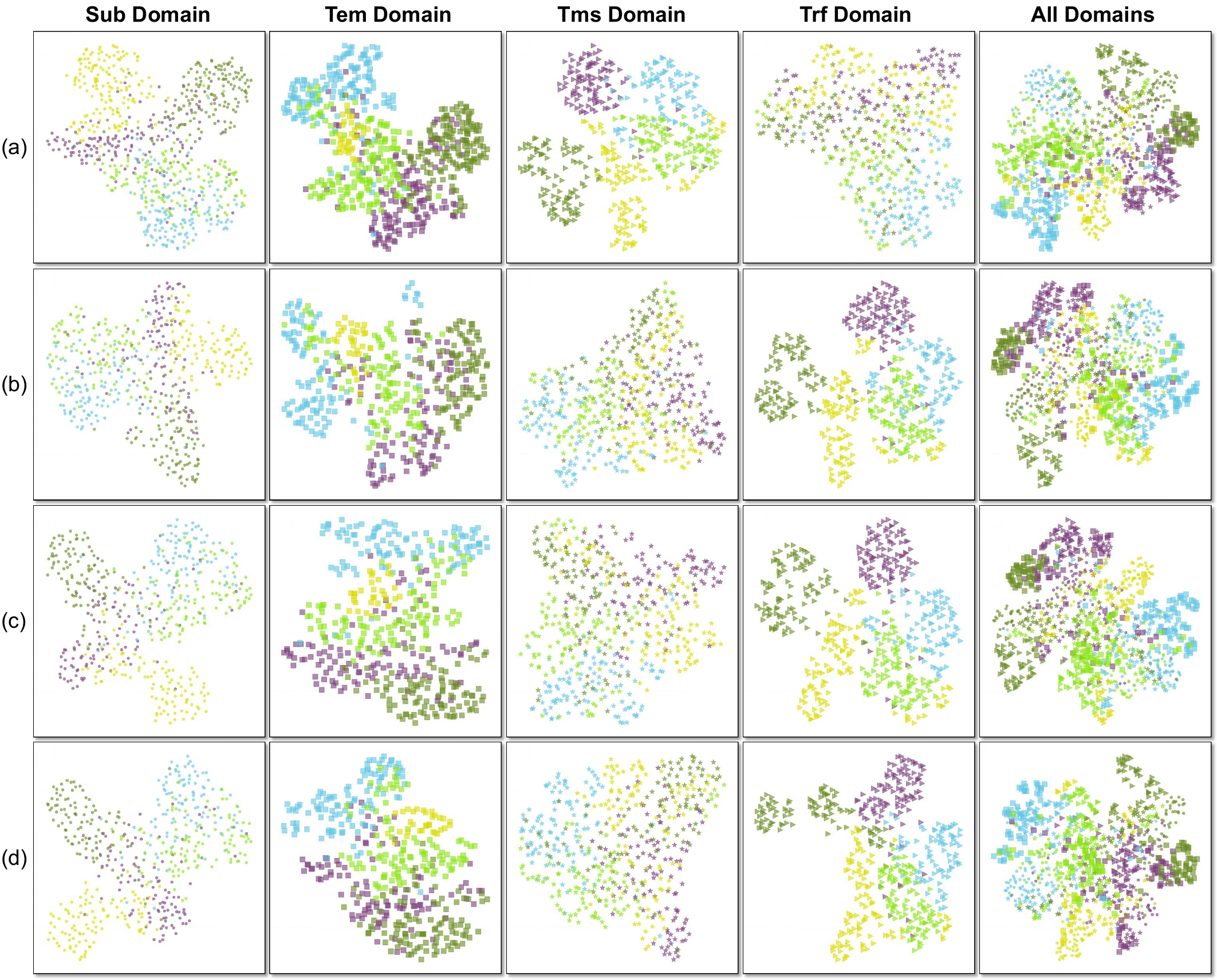}
    \caption{UMAP visualizations of CrosEarth on CASID benchmarks. Image titles like \textbf{Sub Domain} mean the different unseen domains. (a), (b), (c), and (d) separately mean the experiments of adopting different source domains: Sub, Tem, Tms, and Trf. For color-label mapping: \textcolor[RGB]{126, 63, 126}{purple} represents the background class, \textcolor[RGB]{99, 199, 230}{blue} represents the building class, \textcolor[RGB]{107, 141, 34}{green} represents the forest class, \textcolor[RGB]{139, 230, 19}{green} represents the road class, and \textcolor[RGB]{220,220,0}{yellow} represents the clutter class.} 
    \label{umap}
\end{figure*}

\begin{figure*}[t]
    \centering
    \includegraphics[width=0.95\linewidth]{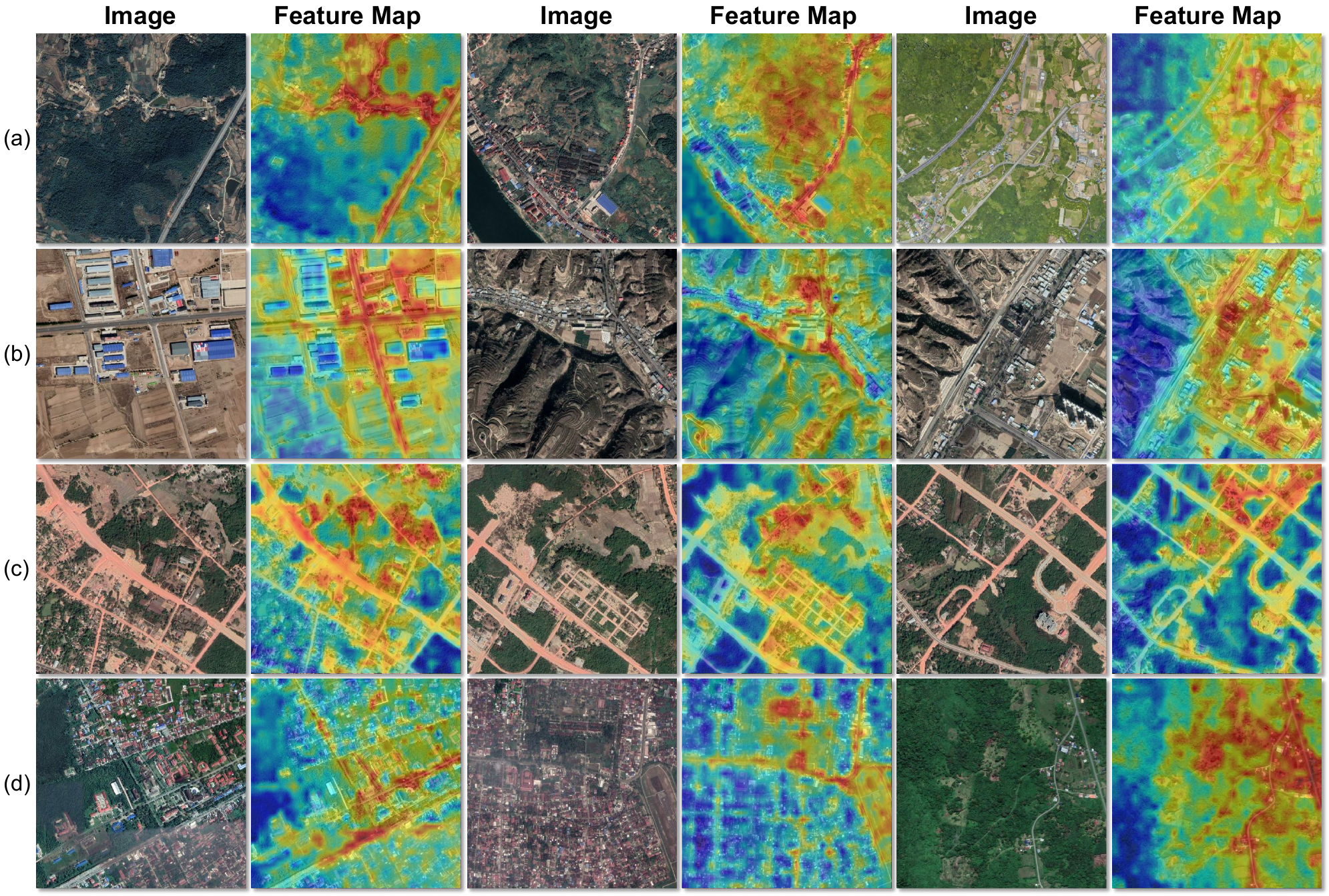}
    \caption{Feature map visualizations of CrossEarth on CASID benchmarks. (a), (b), (c), and (b) separately means Sub2Sub, Sub2Tem, Sub2Tms and Sub2Trf. } 
    \label{heatmap}
\end{figure*}

\begin{figure*}[t]
    \centering
    \includegraphics[width=0.95\linewidth]{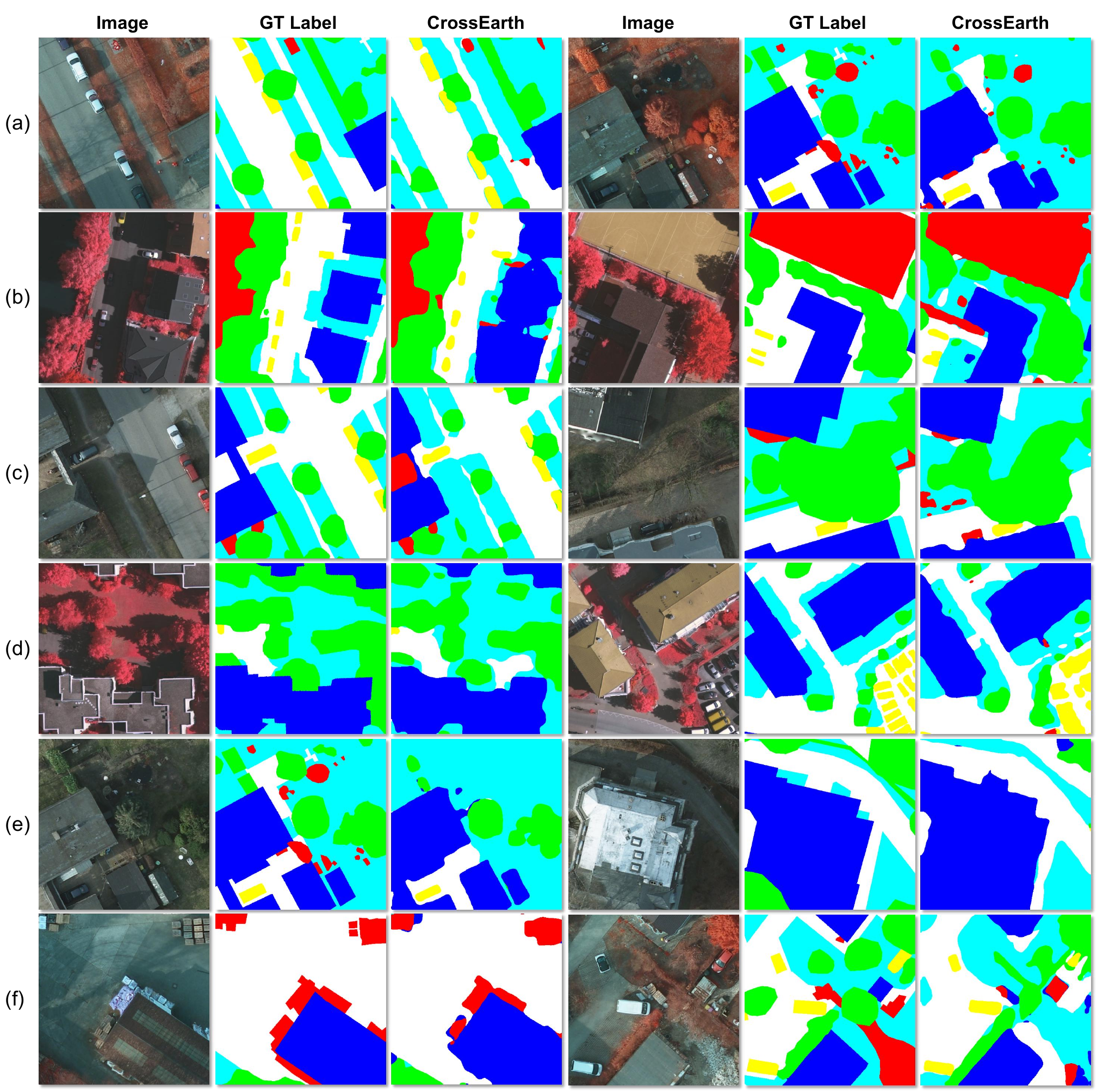}
    \caption{Predicted semantic segmentation maps of CrossEarth on Potsdam and Vaihingen benchmarks. Images from (a) to (f) respectively represent P(r)2P(i), P(r)2V, P(i)2P(r), P(i)2V, V2P(r), and V2P(i). For the color map, white is the impervious surface class, \textcolor[RGB]{255,0,0}{red} is the clutter class, \textcolor[RGB]{0, 0, 255}{blue} is the building class, \textcolor[RGB]{0, 255, 255} {cyan} is the low vegetation class, \textcolor[RGB]{0, 255, 0}{green} is the tree class, and \textcolor[RGB]{255, 255, 0}{yellow} is the car class.} 
    \label{supp_pv_seg}
\end{figure*}

\begin{figure*}[t]
    \centering
    \includegraphics[width=1\linewidth]{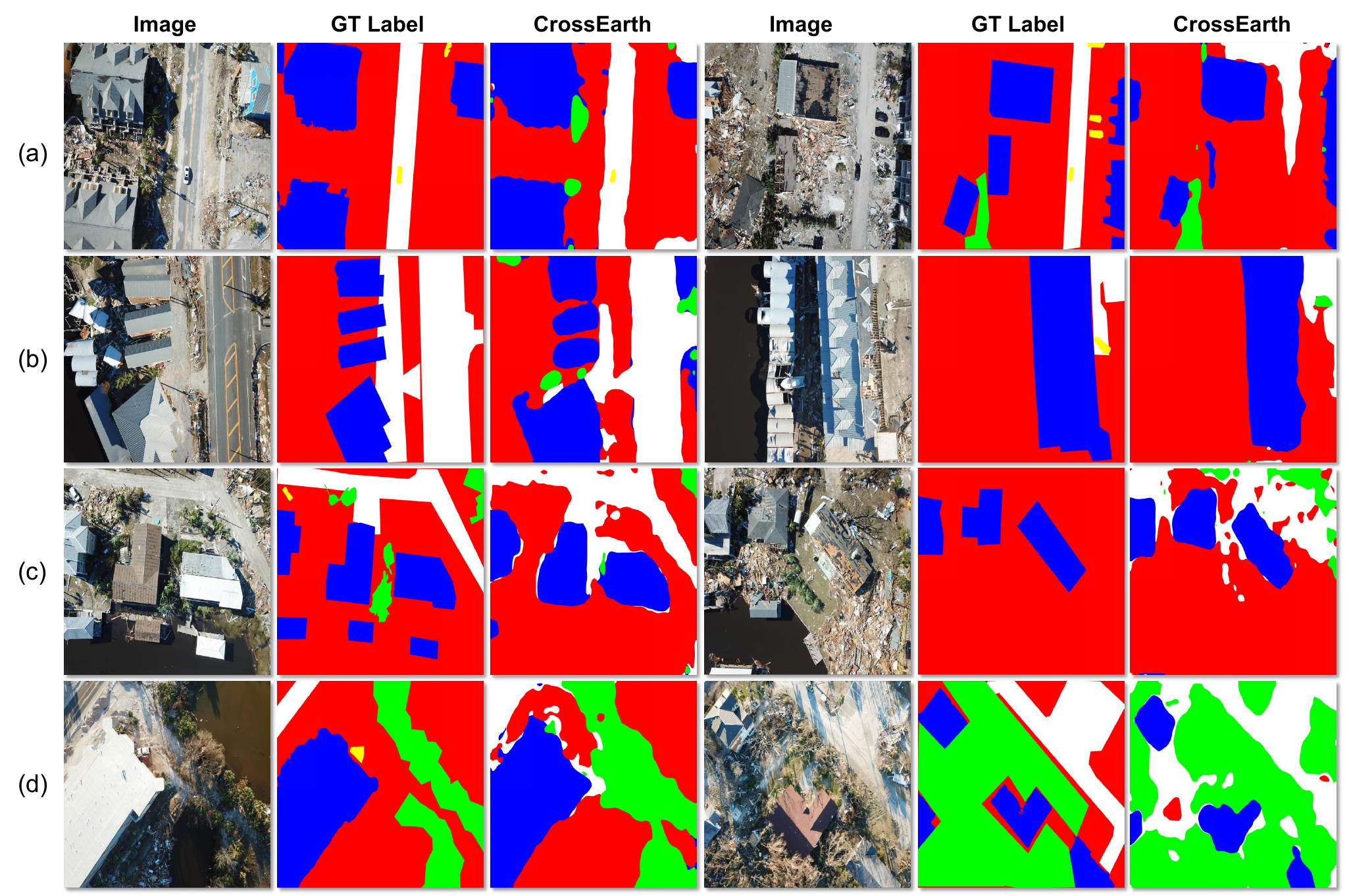}
    \caption{Predicted semantic segmentation maps of CrossEarth on P(r)2Res and P(i)2Res benchmarks. Images in (a) and (b) represent P(r)2Res and the rest represent P(i)2Res. For the color map, white is the impervious surface class, \textcolor[RGB]{255,0,0}{red} is the clutter class, \textcolor[RGB]{0, 0, 255}{blue} is the building class, \textcolor[RGB]{0, 255, 0}{green} is the vegetation class, and \textcolor[RGB]{255, 255, 0}{yellow} is the car class.} 
    \label{supp_p2res}
\end{figure*}

\begin{figure*}[t]
    \centering
    \includegraphics[width=1\linewidth]{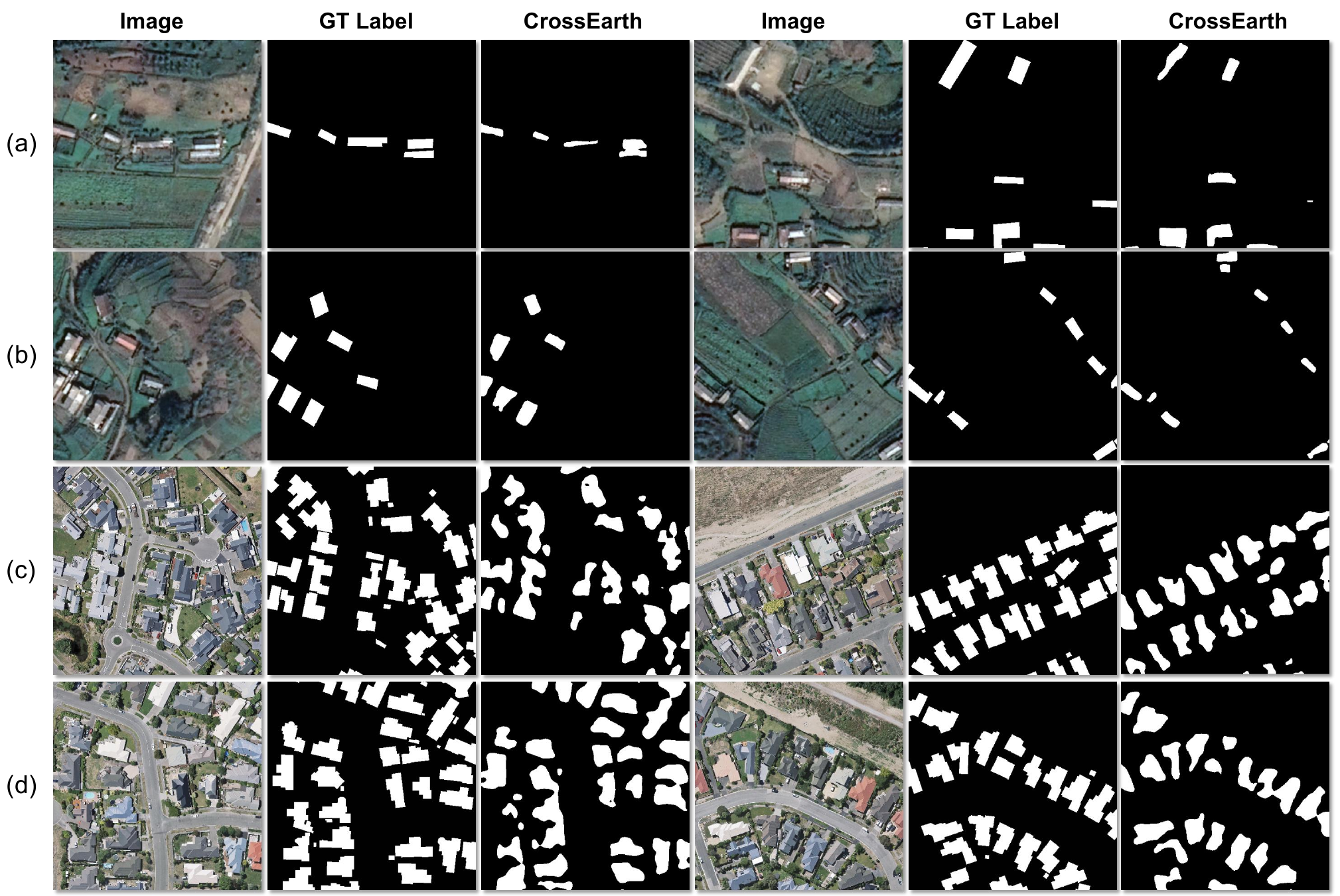}
    \caption{Predicted semantic segmentation maps of CrossEarth on A2S and S2A building detection benchmarks. Images in (a) and (b) represent A2S, and the rest represent S2A. } 
    \label{supp_building_seg}
\end{figure*}

\begin{figure*}[t]
    \centering
    \includegraphics[width=0.95\linewidth]{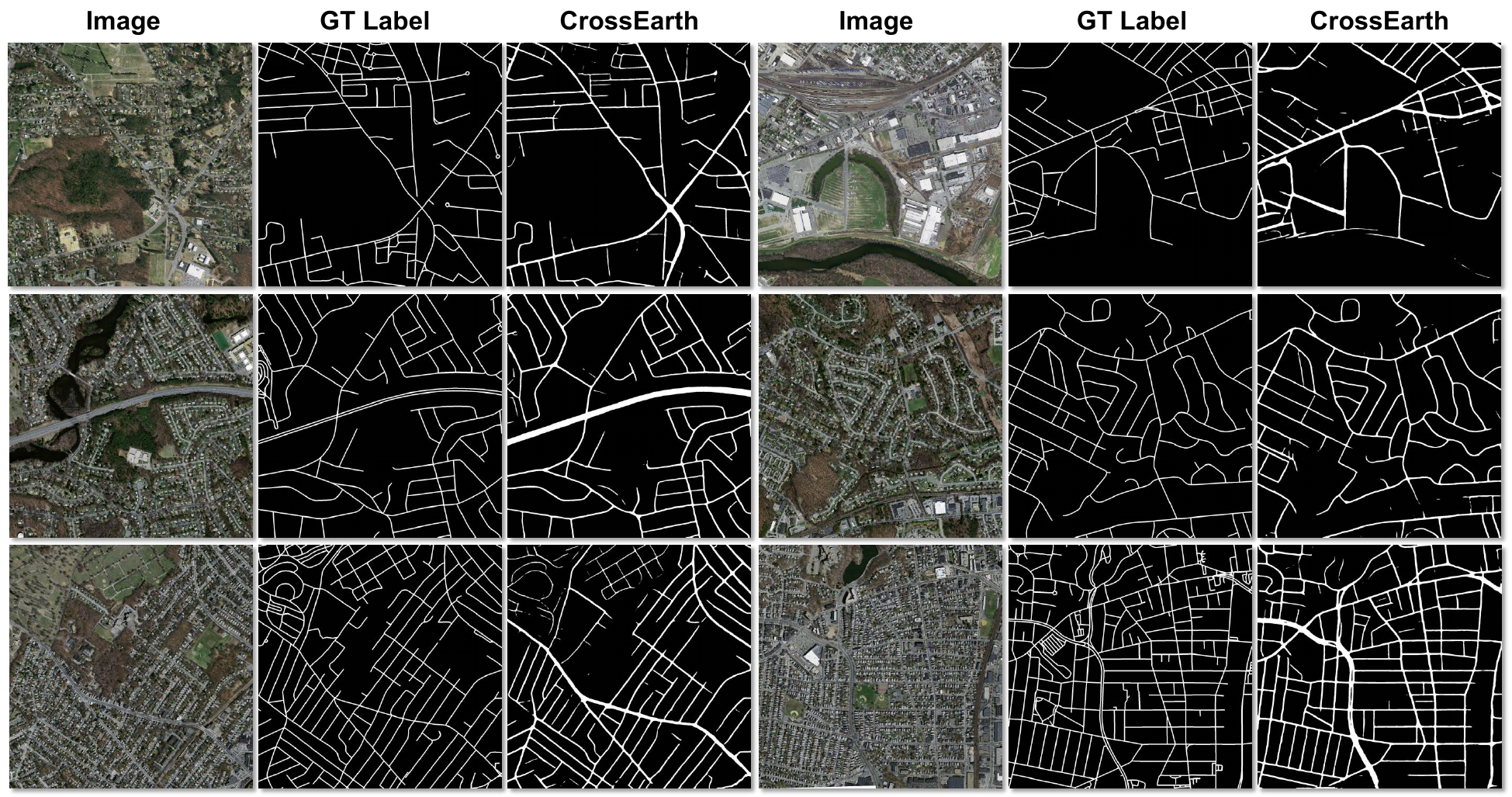}
    \caption{Predicted semantic segmentation maps of CrossEarth on D2M road extraction benchmarks.  } 
    \label{supp_road_seg}
\end{figure*}

\begin{figure*}[t]
    \centering
\includegraphics[width=0.95\linewidth]{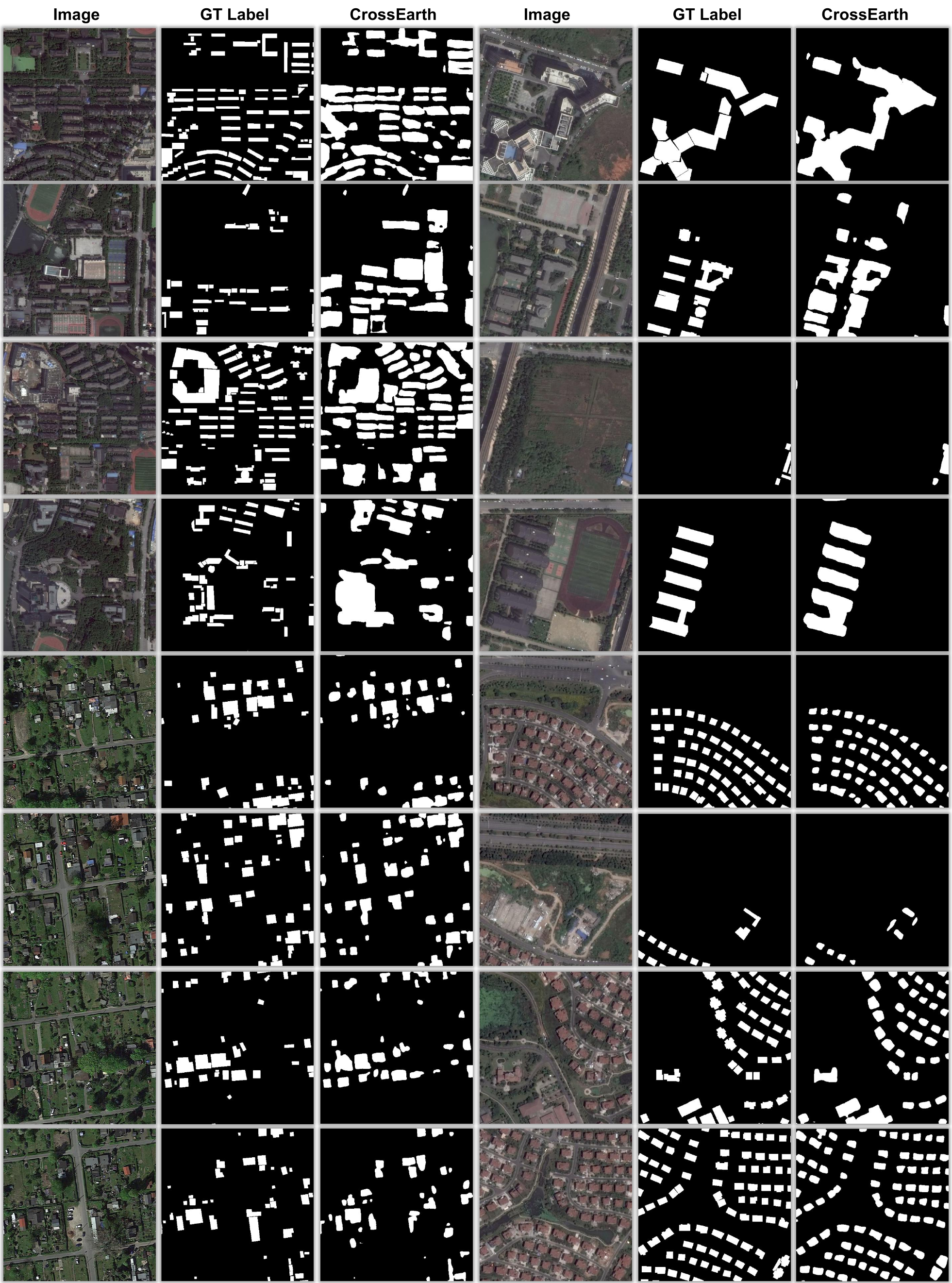}
    \caption{Predicted semantic segmentation maps of CrossEarth on A2S-I building detection benchmarks.}
    \label{supp_a2s-1}
\end{figure*}

\begin{figure*}[t]
    \centering
\includegraphics[width=0.95\linewidth]{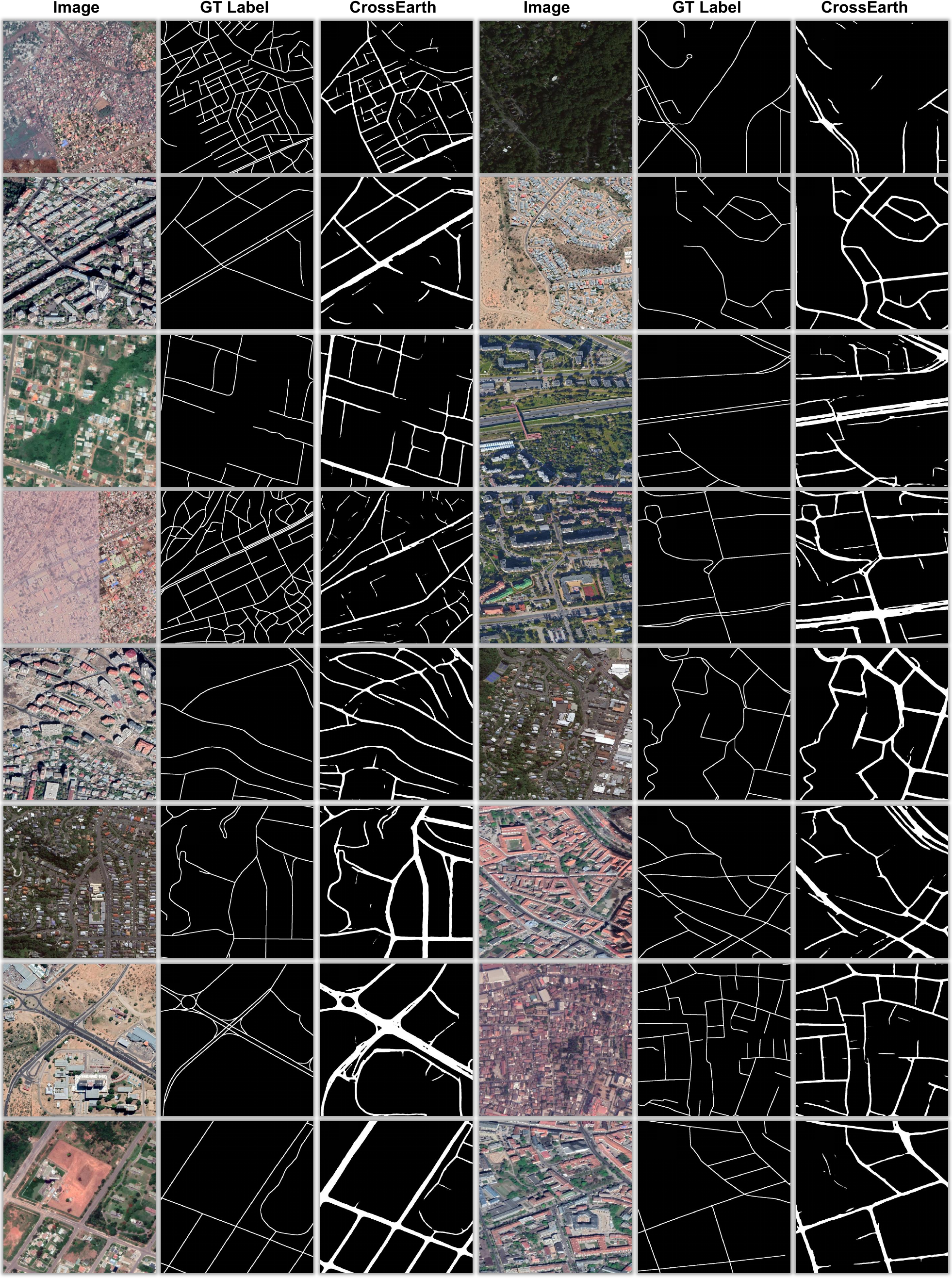}
    \caption{Predicted semantic segmentation maps of CrossEarth on D2G road extraction benchmarks.}
    \label{supp_d2g}
\end{figure*}

\begin{figure*}[t]
    \centering
    \includegraphics[width=0.95\linewidth]{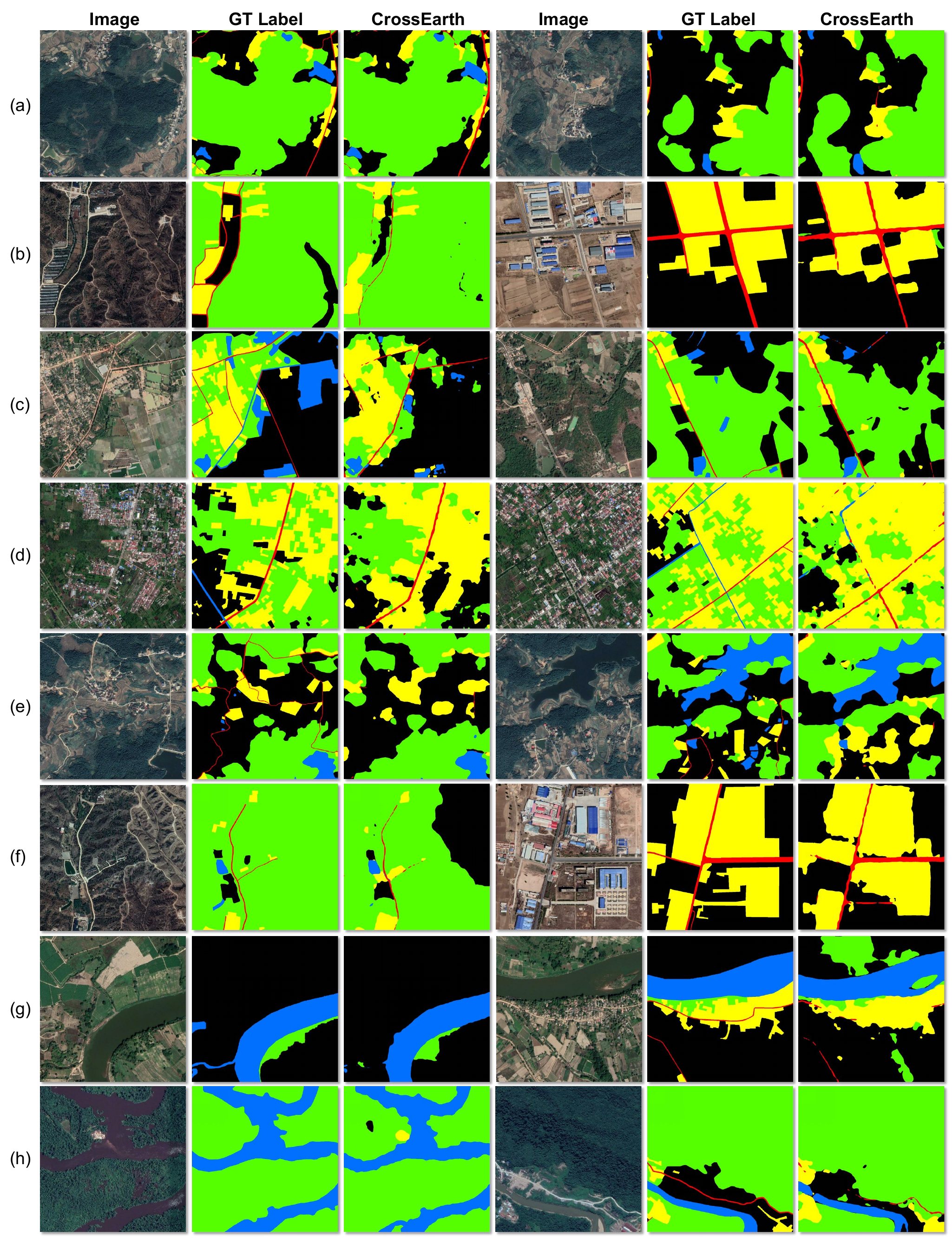}
    \caption{Predicted semantic segmentation maps of CrossEarth on CASID benchmarks \cite{liu2023large}. Images in (a), (b), (c), and (d) respectively represent Sub2Sub, Sub2Tem, Sub2Tms, and Sub2Trf. Images in (e), (f), (g), and (h) respectively represent Tem2Sub, Tem2Tem, Tem2Tms, and Tem2Trf.  \textcolor[RGB]{255, 0, 0}{Red} is the road class, \textcolor[RGB]{255, 255, 0}{yellow} is the building class, \textcolor[RGB]{0, 112, 255}{blue} is the water class, \textcolor[RGB]{85, 255, 0}{green} is the forest class, and black is the background class. } 
    \label{casid1}
\end{figure*}

\begin{figure*}[t]
    \centering
    \includegraphics[width=0.95\linewidth]{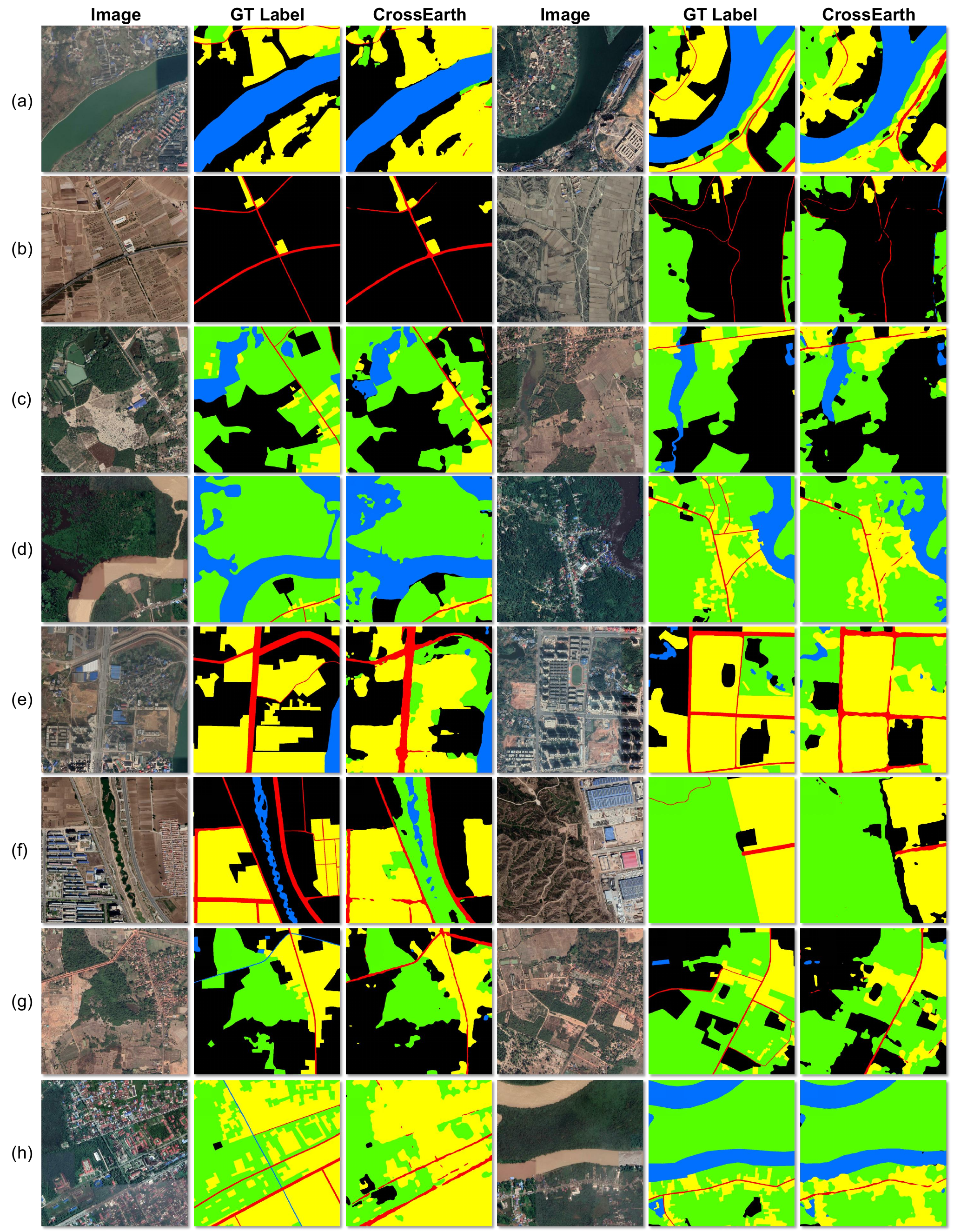}
    \caption{Predicted semantic segmentation maps of CrossEarth on CASID benchmarks \cite{liu2023large}. Images in (a), (b), (c), and (d) respectively represent Tms2Sub, Tms2Tem, Tms2Tms, and Tms2Trf. Images in (e), (f), (g), and (h) respectively represent Trf2Sub, Trf2Tem, Trf2Tms, and Trf2Trf.  } 
    \label{casid2}
\end{figure*}

\begin{figure*}[t]
    \centering
    \includegraphics[width=0.95\linewidth]{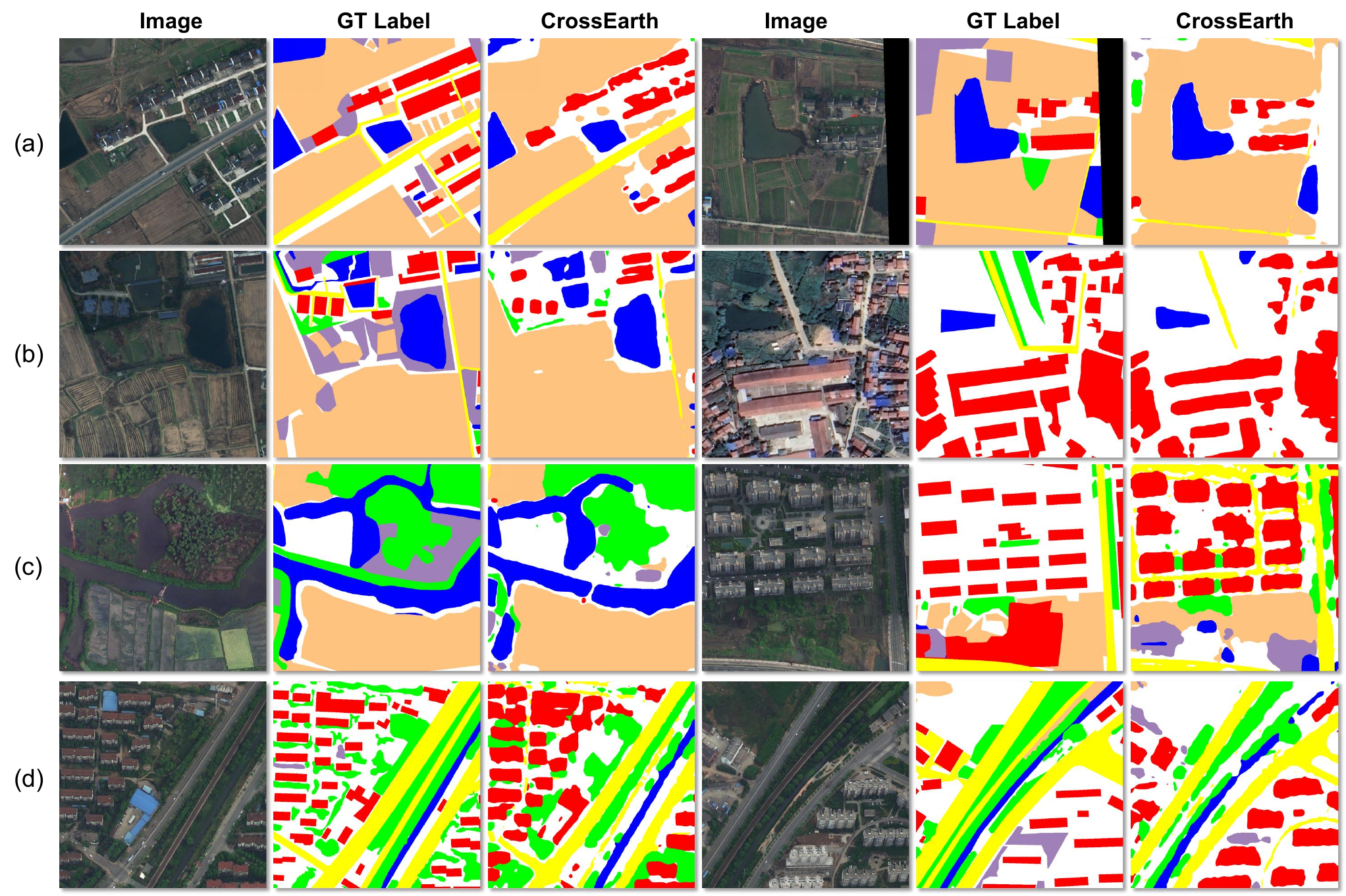}
    \caption{Predicted semantic segmentation maps of CrossEarth on LoveDA benchmarks \cite{wang2021loveda}. Images in (a) and (b) respectively represent U2R. Images in (c) and (d) respectively represent R2U. For the color map, \textcolor[RGB]{255, 0, 0}{red} is the building class, \textcolor[RGB]{255, 255, 0}{yellow} is the road class, \textcolor[RGB]{0,0,255}{blue} is the water class, \textcolor[RGB]{159, 129, 183}{purple} is the barren class, \textcolor[RGB]{0, 255, 0}{green} is the forest class, \textcolor[RGB]{255, 195, 128}{brown} is the agriculture class. } 
    \label{supp_loveda_seg}
\end{figure*}

\begin{figure*}[t]
    \centering
    \includegraphics[width=0.95\linewidth]{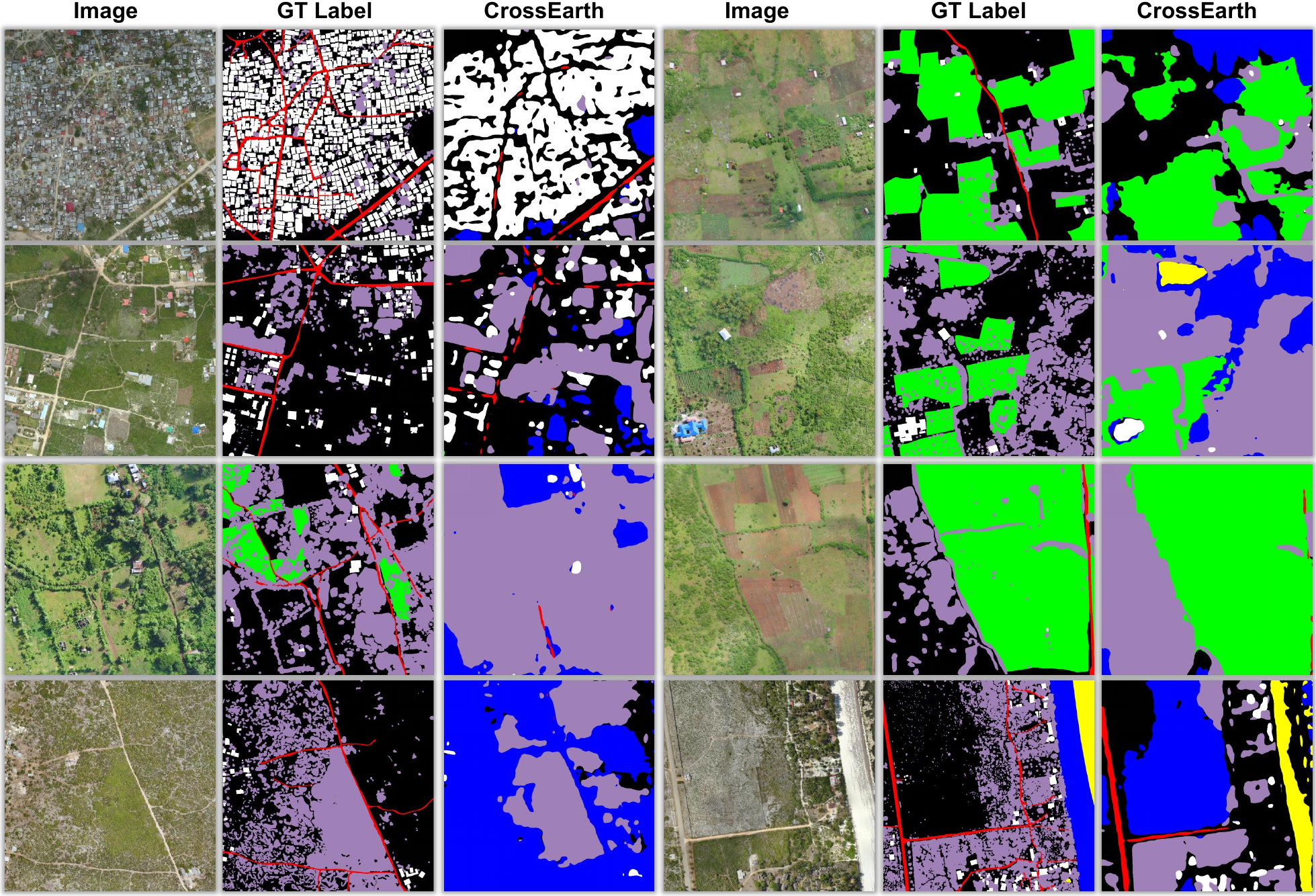}
    \caption{Predicted semantic segmentation maps of CrossEarth on LoveDA-Urban to OpenEarthMap benchmarks \cite{wang2021loveda}. \textcolor[RGB]{255, 0, 0}{red} is the building class, \textcolor[RGB]{255, 255, 0}{yellow} is the road class, \textcolor[RGB]{0,0,255}{blue} is the water class, \textcolor[RGB]{159, 129, 183}{purple} is the barren class, \textcolor[RGB]{0, 255, 0}{green} is the forest class, \textcolor[RGB]{255, 195, 128}{brown} is the agriculture class. }
    \label{supp_u2opm}
\end{figure*}

Figure \ref{umap} displays UMAP visualizations of CASID. Figure \ref{heatmap} showcases representative feature map visualizations of the same dataset. Semantic segmentation maps are presented for Potsdam and Vaihingen in Figure \ref{supp_pv_seg}, while Potsdam and RescueNet are depicted in Figure \ref{supp_p2res}. Building segmentation results are illustrated in Figure \ref{supp_building_seg} and \ref{supp_d2g}, with road extraction results in Figure \ref{supp_road_seg} and Figure \ref{supp_a2s-1}. CASID samples are highlighted in Figure \ref{casid1} and Figure \ref{casid2}, LoveDA outcomes are featured in Figure \ref{supp_loveda_seg}, and OpenEarthMap predictions are shown in Figure \ref{supp_u2opm}.

\input{table6_sup}

\end{document}

%% file: table2.tex
\begin{table*}
    \centering
    \caption{The composition of the curated RSDG benchmark. We classify all task settings based on the primary domain gaps between the source and unseen domain datasets. Task abbreviations, the number of training and testing images, image sizes, and category counts are also provided. $^\bullet$ means the target domain is a global-distributed dataset.}
    \resizebox{\textwidth}{!}{
    \begin{tabular}{ccccccccc}
    \toprule
    \multirow{1}{*}{Domain Gap}&\multicolumn{1}{c}{Dataset} & \multicolumn{1}{c}{Source Domain} &\multicolumn{1}{c}{Target Domain} & Abbreviation & Train Number & Test Number & \multicolumn{1}{c}{Image Size} & Categories \\
    \midrule
    \multirow{8}{*}{Unseen Region} &\multirow{2}{*}{ISPRS Potsdam, Vaihingen}& Potsdam (IRRG) & Vaihingen (IRRG) & P(i)$\thinspace$2$\thinspace$V &3456$\thinspace$ & 398$\thinspace$& \multirow{2}{*}{512$\thinspace\times\thinspace$512}& 6 \\
    & & Vaihingen (IRRG) & Potsdam (IRRG) & V$\thinspace$2$\thinspace$P(i) &344$\thinspace$& 2016$\thinspace$ &  & 6 \\
    &\multirow{2}{*}{LoveDA\cite{wang2021loveda}}& LoveDA-Urban & LoveDA-Rural& U$\thinspace$2$\thinspace$R& 1156 & 992 & \multirow{2}{*}{1024$\thinspace\times\thinspace$1024} &7 \\
    &&LoveDA-Rural & LoveDA-Urban & R$\thinspace$2$\thinspace$U &1366 &667 & & 7\\
  &\multirow{1}{*}{DeepGlobe \cite{chen2019GLNET}, Massachusetts \cite{MnihThesis}}& \multirow{2}{*}{DeepGlobe} & \multirow{2}{*}{Massachusetts} & \multirow{2}{*}{D$\thinspace$2$\thinspace$M} & \multirow{2}{*}{6226}& \multirow{2}{*}{49} &\multirow{2}{*}{1024$\thinspace\times\thinspace$1024} &\multirow{2}{*}{1} \\
    &(Building Extraction)&&&&&&&\\
  &\multirow{1}{*}{ISPRS Potsdam, RescueNet \cite{rahnemoonfar2022rescuenet}}& \multirow{2}{*}{Potsdam (RGB)} & \multirow{2}{*}{RescueNet (RGB)} & \multirow{2}{*}{P(r)$\thinspace$2$\thinspace$Res} & \multirow{2}{*}{3456} & \multirow{2}{*}{449} & \multirow{2}{*}{512$\thinspace\times\thinspace$512}&\multirow{2}{*}{5}\\
    &(Disaster Assessment)&&&&&&&\\
    
    & LoveDA, OpenEarthMap \cite{openearthmap} & LoveDA-Urban & OpenEarthMap$^\bullet$  & U2OEM & 1156 & 384 & 1024$\times$1024 &  6 \\
    
    & (Building Detection)  & & & & & & &\\
    
    & DeepGlobe, GlobalRoadNet \cite{globalroadnet} & DeepGlobe & GlobalRoadNet$^\bullet$  & D2G& 6226 & 241 & 1024$\times$1024 & 1\\
    
    & (Road Extraction)  &  & & & & & &\\
    \midrule
    \multirow{2}{*}{Unseen Spectral Band}&\multirow{2}{*}{ISPRS Potsdam}& Potsdam (RGB) & Potsdam (IRRG)& P(r)$\thinspace$2$\thinspace$P(i) &\multirow{2}{*}{3456$\thinspace$}& \multirow{2}{*}{2016$\thinspace$}& \multirow{2}{*}{512$\thinspace\times\thinspace$512}& \multirow{1}{*}{6} \\
    & & Potsdam (IRRG) & Potsdam (RGB)& P(i)$\thinspace$2$\thinspace$P(r) & &&&6\\
    \midrule
     \multirow{4}{*}{Unseen Region and Spectral Band}&\multirow{2}{*}{ISPRS Potsdam, Vaihingen}& Potsdam (RGB) & Vaihingen (IRRG) & P(r)$\thinspace$2$\thinspace$V &3456$\thinspace$&398$\thinspace$ & \multirow{2}{*}{512$\thinspace\times\thinspace$512}&6\\
    && Vaihingen (IRRG) & Potsdam (RGB) & V$\thinspace$2$\thinspace$P(r)& 344$\thinspace$& 2016$\thinspace$&  & 6 \\
    &\multirow{1}{*}{ISPRS Potsdam, RescueNet \cite{rahnemoonfar2022rescuenet}}& \multirow{2}{*}{Potsdam (IRRG)} & \multirow{2}{*}{RescueNet (RGB)}& \multirow{2}{*}{P(i)$\thinspace$2$\thinspace$Res} & \multirow{2}{*}{3456} & \multirow{2}{*}{449} & \multirow{2}{*}{512$\thinspace\times\thinspace$512}&\multirow{2}{*}{5}\\
    &\multirow{1}{*}{(Disaster Assessment)} &  & & & & & &
    \\
    \midrule
    \multirow{2}{*}{Unseen Region and Platform}&\multirow{2}{*}{WHU Building\cite{ji2018fully}}& Aerial & Satellite \uppercase\expandafter{\romannumeral2} & A$\thinspace$2$\thinspace$S & 4736 & 3726 &\multirow{3}{*}{512$\thinspace\times\thinspace$512} &1 \\
    &&  Satellite \uppercase\expandafter{\romannumeral2} &Aerial &S$\thinspace$2$\thinspace$A &13662 &1036& & 1 \\
    
    &(Building Detection) & Aerial &Satellite \uppercase\expandafter{\romannumeral1}$^\bullet$ & A2S-\uppercase\expandafter{\romannumeral1}& 4736 & 204 &  & 1\\
    \midrule
    \multirow{12}{*}{Unseen Region and Climate}&\multirow{12}{*}{CASID \cite{liu2023large}} & \multirow{3}{*}{Subtropical Monsoon} & Temperate Monsoon& Sub$\thinspace$2$\thinspace$Tem& \multirow{3}{*}{4900}&2075 & \multirow{12}{*}{1024$\thinspace\times\thinspace$1024} & 5 \\
    & & & Tropical Monsoon& Sub$\thinspace$2$\thinspace$Tms& &1650 & & 5\\
    & & & Tropical Rainforest& Sub$\thinspace$2$\thinspace$Trf & &1550 & &5 \\
    & &\multirow{3}{*}{Temperate Monsoon} &  Subtropical Monsoon&Tem$\thinspace$2$\thinspace$Sub&\multirow{3}{*}{5025} &2200 & & 5\\
    & & & Tropical Monsoon& Tem$\thinspace$2$\thinspace$Tms& &1650 & &5 \\
    & & & Tropical Rainforest& Tem$\thinspace$2$\thinspace$Trf & &1550 & &5 \\
    & &\multirow{3}{*}{Tropical Monsoon} & Subtropical Monsoon& Tms$\thinspace$2$\thinspace$Sub&\multirow{3}{*}{3400}&2200 &  &5 \\
    & & & Temperate Monsoon& Tms$\thinspace$2$\thinspace$Tem& &2075 & &5 \\
    & & & Tropical Rainforest&Tms$\thinspace$2$\thinspace$Trf & &1550 & &5 \\
    & &\multirow{3}{*}{Tropical Rainforest} & Subtropical Monsoon&Trf$\thinspace$2$\thinspace$Sub &\multirow{3}{*}{3700} &2200 &  &5 \\
    & & & Temperate Monsoon& Trf$\thinspace$2Tem& &2075 & &5 \\
    & & & Tropical Monsoon& Trf$\thinspace$2$\thinspace$Tms & &1650 & &5 \\
    
    \bottomrule
    \end{tabular}}

    \label{benchmark collection}
\end{table*}

%% file: algorithm1.tex
\begin{codefloat}[t]

\caption{Multi-Task Training with Sample Gate.}
\label{algorithm1}
\footnotesize
\begin{algorithmic}[1]
\REQUIRE Input images $X$, segmentation labels $Y$, probability threshold $p$, max iterations $T$, Mask Generator $p_m(\tau_m, B)$, Style Predictor $p_s$, Style Transfer module $p_t$, DINOv2 backbone $f_\phi$, Geospatial Semantic Extractor $f_G$, Injector $f_I$, Mask2Former decoder $f_\theta$, ASPP decoder $f_A$, learnable visual prompt $v$
\FOR{$t=1$ to $T$}
    \STATE Generate $M$ by Equations~(\ref{eq: M1}) and (\ref{eq: M2}).
    \STATE Generate $X_M$ by Equation~(\ref{eq: XM}).
    \STATE Sample gate: $u \sim \text{Bernoulli}(p)$
    \IF{$u=1$}  
        \STATE Generate $X_S$ by Equations~(\ref{eq:epsilon}) and (\ref{eq: XS}).
        \STATE Compute $\hat{Y}_S$ b y Equation~(\ref{seg_predict}).
        \STATE Compute $L_{\text{seg}}$ with $(Y,\hat{Y}_S)$ following \cite{Wei_2024_CVPR}.
    \ELSE  
        \STATE Compute $\hat{Y}$ by Equation~(\ref{seg_predict}).
        \STATE Compute $L_{\text{seg}}$ with $(Y,\hat{Y})$ following \cite{Wei_2024_CVPR}.
    \ENDIF
    \STATE Compute $\hat{X}_M$ by Equation~(\ref{eq: mim}).
    \IF{$u=1$}
        \STATE Compute $\hat{X}_S$ by Equation~(\ref{eq: mim}).
        \STATE $L_{\text{mim}} = \|\hat{X}_M-X\|_\alpha + \|\hat{X}_S-X_S\|_\alpha$, $\alpha=1$ or $2$
        \STATE $L_\Delta = \|\hat{X}_M-\hat{X}_S\|_1$
    \ELSE
        \STATE $L_{\text{mim}} = \|\hat{X}_M-X\|_\alpha$
        \STATE $L_\Delta = 0$
    \ENDIF
    \STATE Total loss: $L = L_{\text{seg}} + L_{\text{mim}} + L_\Delta$
\ENDFOR
\end{algorithmic}
\end{codefloat}

\color{black}

%% file: table3.tex
\begin{table*}[t]
    \centering
    \caption{Performance comparison on Potsdam and Vaihingen benchmarks. The $*$, $\diamondsuit$, and $\clubsuit$ separately represent DA models, RS semantic segmentation VFMs, and semantic segmentation VFMs. \textbf{Bolds} are the best scores and \underline{underlines} are the second ones. Surf: Impervious surfaces. Bldg: Building. Vegt: Low vegetation. Clut: Clutter.}
    \resizebox{\textwidth}{!}{
    \begin{tabular}{cccccccccccccccccc}
    \toprule
    \multirow{4}{*}{Method} & \multirow{4}{*}{Backbone} & \multicolumn{1}{c}{Domain} & \multicolumn{6}{c}{Classes} & \multirow{2}{*}{mIoU (\%)} & \multicolumn{1}{c}{Domain} & \multicolumn{6}{c}{Classes} & \multirow{2}{*}{mIoU (\%)} 
    \\ \cline{3-9} \cline{11-17}
    & & \multirow{2}{*}{Source $\rightarrow$ Unseen}& \multirow{2}{*}{Surf} &\multirow{2}{*}{Bldg}
&\multirow{2}{*}{Vegt}&\multirow{2}{*}{Tree}&\multirow{2}{*}{Car}&\multirow{2}{*}{Clut}& &\multirow{2}{*}{Source $\rightarrow$ Unseen}& \multirow{2}{*}{Surf} &\multirow{2}{*}{Bldg}
&\multirow{2}{*}{Vegt}&\multirow{2}{*}{Tree}&\multirow{2}{*}{Car}&\multirow{2}{*}{Clut}\\
\\ \midrule 
\multirow{1}{*}{DAFormer* \cite{hoyer2022daformer}}&\multirow{1}{*}{MiT-B5 \cite{xie2021segformer}}&  &73.8                  & 82.9                  & 46.1                  & 70.0                  & 45.9                 & 8.0                   & 54.4& &64.0                   & 66.5                  & 54.1                  & 28.2                  & 66.6                 & 6.0                   & 47.6  \\
\multirow{1}{*}{HRDA* \cite{hoyer2022hrda}}&\multirow{1}{*}{MiT-B5 \cite{xie2021segformer}}& &75.0                  & 78.3                  & 43.3                  & 68.3                  & 50.8                 & 12.8                  & 54.7 & &69.2                  & 70.1                  & 55.6                  & \textbf{38.9}         & 75.6                 & \textbf{10.6}         & 53.3 \\

\multirow{1}{*}{S12-MoCo$^\diamondsuit$ \cite{stewart2023ssl4eo}}&\multirow{1}{*}{ViT-S \cite{dosovitskiy2020image}}& & 14.7&34.8&22.7&39.2&0.5&1.7&18.9& &46.0&40.2&37.9&20.6&21.6&1.3&27.9 \\

\multirow{1}{*}{S12-DINO$^\diamondsuit$ \cite{stewart2023ssl4eo}}&\multirow{1}{*}{ViT-S \cite{dosovitskiy2020image}}& & 19.1&30.5&28.9&24.0&4.5&3.4&18.4& &33.8&35.7&35.4&20.5&10.7&1.5&22.9 \\

\multirow{1}{*}{S12-MAE$^\diamondsuit$ \cite{stewart2023ssl4eo}}&\multirow{1}{*}{ViT-S \cite{dosovitskiy2020image}}& P(i)2V& 14.6&32.6&23.0&40.8&0.3&1.7&18.4& V2P(i) &44.3&42.9&40.2&21.3&15.4&2.7&27.8 \\

\multirow{1}{*}{DOFA$^\diamondsuit$ \cite{xiong2024neural}}&\multirow{1}{*}{ViT-B \cite{dosovitskiy2020image}}& (Unseen Region)& 32.2&40.8&36.3&36.7&5.2&0.6&25.3& (Unseen Region)&43.4&37.5&35.2&23.0&26.1&2.1&27.9 \\

\multirow{1}{*}{SatMAE$^\diamondsuit$ \cite{cong2022satmae}}&\multirow{1}{*}{ViT-L \cite{dosovitskiy2020image}}& & 42.8&50.7&19.6&54.3&14.4&1.8&30.6& &55.7&46.6&43.7&9.8&29.2&2.5&31.3\\

\multirow{1}{*}{ScaleMAE$^\diamondsuit$ \cite{reed2023scale}}&\multirow{1}{*}{ViT-L \cite{dosovitskiy2020image}}& & 45.9 & 60.7 & 37.9& 55.1& 9.1& 4.4 &35.5 & &59.4 & 54.0 & 48.0& 8.1 & 39.9 & 2.7 &35.2\\

\multirow{1}{*}{RemoteCLIP$^\diamondsuit$ \cite{liu2024remoteclip}}&\multirow{1}{*}{ViT-L \cite{dosovitskiy2020image}}& & 4.0 & 31.7 & 14.0 & 35.7 & 0.1 & 0.3 & 14.3 &  & 38.2 & 35.5 & 36.2 & 23.8 & 9.0 & 2.9 & 24.3\\
\multirow{1}{*}{MTP$^\diamondsuit$ \cite{wang2024mtp}}&\multirow{1}{*}{ViT-L \cite{dosovitskiy2020image}}& &75.0                  & 84.5                  & 51.7                  & \underline{70.8}        & \textbf{65.5}        & \underline{25.3}        & \underline{62.6} & &70.3                  & 76.9                  & 50.2                  & 8.2                   & \underline{82.5}       & 1.9                   & 48.3 \\
\multirow{1}{*}{Rein$^\clubsuit$ (Baseline) \cite{Wei_2024_CVPR}}&\multirow{1}{*}{ViT-L \cite{dosovitskiy2020image}}&  &\underline{79.4}        & \underline{90.4}        & \underline{54.0}        & \textbf{71.5}         & 53.6                 & 13.0                  & 60.3 &  &\underline{75.9}        & \textbf{86.5}         & \underline{60.6}        & \underline{37.9}        & 80.6                 & 4.3                   & \underline{57.6}  \\
\multirow{1}{*}{CrossEarth (Ours)}&\multirow{1}{*}{ViT-L \cite{dosovitskiy2020image}}&& \textbf{83.6}         & \textbf{91.7}         & \textbf{63.8}         & 70.2                  & 59.5                 & \textbf{36.3}         & \textbf{67.5 (+7.2)} & & \textbf{77.1}         & \underline{80.8}        & \textbf{61.3}         & \textbf{38.9}         & \textbf{82.9}        & \underline{8.4}         & \textbf{58.2 (+0.6)} \\
\midrule
\multirow{1}{*}{DAFormer* \cite{hoyer2022daformer}}&\multirow{1}{*}{MiT-B5 \cite{xie2021segformer}}&  & 83.5                  & 88.3                  & 65.3                  & 73.1                  & \underline{91.3}       & 31.0                  & 72.1 &  & 82.1                  & 91.6                  & 62.2                  & 70.0                  & \underline{91.0}       & 24.3                  & 70.2 \\
\multirow{1}{*}{HRDA* \cite{hoyer2022hrda}}&\multirow{1}{*}{MiT-B5 \cite{xie2021segformer}}& &82.9                  & 88.0                  & 57.5                  & 74.1                  & 91.1                 & 23.6                  & 69.5 & &80.0                  & 91.1                  & 62.2                  & 71.1                  & 89.7                 & 25.8                  & 70.0  \\

\multirow{1}{*}{S12-MoCo$^\diamondsuit$ \cite{stewart2023ssl4eo}}&\multirow{1}{*}{ViT-S \cite{dosovitskiy2020image}}& & 14.2&23.8&2.6&0.0&2.3&2.7&7.6& &27.5&37.9&5.3&6.4&15.2&0.5&15.5\\

\multirow{1}{*}{S12-DINO$^\diamondsuit$ \cite{stewart2023ssl4eo}}&\multirow{1}{*}{ViT-S \cite{dosovitskiy2020image}}& & 42.1&28.1&1.3&2.0&33.1&0.5&17.9& &37.8&50.3&22.8&16.2&32.4&1.2&26.8\\

\multirow{1}{*}{S12-MAE$^\diamondsuit$ \cite{stewart2023ssl4eo}}&\multirow{1}{*}{ViT-S \cite{dosovitskiy2020image}}& P(r)2P(i)& 56.3&30.5&2.0&1.3&49.5&0.6&23.4& P(i)2P(r) &29.7&43.1&14.1&9.4&2.0&0.6&16.5 \\

\multirow{1}{*}{DOFA$^\diamondsuit$ \cite{xiong2024neural}}&\multirow{1}{*}{ViT-B \cite{dosovitskiy2020image}}& (Unseen Spectral Band) &54.5&32.6&10.4&6.0&43.3&1.0&24.6& (Unseen Spectral Band) &46.7&48.8&30.4&10.3&42.1&1.1&29.9\\

\multirow{1}{*}{SatMAE$^\diamondsuit$ \cite{cong2022satmae}}&\multirow{1}{*}{ViT-L \cite{dosovitskiy2020image}}& &56.4&41.4&20.4&24.8&59.2&8.9&35.2& &63.1& 81.5 & 37.2 & 31.0 & 75.4 & 10.4 & 49.8 \\

\multirow{1}{*}{ScaleMAE$^\diamondsuit$ \cite{reed2023scale}}&\multirow{1}{*}{ViT-L \cite{dosovitskiy2020image}}& & 74.9 & 53.3 & 5.0 & 59.7& 74.2& 12.5 &46.6 & &47.6 & 77.4 & 5.3& 45.7 & 67.7 & 7.3 &41.8 \\

\multirow{1}{*}{RemoteCLIP$^\diamondsuit$ \cite{liu2024remoteclip}}&\multirow{1}{*}{ViT-L \cite{dosovitskiy2020image}}& & 47.8 & 23.9 & 3.5 & 0.5 & 2.6 & 0.1 & 13.1  & & 11.1 & 28.9 & 2.1 & 0.7 & 1.1 & 0.1 & 7.3\\
\multirow{1}{*}{MTP$^\diamondsuit$ \cite{wang2024mtp}}&\multirow{1}{*}{ViT-L \cite{dosovitskiy2020image}}& &83.4                  & 81.3                  & 42.2                  & 68.0                  & \underline{91.3}       & 20.6                  & 64.5 & &78.6                  & 91.3                  & \underline{62.5}        & \underline{71.3}        & \underline{91.0}       & \underline{31.2}        & \underline{71.0}  \\
\multirow{1}{*}{Rein$^\clubsuit$ (Baseline) \cite{Wei_2024_CVPR}}&\multirow{1}{*}{ViT-L \cite{dosovitskiy2020image}}&  &\textbf{86.6}         & \textbf{93.4}         & \underline{73.1}        & \underline{77.7}        & \underline{91.3}       & \underline{35.8}        & \underline{76.3}  & &\underline{82.2}        & \underline{93.2}        & 58.5                  & 70.0                  & 87.8                 & 23.7                  & 69.2  \\
\multirow{1}{*}{CrossEarth (Ours)}&\multirow{1}{*}{ViT-L\cite{dosovitskiy2020image}}& & \underline{86.1}        & \underline{91.5}        & \textbf{73.8}         & \textbf{79.7}         & \textbf{91.9}        & \textbf{43.8}         & \textbf{77.8 (+1.5)} & &\textbf{86.3}         & \textbf{93.5}         & \textbf{73.8}         & \textbf{74.0}         & \textbf{91.2}        & \textbf{37.2}         & \textbf{76.0 (+6.8)}\\
\midrule
\multirow{1}{*}{DAFormer* \cite{hoyer2022daformer}}&\multirow{1}{*}{MiT-B5 \cite{xie2021segformer}}&  & 64.2                  & 73.4                  & 4.5                   & 9.7                   & 42.2                 & 1.5                   & 32.6 & & 46.0                  & 59.4                  & 12.6                  & 5.8                   & 63.5                 & 2.9                   & 31.7  \\
\multirow{1}{*}{HRDA* \cite{hoyer2022hrda}}&\multirow{1}{*}{MiT-B5 \cite{xie2021segformer}}& &  67.1                  & 66.9                  & 4.1                   & 17.5                  & 43.0                 & 1.8                   & 33.4& & 54.7                  & 54.7                  & 11.6                  & \underline{14.4}        & 72.3                 & \underline{6.5}         & 35.7  \\

\multirow{1}{*}{S12-MoCo$^\diamondsuit$ \cite{stewart2023ssl4eo} }&\multirow{1}{*}{ViT-S \cite{dosovitskiy2020image}}& & 11.5&25.2&1.8&0.0&2.6&1.7&7.1& &42.7&27.2&4.3&2.2&20.8&1.2&16.4\\

\multirow{1}{*}{S12-DINO$^\diamondsuit$ \cite{stewart2023ssl4eo}}&\multirow{1}{*}{ViT-S \cite{dosovitskiy2020image}}& & 10.0&24.4&1.5&0.0&4.7&1.4&7.0& &28.8&25.1&3.9&1.3&12.5&1.0&12.1\\

\multirow{1}{*}{S12-MAE$^\diamondsuit$ \cite{stewart2023ssl4eo}}&\multirow{1}{*}{ViT-S \cite{dosovitskiy2020image}}& P(r)2V& 9.0&25.5&1.3&0.0&5.7&1.1&7.1& V2P(r)&40.9&29.9&9.1&2.0&21.1&1.4&33.7 \\

\multirow{1}{*}{DOFA$^\diamondsuit$ \cite{xiong2024neural}}&\multirow{1}{*}{ViT-B \cite{dosovitskiy2020image}}& (Unseen Region and Spectral Band) & 27.2&22.3&2.4&0.0&4.7&1.1&9.6 & (Unseen Region and Spectral Band) &38.4&24.2&7.6&3.0&23.9&0.8&16.3\\

\multirow{1}{*}{SatMAE$^\diamondsuit$ \cite{cong2022satmae}}&\multirow{1}{*}{ViT-L \cite{dosovitskiy2020image}}& & 53.6 & 26.4 & 3.5& 0.6& 20.1&8.2&18.7 & &45.0 &38.0&20.9&4.2&29.4&2.4&23.3\\

\multirow{1}{*}{ScaleMAE$^\diamondsuit$ \cite{reed2023scale}}&\multirow{1}{*}{ViT-L \cite{dosovitskiy2020image}}& & 43.3 & 30.9 & 1.1& 1.8& 7.4& 8.1 & 15.4 & &43.9 & 51.9& 25.2 & 0.8 & 35.2 &3.2&26.7\\

\multirow{1}{*}{RemoteCLIP$^\diamondsuit$ \cite{liu2024remoteclip}}&\multirow{1}{*}{ViT-L \cite{dosovitskiy2020image}}& &  4.0 & 23.2 & 3.1 & 0.0 & 0.0 & 0.1 & 5.1  & & 33.1 & 25.5 & 2.1 & 1.6 & 6.9 & 1.6 & 11.8\\
\multirow{1}{*}{MTP$^\diamondsuit$ \cite{wang2024mtp}}&\multirow{1}{*}{ViT-L \cite{dosovitskiy2020image}}& &51.0                  & 64.8                  & 4.4                   & 7.5                   & 56.8                 & 1.9                   & 31.1 & &58.5                  & \underline{76.5}        & 44.4                  & 10.8                  & \textbf{82.0}        & 1.4                   & 48.6 \\
\multirow{1}{*}{Rein$^\clubsuit$ (Baseline) \cite{Wei_2024_CVPR}}&\multirow{1}{*}{ViT-L \cite{dosovitskiy2020image}}&   & \textbf{79.1}         & \textbf{91.2}         & \underline{37.0}        & \underline{62.2}        & \underline{61.1}       & \underline{5.7}         & \underline{56.1}&  &\underline{70.4}        & \textbf{77.4}         & \underline{58.6}        & 13.9                  & 78.5                 & 4.4                   & \underline{50.6}  \\
\multirow{1}{*}{CrossEarth (Ours)}&\multirow{1}{*}{ViT-L \cite{dosovitskiy2020image}}& &\underline{78.1}        & \underline{89.3}        & \textbf{55.2}         & \textbf{72.5}         & \textbf{61.9}        & \textbf{14.5}         & \textbf{61.9 (+5.8)} & &\textbf{72.5}         & 73.6                  & \textbf{58.7}         & \textbf{22.4}         & \underline{81.1}       & \textbf{7.6}          & \textbf{52.7 (+2.1)} \\ 
\bottomrule
    \end{tabular}}
    \label{p2v}
\end{table*}

%% file: table4.tex
\begin{table}[t!]
    \centering
    \caption{Performance comparison on LoveDA-Rural to Urban and Urban to Rural benchmarks (testing on official validation sets). Bkgd: Background. Bldg: Building. Rd: Road. Wtr: Water. Barr: Barren. Frst: Forest. Agri: Agriculture.}
     \resizebox{1.0\linewidth}{!}{
    \begin{tabular}{c|ccccccc|c}
    \toprule
        Models& Bkgd &Bldg &Rd &Wtr &Barr &Frst &Agri& mIoU (\%) \\
       \midrule
       \multicolumn{9}{c}{Performance Comparison on LoveDA-Rural to Urban }\\
       \midrule
          DAFormer \cite{hoyer2022daformer} & 40.1&55.2 &51.7 &69.9 &43.3 &51.9 &49.0 &51.6 \\
          HRDA \cite{hoyer2022hrda} & \textbf{41.6} & 57.1& 53.1&63.2 &45.6 &51.8 &56.0 &52.6\\
          
          S12-MoCo \cite{stewart2023ssl4eo}& 33.0 & 7.8 & 1.4 & 21.0 & 0.0 & 29.3 & 30.9 & 17.6\\
          
          S12-DINO \cite{stewart2023ssl4eo}& 33.1 & 11.0 & 2.5 & 48.8 & 0.3 & 23.2 & 34.7 & 21.9\\
          
          S12-MAE \cite{stewart2023ssl4eo}& 33.2 & 9.1 & 1.8 & 48.2 & 0.0 & 14.8 & 34.5 & 20.1\\
          
          DOFA \cite{xiong2024neural}& 27.6 & 36.7 & 29.4 & 30.5 & 12.8& 28.3 & 2.3 & 23.9\\
          
          SatMAE \cite{cong2022satmae}& 36.5 & 42.3 & 43.7 & 57.1 & 28.5 & 38.6 & 40.5 & 41.0\\
          
          ScaleMAE  \cite{reed2023scale}& 39.7 & 54.1 & 54.7 & 71.6 & 45.9 &47.1 & 52.7 & 52.3\\
          
          RemoteCLIP \cite{liu2024remoteclip}& 24.4 & 3.7 & 0.7 & 12.4 & 0.0 & 19.7 & 26.7 & 12.5 \\
          MTP \cite{wang2024mtp} & \underline{40.8}&59.6 &\textbf{58.3} &74.4 &46.6&47.4 &54.6&54.5\\
          Rein (Baseline) \cite{Wei_2024_CVPR} &40.3 &\textbf{64.0} &56.6 &\textbf{76.0} &\underline{50.4}&\textbf{55.4} &\underline{61.1} &\underline{57.7}\\
          CrossEarth (Ours) & 39.8& \underline{63.3}& \underline{57.3} & \underline{75.9} &\textbf{51.8} &\underline{55.0} &\textbf{62.5}&\textbf{57.9 (+0.2)}\\
       \midrule
       \multicolumn{9}{c}{Performance Comparison on LoveDA-Urban to Rural}\\
       \midrule
          DAFormer \cite{hoyer2022daformer} & \underline{57.1}& 46.9 &36.5 &62.9 &\underline{12.1}&18.5&51.3&40.8 \\
          HRDA \cite{hoyer2022hrda} & 50.2 & 46.1& 40.0&66.6 &6.8 &26.9 &\underline{58.1}&42.1\\
          
          S12-MoCo \cite{stewart2023ssl4eo}& 42.5 & 14.7 &16.9 & 11.6 & 7.0 & 1.8 & 0.3 &23.9\\
          
          S12-DINO \cite{stewart2023ssl4eo}& 44.8 & 21.0 & 24.9 &15.8 &7.0 &0.8 &0.3& 16.8\\
          
          S12-MAE \cite{stewart2023ssl4eo}&42.0 & 17.7 & 7.1 & 11.6 & 7.6 &4.2 & 0.1 & 12.9\\
          
          DOFA \cite{xiong2024neural}& 27.6 & 36.7&29.4&30.5&\textbf{12.8}&28.3&2.3 &13.2\\
          
          SatMAE \cite{cong2022satmae}& 52.1 & 43.4 & 40.3 & 53.9 & 7.0 & \underline{31.5} & 34.1 & 37.5 \\
          
          ScaleMAE  \cite{reed2023scale}& 54.1 & 52.8 & 38.8 & 52.3 & \textbf{12.8} & 31.0 & 42.2 & 40.6\\
          
          RemoteCLIP \cite{liu2024remoteclip}& 40.7 & 13.4 & 9.5 & 5.2 & 3.4 & 1.5 & 0.0 & 10.5 \\
          MTP \cite{wang2024mtp} &55.5&44.4 &46.3 &\underline{66.7} &7.3&\textbf{37.0} &50.7&44.0\\
          Rein (Baseline) \cite{Wei_2024_CVPR} &57.0 &\underline{56.3} &\textbf{49.4} &\textbf{68.2} &9.3&25.5 &53.5 & \underline{45.6}\\
          CrossEarth (Ours) & \textbf{57.5} &\textbf{61.5} & \underline{48.5} & 65.6 &9.6 & 25.4 & \textbf{58.2} &\textbf{46.6 (+1.0)}\\
          \midrule
          
       \multicolumn{9}{c}{Performance Comparison on LoveDA-Urban to OpenEarthMap}\\
       \midrule
          
          DAFormer \cite{hoyer2022daformer} &\textbf{47.7}& 58.6& 31.3& 31.2& 2.9& \underline{45.2}& 34.9 & 36.0 \\
          
          HRDA \cite{hoyer2022hrda} &  46.1& 53.3& 37.1& 32.5& 1.5& 35.1& 41.9 & 35.4\\
          
          S12-MoCo \cite{stewart2023ssl4eo}& 36.9 &26.7&13.3&9.9&3.3&9.8&1.3 & 14.5 \\
          
          S12-DINO \cite{stewart2023ssl4eo}& 36.4 &31.5&14.7&17.9&2.5&7.4&0.4&15.8 \\
          
          S12-MAE \cite{stewart2023ssl4eo}& 36.1 &27.4&10.8&20.7 &2.6&10.3&0.4&15.4\\
          
          DOFA \cite{xiong2024neural}& 35.2 &32.7&14.2&24.8&3.8&8.2&1.9&17.3\\
          
          SatMAE \cite{cong2022satmae}&  38.7 & 43.4 & 23.0 & 25.1 & \underline{6.4} & 10.0 & 10.7 & 23.7 \\
          
          ScaleMAE  \cite{reed2023scale}& 40.5 & 44.6 & 31.6 & 21.3 & 0.7 & 28.1 & 15.5 & 26.0 \\
          
          RemoteCLIP \cite{liu2024remoteclip}&  40.2 & 9.9 & 7.0 & 5.2 & 2.3 & 8.6 & 0.0 & 10.5\\
          
          MTP \cite{wang2024mtp} & 40.2 & 57.8 & 36.4 & 55.4 & 2.7 & \textbf{46.2} & 38.0 & 39.5\\
          
          Rein (Baseline) \cite{Wei_2024_CVPR} & 45.2& \underline{63.4} & \underline{38.1} & \underline{62.7} & 4.2 & 42.5 & \underline{65.2} & \underline{45.9}\\
          
          CrossEarth (Ours) & \underline{47.6} & \textbf{65.1} & \textbf{39.9} & \textbf{69.9} &\textbf{13.2 }&42.4&\textbf{66.3} &\textbf{49.2 (+3.3)}\\
        \bottomrule
    \end{tabular}}
    \label{loveda}
\end{table}

%% file: table11.tex
\begin{table}[t!]
    \centering
    \caption{Performance comparison on building extraction and road extraction tasks across A2S, A2S-\uppercase\expandafter{\romannumeral1}, S2A, D2M, and D2G benchmarks using mIoU scores (\%).}
     \resizebox{\linewidth}{!}{
    \begin{tabular}{c|cccccc}
    \toprule
        \multirow{2}{*}{Models} & \multicolumn{3}{c}{Building} & \multicolumn{2}{c}{Road} \\
        \cmidrule{2-4} \cmidrule{5-6}
        & A2S & A2S-\uppercase\expandafter{\romannumeral1} & S2A & D2M & D2G  \\
       \midrule
          DAFormer \cite{hoyer2022daformer} & 7.1 & 0.6&64.3& 41.9& 18.3 \\
          HRDA \cite{hoyer2022hrda} & 7.9 &4.9 & 62.7& 50.3 & 20.1\\
          
          S12-MoCo \cite{stewart2023ssl4eo}& 7.7 & 20.1 & 20.9 & 7.8 & 11.2\\
          
          S12-DINO \cite{stewart2023ssl4eo}& 6.8 & 17.1 & 19.9 & 5.9 & 9.3\\
          
          S12-MAE \cite{stewart2023ssl4eo}& 5.4 & 22.3 & 19.3 & 7.1 & 11.5\\
          
          DOFA \cite{xiong2024neural}& 5.7 & 23.9 & 19.6 & 8.4 & 12.2\\
          
          SatMAE \cite{cong2022satmae}& 12.9 & 32.8 & 18.1 &23.3& 20.2\\
          
          ScaleMAE \cite{reed2023scale}& 0.6& 38.0 & 19.4 &32.1 & 23.4\\
          
          RemoteCLIP \cite{liu2024remoteclip}& 4.8 & 20.9 & 18.0 & 5.9 & 6.8\\
          MTP \cite{wang2024mtp} & 23.6& 46.0 & 47.2 &\textbf{54.3} & 22.6\\
          Rein (Baseline) \cite{Wei_2024_CVPR} &\underline{30.8} & \underline{48.0} & \textbf{66.8} & 49.7 & \underline{23.0}\\
          CrossEarth (Ours) & \textbf{44.3 (+13.5)} &\textbf{57.1 (+9.1)} &\underline{64.5 (-2.3)}& \underline{50.5 (+0.8)} & \textbf{24.2 (+1.2)}\\
        \bottomrule
    \end{tabular}}
    \label{road_building_table}
\end{table}

%% file: table5.tex
\begin{table*}[t]
    \caption{Performance comparison on P(r)2Res and P(i)2Res benchmarks. The gray blocks highlight significant improvements of CrossEarth compared to other methods. Surf: Impervious surfaces. Bldg: Building. Clut: Clutter}
    \centering
    \resizebox{\textwidth}{!}{
    \begin{tabular}{cccccccccccccccc}
    \toprule
    \multirow{4}{*}{Method} & \multirow{4}{*}{Backbone} & \multicolumn{1}{c}{Domain} & \multicolumn{5}{c}{Classes} & \multirow{2}{*}{mIoU (\%)} & \multicolumn{1}{c}{Domain} & \multicolumn{5}{c}{Classes} & \multirow{2}{*}{mIoU (\%)} 
    \\ \cline{3-8} \cline{10-15}
    & & \multirow{2}{*}{Source $\rightarrow$ Unseen}& \multirow{2}{*}{Surf} &\multirow{2}{*}{Bldg}
&\multirow{2}{*}{Tree}&\multirow{2}{*}{Car}&\multirow{2}{*}{Clut}& &\multirow{2}{*}{Source $\rightarrow$ Unseen}& \multirow{2}{*}{Surf} &\multirow{2}{*}{Bldg}
&\multirow{2}{*}{Tree}&\multirow{2}{*}{Car}&\multirow{2}{*}{Clut}\\
\\ \midrule \multicolumn{16}{c}{Performance Comparison in \textbf{Cross-Domain Generalization} setting on RescueNet benchmark}  \\ \midrule
\multirow{1}{*}{DeepLabv3 \cite{chen2017rethinking}}&\multirow{1}{*}{ResNet-101 \cite{he2016deep}}&  & 26.9 & 23.3 & 1.3&2.5 &53.3 & 21.5& & 12.8& 9.7 & 40.4 &9.4 & 17.6&18.0\\
\multirow{1}{*}{DAFormer \cite{hoyer2022daformer}}&\multirow{1}{*}{MiT-B5 \cite{xie2021segformer}} & & {33.0} & {32.6} &\underline{55.0} &7.6 &\underline{66.7}& {39.0} & &34.1 &43.6 &41.7 &12.0 &58.9 & 38.1\\
\multirow{1}{*}{HRDA \cite{hoyer2022hrda}} & \multirow{1}{*}{MiT-B5 \cite{xie2021segformer}} & & 28.8 &27.2 &\textbf{56.8} &4.0&66.4&36.6 & &26.1&29.2&44.3&5.2&61.6&33.3\\

S12-MoCo \cite{stewart2023ssl4eo}& ViT-S \cite{dosovitskiy2020image} & & 7.8 & 9.4 & 5.0 & 1.5 & 0.6 & 4.9 & &2.9 & 9.1 & 2.1 & 1.1 & 0.3 &3.1 \\

S12-DINO \cite{stewart2023ssl4eo}& ViT-S \cite{dosovitskiy2020image}& P(r)2Res& 6.8 & 9.8 & 18.8 & 1.0 & 0.2 & 7.3 & P(i)2Res& 0.8 & 10.2 & 22.2 & 2.1 & 1.4 & 7.3\\

S12-MAE \cite{stewart2023ssl4eo}& ViT-S \cite{dosovitskiy2020image}& & 2.5 &9.3 & 6.8 & 0.4 & 0.2 & 3.8 & &5.6 &10.1 &10.3 &2.5 &0.9 & 5.9\\

\multirow{1}{*}{DOFA} \cite{xiong2024neural}& \multirow{1}{*}{ViT-B \cite{dosovitskiy2020image}} &  (Unseen Region)&5.0 & 9.0 & 3.5 & 1.3 & 0.3 &3.8 &(Unseen Region and Spectral
Band) &2.6 & 8.9& 0.8& 1.0 & 0.6 & 2.8 \\

\multirow{1}{*}{SatMAE} \cite{cong2022satmae}& \multirow{1}{*}{ViT-L \cite{dosovitskiy2020image}} & &13.7 & 13.8 & 26.1 & 3.7 & 27.1 & 16.9 & & 15.5 & 13.7 &24.4& 3.0&21.1& 15.5 \\

\multirow{1}{*}{ScaleMAE} \cite{reed2023scale}& \multirow{1}{*}{ViT-L \cite{dosovitskiy2020image}} & & 23.5 & 31.9 & 45.4 & 0.1 & 47.7 & 29.7 & & 13.9 & 35.0 & \underline{48.7} & 0.3 & 49.4 & 29.4\\

RemoteCLIP \cite{liu2024remoteclip}& \multirow{1}{*}{ViT-L \cite{dosovitskiy2020image}} & & 2.2 & 8.2 & 2.6 & 0.0 & 0.0 & 2.6& & 1.3 & 10.1 & 27.5 & 0.2 &0.8& 8.0 \\ 
\multirow{1}{*}{MTP \cite{wang2024mtp}} & \multirow{1}{*}{ViT-L \cite{dosovitskiy2020image}} &  &34.3 &40.9&41.7&\underline{16.3}&62.7&39.2& & 31.3&36.8&0.5&\textbf{14.4}&57.5&28.1 \\
\multirow{1}{*}{Rein (Baseline) \cite{Wei_2024_CVPR}} & \multirow{1}{*}{ViT-L \cite{dosovitskiy2020image}} & &\textbf{53.6}&\underline{49.5}&44.2&13.7&\textbf{68.6}&\underline{45.9}&&\textbf{51.3}&\underline{47.3}&48.5&\underline{13.3}&\textbf{66.5}&\underline{45.4}\\
\multirow{1}{*}{CrossEarth (Ours)}&\multirow{1}{*}{ViT-L \cite{dosovitskiy2020image}}& & \underline{43.6}& \cellcolor[HTML]{C0C0C0}\textbf{59.7 (+10.2)}& {51.0} & \textbf{16.4}& {63.2} &\textbf{46.8 (+0.9)} & &\underline{41.9} &\cellcolor[HTML]{C0C0C0}\textbf{60.6 (+13.3)} & \textbf{51.3}& 10.6&\underline{62.9} &\textbf{45.5 (+0.1)} \\

\bottomrule
    \end{tabular}}

    \label{Rescue}
\end{table*}

%% file: table6-1.tex
\begin{table*}[t]
\caption{Performance comparison in CASID benchmarks. The comparison consists of 4 in-domain and 12 out-of-domain experiments between various specialized models and VFMs. Sub: Subtropical Monsoon. Tem: Temperate Monsoon. Tms: Tropical Monsoon. Trf: Tropical Rainforest.}
    \centering
    \resizebox{\textwidth}{!}{
\begin{tabular}{cccccccccc}
\toprule
\multirow{2}{*}{Method} & \multirow{2}{*}{Backbone} & \multicolumn{8}{c}{Domain(Source $\rightarrow$ Unseen) ------ mIoU(\%)} \\ \cmidrule{3-10} 
 &  & Sub (Source-Only) & Sub2Tem & \multicolumn{1}{l}{Sub2Tms} & \multicolumn{1}{l}{Sub2Trf} & \multicolumn{1}{l}{Tem (Source-Only)} & \multicolumn{1}{l}{Tem2Sub} & \multicolumn{1}{l}{Tem2Tms} & \multicolumn{1}{l}{Tem2Trf} \\ \midrule
DeepLabv2 \cite{chen2017deeplab} & ResNet-101 \cite{he2016deep} & 60.9 & 10.0 & 12.8 & 15.3 & 39.9 & 9.4 & 10.9 & 12.2 \\
DAFormer \cite{hoyer2022daformer} & MiT-B5 \cite{xie2021segformer} & \underline{68.0} & 36.0 & \underline{63.5} & 59.0 & 45.0 & 59.8 & \underline{63.8} & 51.7 \\
HRDA \cite{hoyer2022hrda} & MiT-B5 \cite{xie2021segformer} & 66.3 & 34.6 & 63.0 & 56.2 & 46.3 & 62.7 & 63.3 & 44.9 \\
 S12-MoCo \cite{stewart2023ssl4eo} & ViT-S \cite{dosovitskiy2020image} & 40.3 & 19.6 & 29.3 & 32.1 & 18.3 & 21.5 & 17.1 & 12.2 \\
 S12-DINO \cite{stewart2023ssl4eo} & ViT-S \cite{dosovitskiy2020image} & 27.6 & 18.7 & 21.3 & 26.2 & 22.4 & 28.1 & 21.9 & 24.7 \\
 S12-MAE \cite{stewart2023ssl4eo} & ViT-S \cite{dosovitskiy2020image} & 28.2 & 18.2 & 19.9 & 26.3 & 21.2 & 25.4 & 22.4 & 24.9 \\
 DOFA \cite{xiong2024neural} & ViT-B \cite{dosovitskiy2020image} & 33.6 & 18.5 & 25.3 & 30.1 & 21.2 & 26.4 & 22.9 & 20.8 \\
 SatMAE \cite{cong2022satmae} & ViT-L \cite{dosovitskiy2020image} & 60.8 & 33.1 & 52.9 & 48.7 & 36.2 & 51.8 & 43.7 & 36.2 \\
 ScaleMAE \cite{reed2023scale} & ViT-L \cite{dosovitskiy2020image} & 66.3 & 39.2 & 60.7 & 56.4 & 50.4 & 62.5 & 57.3 & 45.9 \\
 RemoteCLIP \cite{liu2024remoteclip} & ViT-L \cite{dosovitskiy2020image} & 16.6 & 8.5 & 15.1 & 19.7 & 18.0 & 54.3 & 57.8 & 16.7 \\
MTP \cite{wang2024mtp} & ViT-L \cite{dosovitskiy2020image} & \underline{68.0} & 39.4 & 60.0 & 61.6 & 42.9 & \textbf{66.2} & \textbf{64.7} & 48.3 \\
Rein (Baseline) \cite{Wei_2024_CVPR} & ViT-L \cite{dosovitskiy2020image} & 67.2 & \underline{46.4} & 59.7 & \underline{63.4} & \underline{51.3} & \underline{65.5} & 62.9 & \underline{55.9} \\
CrossEarth (Ours) & ViT-L \cite{dosovitskiy2020image} & \textbf{69.4 (+2.2)} & \textbf{48.1 (+1.7)} & \textbf{64.6 (+4.9)} & \textbf{64.2 (+0.8)} & \textbf{53.5 (+2.2)} & 63.5 (-2.0)  & 61.8 (-1.1) & \textbf{57.8 (+1.9)} \\ 
\midrule
\multirow{2}{*}{Method} & \multirow{2}{*}{Backbone} & \multicolumn{8}{c}{Domain(Source $\rightarrow$ Unseen) ------ mIoU(\%)} \\ \cmidrule{3-10} 
 &  & Tms (Source-Only) & Tms2Sub & \multicolumn{1}{l}{Tms2Tem} & \multicolumn{1}{l}{Tms2Trf} & \multicolumn{1}{l}{Trf (Source-Only)} & \multicolumn{1}{l}{Trf2Sub} & \multicolumn{1}{l}{Trf2Tem} & \multicolumn{1}{l}{Trf2Tms} \\ \midrule
DeepLabv2 \cite{chen2017deeplab} & ResNet-101 \cite{he2016deep} & 56.5 & 12.8 & 12.2 & 9.1 & 49.0 & 10.2 & 8.5 & 10.9 \\
 DAFormer \cite{hoyer2022daformer} & MiT-B5 \cite{xie2021segformer} & 64.1 & 61.8 & 31.4 & 56.2 & \underline{64.6} & 62.9 & 39.6 & 62.7 \\
HRDA \cite{hoyer2022hrda} & MiT-B5 \cite{xie2021segformer} & 62.6 & 63.8 & 33.3 & 56.9 & 63.5 & 63.3 & \textbf{43.0} & 63.1\\
 S12-MoCo \cite{stewart2023ssl4eo} & ViT-S \cite{dosovitskiy2020image} & 21.1 & 28.8 & 21.6 & 27.3 & 28.9 & 33.0 & 18.7 & 26.3 \\
 S12-DINO \cite{stewart2023ssl4eo} & ViT-S \cite{dosovitskiy2020image} & 17.5 & 22.9 & 18.0 & 23.0 & 31.5 & 35.0 & 17.3 & 21.5 \\
 S12-MAE \cite{stewart2023ssl4eo} & ViT-S \cite{dosovitskiy2020image} & 17.5 & 26.6 & 18.9 & 27.6 & 32.5 & 34.0 & 14.5 & 19.5 \\
 DOFA \cite{xiong2024neural} & ViT-B \cite{dosovitskiy2020image} & 24.1 & 29.8 & 18.8 & 29.1 & 31.2 & 36.1 & 17.1 & 35.7 \\
 SatMAE \cite{cong2022satmae} & ViT-L \cite{dosovitskiy2020image} & 57.4 & 60.2 & 32.5 & 53.0 & 57.8 & 59.1 & 32.8 & 56.7 \\
 ScaleMAE \cite{reed2023scale} & ViT-L \cite{dosovitskiy2020image} & 61.1 & 62.1 & 35.2 & 55.4 & 59.5 & 60.6 & 30.9 & 58.0 \\
 RemoteCLIP \cite{liu2024remoteclip} & ViT-L \cite{dosovitskiy2020image} & 18.7 & 22.5 & 12.3 & 22.5 & 26.4 & 32.1 & 26.0 & 23.6 \\
MTP \cite{wang2024mtp} & ViT-L \cite{dosovitskiy2020image} & 62.5 & 56.1 & 32.3 & \underline{60.1} & 62.8 & 55.6 & 37.0 & 59.8 \\
Rein (Baseline) \cite{Wei_2024_CVPR} & ViT-L \cite{dosovitskiy2020image} & \underline{65.0} & \underline{68.6} & \underline{37.2} & 59.7 & 61.0 & \underline{67.6} & 37.8 & \textbf{64.9} \\
CrossEarth (Ours) & ViT-L \cite{dosovitskiy2020image} & \textbf{68.9 (+3.9)} & \textbf{69.1 (+0.5)} & \textbf{40.5 (+3.3)} & \textbf{60.7 (+1.0)} & \textbf{65.3 (+4.3)} & \textbf{67.9 (+0.3)} & \underline{42.3 (+4.5)}  & \underline{64.4 (-0.5)}  \\ 
\bottomrule
\label{casid_exp}
\end{tabular}}
\end{table*}

%% file: main_ablation_table.tex
\begin{table*}[t]
\centering
\scriptsize

\caption{Comprehensive ablation study on the CASID dataset.
All models are trained on Subms with identical budgets and evaluated on Subms/Temms/Trms/Trorf (mIoU \%). Here, CA denotes Cross-Attention, MIM dec. denotes the decoder of MIM, and Style refers to the Earth-Style Injection module. The training schedule is fixed across all methods: 30k iterations, batch size = 1, AdamW optimizer, and an initial learning rate of $1 \times 10^{-4}$. For mask generation, the styled images use a masking ratio $\tau_m = 0.1$ with block size $B = 64$, while the masked images use $\tau_m = 0.7$ with $B = 64$. Notably, $L_{MIM}$ is implemented with $\ell_1$ loss on the CASID dataset. Subms: Subtropical Monsoon. Temms: Temperate Monsoon. Troms: Tropical Monsoon. Trorf: Tropical Rainforest.}
\resizebox{0.95\linewidth}{!}{
\begin{tabular}{lccccccccccc}
\toprule
\multirow{2}{*}{Setting} &
\multirow{2}{*}{GSE} &
\multirow{2}{*}{Injector} &
\multirow{2}{*}{MIM dec.} &
\multirow{2}{*}{$L_{MIM}$} &
\multirow{2}{*}{$L_\Delta$} &
\multirow{2}{*}{$p$} &
\multicolumn{5}{c}{Subms ------ mIoU (\%)} \\
\cmidrule(lr){8-12}
& $f_G$ & $f_I$ & $f_A$ &  &  &  &
Subms &Temms &Troms &Trorf & Avg. \\
\midrule
\rowcolor{gray!15} \multicolumn{12}{l}{$\blacktriangledown$ \emph{Effectiveness of each main component (GSE, MIM, Style)}} \\
 R1 \; Rein (Baseline) & -- & -- & -- & -- & -- & -- & 67.2  & 46.4 & 59.7 & 63.4 & 59.2  \\
 R2 \; R1+GSE & \checkmark & CA & -- & -- & -- & -- & 69.0 &  46.1 & 60.2 & 62.8 &59.6 (+0.4) \\
 R3 \; R1+MIM & -- & -- & ASPP & $\ell_1$ & -- & -- & 68.7 &  47.2 & 62.7 & 63.1 &60.4 (+1.2)  \\
 R4 \; R1+Style & -- & -- & -- & -- & -- & 0.1 & 68.4 & 45.6 & 61.5 & 63.7 & 59.8 (+0.6) \\
 R5 \; R2+MIM & \checkmark & CA & ASPP & $\ell_1$ & -- & -- & 69.6 & 47.4 & 63.5 & 63.2 & 60.9 (+1.7) \\
 R6 \; R2+Style & \checkmark & CA & -- & -- & -- & 0.1 & 69.3 & 46.9 & 63.3 & 62.2 & 60.4 (+1.2) \\
 R7 \; R3+Style & -- & -- & ASPP & $\ell_1$ & $\ell_1$ & 0.1 & 69.1 & 47.8 & 64.0 & 64.2 & 61.3 (+2.1) \\
 R8 \; R6+MIM (CrossEarth) & \checkmark & CA & ASPP & $\ell_1$ & $\ell_1$ & 0.1 & 69.4 & 48.1 & 64.6 & 64.2 &\textbf{ 61.6 (+2.4) } \\
\midrule
\rowcolor{gray!15} \multicolumn{12}{l}{$\blacktriangledown$ \emph{Ablation of injector}} \\
 R9 \; R8 ($f_I=$ FFN)  & \checkmark & \textbf{FFN} & ASPP & $\ell_1$ & $\ell_1$ & 0.1 & 68.9 & 47.8 & 63.9 & 64.0 & 61.2 (+2.0)\\
\midrule
\rowcolor{gray!15} \multicolumn{12}{l}{$\blacktriangledown$ \emph{Ablation of decoder in MIM}} \\
 R10\; R8 ($f_A=$ Linear) & \checkmark & CA & \textbf{Linear} & $\ell_1$ & $\ell_1$ & 0.1 & 65.1 & 49.1 & 57.6 & 64.4 & 59.1 (-0.1) \\
\midrule
\rowcolor{gray!15} \multicolumn{12}{l}{$\blacktriangledown$ \emph{Ablation of loss and sample gate in Style}} \\
 R11 \; R8 ($p=0.3$) & \checkmark & CA & ASPP & $\ell_1$ & $\ell_1$ & \textbf{0.3} & 69.0 & 47.5 & 63.8 & 63.5 & 60.9 (+1.7) \\
 R12 \; R8 ($p=0.5$) & \checkmark & CA & ASPP & $\ell_1$ & $\ell_1$ & \textbf{0.5} & 68.6 & 47.0 & 63.2 & 62.9 & 60.4 (+1.2) \\
 R13 \; R8 ($p=0.7$) & \checkmark & CA & ASPP & $\ell_1$ & $\ell_1$ & \textbf{0.7} & 67.9 & 46.3 & 62.5 & 62.4 & 59.8 (+0.6) \\
 R14 \; R8 ($p=0.9$) & \checkmark & CA & ASPP & $\ell_1$ & $\ell_1$ & \textbf{0.9} & 67.0 & 45.9 & 60.8 & 61.0 & 58.7 (-0.5) \\
 R15 \; R8 ($\ell_{mask}=\ell_2$) & \checkmark & CA & ASPP & $\bm{\ell_2}$  & $\ell_1$ & 0.1 & 67.3 & 46.8 & 62.0 & 62.9 & 59.8 (+0.6) \\
 R16 \; R8 ($\ell_{\Delta}=\ell_2$) & \checkmark & CA & ASPP & $\ell_1$ &  $\bm{\ell_2}$  & 0.1 & 68.7 & 47.2 & 63.3 & 62.9 & 60.5 (+1.3) \\
\bottomrule
\label{all_ablation}
\end{tabular}
}
\end{table*}

%% file: table8.tex
\begin{table*}[]
\caption{Ablation studies of Mask Ratio $\tau_m$ and Patch size $B$ for generating Styled Images $X_S$ on the CASID dataset \cite{liu2023large}. Notably, the green up-arrow indicates that this performance surpasses the baseline Rein's result, while the red down-arrow signifies the opposite. The values of $\tau_m$ and $B$ of the $X_M$ group keep 0.7 and 64 in these experiments.  }
\resizebox{\textwidth}{!}{
\begin{tabular}{c|ccccc|ccccc|ccccc}
\toprule
\multicolumn{16}{c}{Mask Ratio ($\tau_m$) and Patch Size ($B$) of Styled Image ($X_S$)}                                                     \\ \midrule
   \multicolumn{1}{c}{} & \multicolumn{5}{c}{Sub2Tem} & \multicolumn{5}{c}{Sub2Tms} & \multicolumn{5}{c}{Sub2Trf} \\ \hline
 \diagbox{$\tau_m$}{$B$}   & 16    & 32    & 64   & 128  & 256  & 16    & 32    & 64   & 128  & 256  & 16    & 32    & 64   & 128  & 256  \\ \hline
0.1 & 51.3 \textcolor[RGB]{182,207,58}{($\uparrow$)}  & 46.3 \textcolor[RGB]{182,207,58}{($\uparrow$)}  & 48.1 \textcolor[RGB]{182,207,58}{($\uparrow$)}  & 45.4 \textcolor[RGB]{245,183,191}{($\downarrow$)} & 41.7 \textcolor[RGB]{245,183,191}{($\downarrow$)} & 63.2 \textcolor[RGB]{182,207,58}{($\uparrow$)}   & 62.9 \textcolor[RGB]{182,207,58}{($\uparrow$)} & 64.2 \textcolor[RGB]{182,207,58}{($\uparrow$)}  & 61.5 \textcolor[RGB]{182,207,58}{($\uparrow$)}  & 58.1 \textcolor[RGB]{245,183,191}{($\downarrow$)} & 63.8 \textcolor[RGB]{182,207,58}{($\uparrow$)}   & 63.9 \textcolor[RGB]{182,207,58}{($\uparrow$)}  & 64.6 \textcolor[RGB]{182,207,58}{($\uparrow$)}  & 63.3 \textcolor[RGB]{245,183,191}{($\downarrow$)} & 63.4 (-) \\
0.3 & 38.4 \textcolor[RGB]{245,183,191}{($\downarrow$)}  & 47.5 \textcolor[RGB]{182,207,58}{($\uparrow$)}  & 47.3 \textcolor[RGB]{182,207,58}{($\uparrow$)} & 46.7 \textcolor[RGB]{182,207,58}{($\uparrow$)} & 41.3 \textcolor[RGB]{245,183,191}{($\downarrow$)} & 61.2 \textcolor[RGB]{182,207,58}{($\uparrow$)}  & 66.1 \textcolor[RGB]{182,207,58}{($\uparrow$)}  & 63.7 \textcolor[RGB]{182,207,58}{($\uparrow$)} & 62.5 \textcolor[RGB]{182,207,58}{($\uparrow$)} & 57.6 \textcolor[RGB]{245,183,191}{($\downarrow$)} & 60.9  \textcolor[RGB]{245,183,191}{($\downarrow$)} & 62.3 \textcolor[RGB]{245,183,191}{($\downarrow$)} & 64.6 \textcolor[RGB]{182,207,58}{($\uparrow$)} & 64.6 \textcolor[RGB]{182,207,58}{($\uparrow$)}& 63.7 \textcolor[RGB]{182,207,58}{($\uparrow$)}\\
0.5 & 49.8 \textcolor[RGB]{182,207,58}{($\uparrow$)}  & 44.9 \textcolor[RGB]{245,183,191}{($\downarrow$)} & 47.8 \textcolor[RGB]{182,207,58}{($\uparrow$)} & 41.4 \textcolor[RGB]{245,183,191}{($\downarrow$)}& 38.5 \textcolor[RGB]{245,183,191}{($\downarrow$)} & 64.8 \textcolor[RGB]{182,207,58}{($\uparrow$)}  & 64.4 \textcolor[RGB]{182,207,58}{($\uparrow$)} & 58.7 \textcolor[RGB]{245,183,191}{($\downarrow$)} & 58.8 \textcolor[RGB]{245,183,191}{($\downarrow$)} & 57.5 \textcolor[RGB]{245,183,191}{($\downarrow$)} & 63.5 \textcolor[RGB]{182,207,58}{($\uparrow$)} & 63.4 (-) & 62.8 \textcolor[RGB]{245,183,191}{($\downarrow$)} & 62.9 \textcolor[RGB]{245,183,191}{($\downarrow$)}& 62.7 \textcolor[RGB]{245,183,191}{($\downarrow$)}\\
0.7 & 48.2 \textcolor[RGB]{182,207,58}{($\uparrow$)} & 46.0  \textcolor[RGB]{245,183,191}{($\downarrow$)}  & 42.2  \textcolor[RGB]{245,183,191}{($\downarrow$)}& 39.1 \textcolor[RGB]{245,183,191}{($\downarrow$)}& 47.8 \textcolor[RGB]{182,207,58}{($\uparrow$)} & 63.3 \textcolor[RGB]{182,207,58}{($\uparrow$)} & 60.3 \textcolor[RGB]{182,207,58}{($\uparrow$)} & 61.5 \textcolor[RGB]{182,207,58}{($\uparrow$)} & 62.6 \textcolor[RGB]{182,207,58}{($\uparrow$)} & 63.5 \textcolor[RGB]{182,207,58}{($\uparrow$)}& 64.3 \textcolor[RGB]{182,207,58}{($\uparrow$)} & 63.9 \textcolor[RGB]{182,207,58}{($\uparrow$)} & 62.5 \textcolor[RGB]{245,183,191}{($\downarrow$)} & 60.5 \textcolor[RGB]{245,183,191}{($\downarrow$)}& 63.8 \textcolor[RGB]{182,207,58}{($\uparrow$)} \\
0.9 & 44.1 \textcolor[RGB]{245,183,191}{($\downarrow$)} & 40.7 \textcolor[RGB]{245,183,191}{($\downarrow$)} & 47.8 \textcolor[RGB]{182,207,58}{($\uparrow$)} & 42.4 \textcolor[RGB]{245,183,191}{($\downarrow$)}& 44.4 \textcolor[RGB]{245,183,191}{($\downarrow$)}& 66.0 \textcolor[RGB]{182,207,58}{($\uparrow$)} & 59.2 \textcolor[RGB]{245,183,191}{($\downarrow$)} & 64.6 \textcolor[RGB]{182,207,58}{($\uparrow$)} & 64.2 \textcolor[RGB]{182,207,58}{($\uparrow$)} & 59.0  \textcolor[RGB]{245,183,191}{($\downarrow$)} & 62.2 \textcolor[RGB]{245,183,191}{($\downarrow$)} & 62.0 \textcolor[RGB]{245,183,191}{($\downarrow$)} & 63.0 \textcolor[RGB]{245,183,191}{($\downarrow$)}  & 61.0  \textcolor[RGB]{245,183,191}{($\downarrow$)} & 63.6 \textcolor[RGB]{182,207,58}{($\uparrow$)} \\ \bottomrule
\end{tabular}}
\label{x_s ablation}
\end{table*}

%% file: table9.tex
\begin{table*}[]
\caption{Ablation studies of Mask Ratio $\tau_m$ and Patch size $B$ for generating Masked Images $X_M$ on the CASID dataset \cite{liu2023large}. Notably, colored fonts show the same meanings as Table 7, and the values of $\tau_m$ and $B$ of the $X_S$ group keep 0.1 and 64 in these experiments. }
\resizebox{\textwidth}{!}{
\begin{tabular}{c|ccccc|ccccc|ccccc}
\toprule
\multicolumn{16}{c}{Mask Ratio ($\tau_m$) and Patch Size ($B$) of Masked Image ($X_M$)}     
                                                  \\ \midrule
   \multicolumn{1}{c}{} & \multicolumn{5}{c}{Sub2Tem} & \multicolumn{5}{c}{Sub2Tms} & \multicolumn{5}{c}{Sub2Trf} \\ \hline
  \diagbox{$\tau_m$}{$B$}   & 16    & 32    & 64   & 128  & 256  & 16    & 32    & 64   & 128  & 256  & 16    & 32    & 64   & 128  & 256  \\ \hline
0.1 & 44.2 \textcolor[RGB]{245,183,191}{($\downarrow$)} & 46.7 \textcolor[RGB]{182,207,58}{($\uparrow$)}  & 42.4 \textcolor[RGB]{245,183,191}{($\downarrow$)} & 38.9 \textcolor[RGB]{245,183,191}{($\downarrow$)}& 44.7 \textcolor[RGB]{245,183,191}{($\downarrow$)}& 65.9 \textcolor[RGB]{182,207,58}{($\uparrow$)}  & 60.1 \textcolor[RGB]{182,207,58}{($\uparrow$)} & 53.0  \textcolor[RGB]{245,183,191}{($\downarrow$)} & 58.4 \textcolor[RGB]{245,183,191}{($\downarrow$)} & 57.3 \textcolor[RGB]{245,183,191}{($\downarrow$)} & 63.0 \textcolor[RGB]{245,183,191}{($\downarrow$)}   & 63.4 (-) & 57.9 \textcolor[RGB]{245,183,191}{($\downarrow$)}& 59.5 \textcolor[RGB]{245,183,191}{($\downarrow$)}& 63.6 \textcolor[RGB]{182,207,58}{($\uparrow$)} \\
0.3 & 39.0 \textcolor[RGB]{245,183,191}{($\downarrow$)}   & 44.8 \textcolor[RGB]{245,183,191}{($\downarrow$)} & 45.6 \textcolor[RGB]{245,183,191}{($\downarrow$)} & 45.9 \textcolor[RGB]{245,183,191}{($\downarrow$)} & 47.9 \textcolor[RGB]{182,207,58}{($\uparrow$)} & 57.6 \textcolor[RGB]{245,183,191}{($\downarrow$)}  & 61.9 \textcolor[RGB]{182,207,58}{($\uparrow$)} & 61.4 \textcolor[RGB]{182,207,58}{($\uparrow$)} & 62.0 \textcolor[RGB]{182,207,58}{($\uparrow$)}  & 61.4 \textcolor[RGB]{182,207,58}{($\uparrow$)}& 62.2 \textcolor[RGB]{245,183,191}{($\downarrow$)} & 59.5 \textcolor[RGB]{245,183,191}{($\downarrow$)} & 63.7 \textcolor[RGB]{182,207,58}{($\uparrow$)}& 63.6 \textcolor[RGB]{182,207,58}{($\uparrow$)}& 64.3 \textcolor[RGB]{182,207,58}{($\uparrow$)} \\
0.5 & 48.3 \textcolor[RGB]{182,207,58}{($\uparrow$)}  & 48.0  \textcolor[RGB]{182,207,58}{($\uparrow$)}   & 48.3 \textcolor[RGB]{182,207,58}{($\uparrow$)} & 49.2 \textcolor[RGB]{182,207,58}{($\uparrow$)}  & 51.4 \textcolor[RGB]{182,207,58}{($\uparrow$)} & 61.1 \textcolor[RGB]{182,207,58}{($\uparrow$)}   & 61.3 \textcolor[RGB]{182,207,58}{($\uparrow$)}  & 59.2 \textcolor[RGB]{245,183,191}{($\downarrow$)}  & 61.2 \textcolor[RGB]{182,207,58}{($\uparrow$)} & 59.8 \textcolor[RGB]{182,207,58}{($\uparrow$)}& 64.5 \textcolor[RGB]{182,207,58}{($\uparrow$)}  & 64.5 \textcolor[RGB]{182,207,58}{($\uparrow$)}  & 63.0 \textcolor[RGB]{245,183,191}{($\downarrow$)}  & 64.4 \textcolor[RGB]{182,207,58}{($\uparrow$)} & 63.0  \textcolor[RGB]{245,183,191}{($\downarrow$)} \\
0.7 & 43.9 \textcolor[RGB]{245,183,191}{($\downarrow$)} & 49.0    \textcolor[RGB]{182,207,58}{($\uparrow$)}& 48.1 \textcolor[RGB]{182,207,58}{($\uparrow$)} & 42.1 \textcolor[RGB]{245,183,191}{($\downarrow$)}& 43.1 \textcolor[RGB]{245,183,191}{($\downarrow$)} & 57.8  \textcolor[RGB]{245,183,191}{($\downarrow$)}& 63.6 \textcolor[RGB]{182,207,58}{($\uparrow$)} & 64.2 \textcolor[RGB]{182,207,58}{($\uparrow$)}& 63.4 \textcolor[RGB]{182,207,58}{($\uparrow$)}& 62.7 \textcolor[RGB]{182,207,58}{($\uparrow$)}& 63.0 \textcolor[RGB]{245,183,191}{($\downarrow$)}& 63.8 \textcolor[RGB]{182,207,58}{($\uparrow$)}  & 64.6 \textcolor[RGB]{182,207,58}{($\uparrow$)}& 61.6 \textcolor[RGB]{245,183,191}{($\downarrow$)}& 62.1 \textcolor[RGB]{245,183,191}{($\downarrow$)}\\
0.9 & 42.9  \textcolor[RGB]{245,183,191}{($\downarrow$)}& 50.7 \textcolor[RGB]{182,207,58}{($\uparrow$)} & 43.8 \textcolor[RGB]{245,183,191}{($\downarrow$)}& 41.8 \textcolor[RGB]{245,183,191}{($\downarrow$)}& 46.6 \textcolor[RGB]{182,207,58}{($\uparrow$)}& 59.7 (-) & 62.4 \textcolor[RGB]{182,207,58}{($\uparrow$)} & 62.8 \textcolor[RGB]{182,207,58}{($\uparrow$)}& 62.2 \textcolor[RGB]{182,207,58}{($\uparrow$)} & 61.9 \textcolor[RGB]{182,207,58}{($\uparrow$)}& 63.5  \textcolor[RGB]{182,207,58}{($\uparrow$)}& 64.4  \textcolor[RGB]{182,207,58}{($\uparrow$)}& 63.8 \textcolor[RGB]{182,207,58}{($\uparrow$)}& 60.2 \textcolor[RGB]{245,183,191}{($\downarrow$)}& 64.0 \textcolor[RGB]{182,207,58}{($\uparrow$)}  \\ \bottomrule
\end{tabular}}
\label{x_m ablation}
\end{table*}

%% file: rescue-to-potsdam.tex
\begin{table}[H]
    \centering
    \caption{Mapping RescueNet labels to Potsdam}
    \resizebox{1\linewidth}{!}{
    
    \begin{tabular}{c|c|c}
    \toprule
        Dataset &Label change & Final Label Category  \\
       \midrule
          Potsdam  & Vegt $\times$  & \multirow{6}{*}{Surf, Bldg, Tree, Car, Clut} \\
          RescueNet & Bldg-related $\rightarrow$ Bldg & \\
          RescueNet & Vehicle $\rightarrow$ Car \\
          RescueNet & Water, Pool $\rightarrow$ Bkdg \\
         RescueNet & Road-related $\rightarrow$ Surf & \\
          RescueNet& Bkdg $\rightarrow$ Clut& \\
         \bottomrule
    \end{tabular}}
    
    \label{label change: rescue2potsdam}
\end{table}

%% file: oem-to-loveda.tex
\begin{table}[H]
    \centering
    \caption{Mapping OpenEarthMap (OEM) labels to LoveDA}
    
    \resizebox{1\linewidth}{!}{
    \begin{tabular}{c|c|c}
    \toprule
        Dataset & Label change & Final Label Category  \\
       \midrule
          OEM & bareland $\rightarrow$ Barr & \multirow{8}{*}{Bkgd, Bldg, Rd, Wtr, Barr, Frst, Agri} \\
          OEM & rangeland $\rightarrow$ Bkgd & \\
          OEM & developed space $\rightarrow$ Bkgd & \\
          OEM & road $\rightarrow$ Rd & \\
          OEM & tree $\rightarrow$ Frst & \\
          OEM & water $\rightarrow$ Wtr & \\
          OEM & agriculture land $\rightarrow$ Agri & \\
          OEM & building $\rightarrow$ Bldg & \\
         \bottomrule
    \end{tabular}}
    \label{label_change_oem}
\end{table}

%% file: table6_sup.tex
\onecolumn
\begin{table}[t]
\caption{Performance comparison in CASID benchmarks. The comparison consists of 4 in-domain and 12 out-of-domain experiments between various specialized models and VFMs. Sub: Subtropical Monsoon. Tem: Temperate Monsoon. Tms: Tropical Monsoon. Trf: Tropical Rainforest.}
    \tiny
    \centering
    \resizebox{0.9\textwidth}{!}{
    \begin{tabular}{cccccccccccccccc}
    \toprule
    \multirow{4}{*}{Method} & \multirow{4}{*}{Backbone} & \multicolumn{1}{c}{Domain} & \multicolumn{5}{c}{Classes} & \multirow{2}{*}{mIoU (\%)} & \multicolumn{1}{c}{Domain} & \multicolumn{5}{c}{Classes} & \multirow{2}{*}{mIoU (\%)} 
    \\ \cline{3-8} \cline{10-15}
    & &  & \multirow{2}{*}{Bkgd} &\multirow{2}{*}{Bldg}
&\multirow{2}{*}{Frst}&\multirow{2}{*}{Rd}&\multirow{2}{*}{Wtr}& &\multirow{2}{*}{Source2Unseen}&\multirow{2}{*}{Bkgd} &\multirow{2}{*}{Bldg}&\multirow{2}{*}{Frst}&\multirow{2}{*}{Rd}&\multirow{2}{*}{Wtr}&\\
\\ \midrule \multicolumn{16}{c}{Performance Comparison in \textbf{Cross-Domain Generalization} Setting on CASID benchmark}  \\ \midrule
\multirow{1}{*}{DeepLabv2 \cite{chen2017deeplab}}&\multirow{1}{*}{ResNet-101 \cite{he2016deep}}& \multirow{11}{*}{Sub (Source-Only)} & 56.0&78.2&\underline{83.8}&24.4&62.3&60.9 & & 0.3&8.3&41.2&0.1&0.4&10.0 \\
\multirow{1}{*}{DAFormer \cite{hoyer2022daformer}}&\multirow{1}{*}{MiT-B5 \cite{xie2021segformer}}&  & 60.3&81.5& 83.1&43.5&\underline{71.6}&\underline{68.0} & & 40.2&61.8&30.0&40.4&7.7&36.0 \\
\multirow{1}{*}{HRDA \cite{hoyer2022hrda}}&\multirow{1}{*}{MiT-B5 \cite{xie2021segformer}}&  & 56.2& 81.5&78.4&44.1&71.1&66.3 & & 37.4&63.8&19.8&39.8&12.2&34.6 \\

\multirow{1}{*}{S12-MoCo \cite{stewart2023ssl4eo}}&\multirow{1}{*}{ViT-S \cite{dosovitskiy2020image} }&  & 43.0 & 68.7 &65.2 & 0.6 & 24.1 & 40.3 & & 39.7 &25.4 & 31.3 & 1.1 & 0.7 & 19.6 \\

\multirow{1}{*}{S12-DINO \cite{stewart2023ssl4eo}}&\multirow{1}{*}{ViT-S \cite{dosovitskiy2020image} }&  & 35.2 & 39.4 & 61.6 & 0.2 & 1.4 & 27.6 & &35.9 & 33.1 & 23.4 & 0.1 & 0.8 & 18.7    \\

\multirow{1}{*}{S12-MAE \cite{stewart2023ssl4eo}}&\multirow{1}{*}{ViT-S \cite{dosovitskiy2020image} }&  & 35.1 & 51.7 & 53.3 & 0.1 & 0.5 & 28.2 & & 35.8 & 28.2 & 24.6 & 0.1 & 2.2 & 18.2   \\

\multirow{1}{*}{DOFA \cite{xiong2024neural}}&\multirow{1}{*}{ViT-B \cite{dosovitskiy2020image} }&  & 39.2 & 60.7 & 60.7 & 0.2 & 7.4 & 33.6 & & 35.4& 32.0 & 23.9 & 0.1 & 1.1 & 18.5  \\

\multirow{1}{*}{SatMAE \cite{cong2022satmae} }&\multirow{1}{*}{ViT-L \cite{dosovitskiy2020image} }&  Source2Unseen&  52.8 & 78.6 & 76.2 & 29.8 & 66.4& 60.8  & Sub2Tem& 39.3 & 59.8 & 28.5 & 31.9 & 6.1 & 33.1  \\

\multirow{1}{*}{ScaleMAE  \cite{reed2023scale}}&\multirow{1}{*}{ViT-L \cite{dosovitskiy2020image} }&  & 58.7 & 81.1& 81.2 & 38.8 & 71.4 &66.3 & & 40.7 & 62.7 & 33.2 & 36.1 & 23.2 & 39.2    \\

\multirow{1}{*}{RemoteCLIP \cite{liu2024remoteclip}}&\multirow{1}{*}{ViT-L \cite{dosovitskiy2020image} }&  & 26.8  & 12.7 & 43.4 & 0.0 & 0.0 & 16.6& & 32.6 & 2.9& 6.9&0.0 &0.0 &8.5 \\
\multirow{1}{*}{MTP \cite{wang2024mtp}}&\multirow{1}{*}{ViT-L \cite{dosovitskiy2020image} }&  & \underline{62.0}&80.1&83.7&44.3& 69.9&\underline{68.0} & &41.7&62.3&34.3&39.5&19.3&39.4 \\
\multirow{1}{*}{Rein (Baseline) \cite{Wei_2024_CVPR}}&\multirow{1}{*}{ViT-L \cite{dosovitskiy2020image}}&  & 59.1& \textbf{82.3} &80.0& \underline{44.4}&70.2&67.2 & &\underline{45.7}&\underline{69.6}&\underline{46.5}&\underline{41.1}&\textbf{29.3}& \underline{46.4}\\
\multirow{1}{*}{CrossEarth (Ours)}&\multirow{1}{*}{ViT-L \cite{dosovitskiy2020image}}&  & \textbf{63.3} &\underline{82.2}&\textbf{84.7}&\textbf{44.9}&\textbf{71.7}&\textbf{69.4 (+2.2)} & &\textbf{47.1}&\textbf{70.0}&\textbf{50.3}&\textbf{43.9}&\underline{28.6}& \textbf{48.1 (+1.7)}  \\\midrule

\multirow{1}{*}{DeepLabv2 \cite{chen2017deeplab} }&\multirow{1}{*}{ResNet-101 \cite{he2016deep} }&  & 1.1&4.8&53.3&0.1&4.7&12.8 & & 0.4&2.6&70.1&0.1&3.5&15.3 \\
\multirow{1}{*}{DAFormer \cite{hoyer2022daformer} }&\multirow{1}{*}{MiT-B5 \cite{xie2021segformer} }&  & \underline{69.5}&65.0&\underline{78.6}&34.3&\underline{69.9}&\underline{63.5} & &29.2&73.8&\underline{94.7}&40.9&56.4&59.0 \\
\multirow{1}{*}{HRDA \cite{hoyer2022hrda}}&\multirow{1}{*}{MiT-B5 \cite{xie2021segformer}}&  & 68.9&64.6&77.3&33.6&\textbf{70.3}&63.0 & & 23.2&68.1&91.9&40.7&57.4&56.2 \\

\multirow{1}{*}{S12-MoCo \cite{stewart2023ssl4eo}}&\multirow{1}{*}{ViT-S \cite{dosovitskiy2020image} }& & 54.3 & 23.8 & 55.0 & 0.1 & 13.3 &29.3 & & 19.6 & 49.6 & 90.4 & 0.8 & 0.0 & 32.1\\

\multirow{1}{*}{S12-DINO \cite{stewart2023ssl4eo}}&\multirow{1}{*}{ViT-S \cite{dosovitskiy2020image} }&  & 47.7 & 18.0 &40.6 & 0.0 & 0.1 & 21.3  & & 17.1 & 22.5 & 90.0 & 1.3 & 0.2 & 26.2 \\

\multirow{1}{*}{S12-MAE \cite{stewart2023ssl4eo}}&\multirow{1}{*}{ViT-S \cite{dosovitskiy2020image} }&  & 44.6 &19.7 & 35.2 & 0.0 & 0.2 &19.9 & & 15.1 & 29.2 & 86.9 & 0.4 & 0.1 & 26.3 \\

\multirow{1}{*}{DOFA \cite{xiong2024neural} \cite{xiong2024neural}}&\multirow{1}{*}{ViT-B \cite{dosovitskiy2020image} }&  & 49.3 &31.7 & 45.3 & 0.1 & 0.5 & 25.3 & & 17.7 & 40.4 & 90.6 & 1.8 &0.1 & 30.1 \\

\multirow{1}{*}{SatMAE \cite{cong2022satmae} }&\multirow{1}{*}{ViT-L \cite{dosovitskiy2020image} }&  Sub2Tms& 61.7 & 55.7 & 68.9 & 15.7 & 62.6 & 52.9 &  Sub2Trf&22.8& 60.1 & 92.5 & 31.6 & 36.5 & 48.7 \\

\multirow{1}{*}{ScaleMAE \cite{reed2023scale}}&\multirow{1}{*}{ViT-L \cite{dosovitskiy2020image} }&  &  66.3 & 62.5 & 74.3 & 29.4 & 70.8 & 60.7 & & 27.7 & 70.5 & 93.8 & 36.5 & 53.3 &  56.4\\

\multirow{1}{*}{RemoteCLIP \cite{liu2024remoteclip}}&\multirow{1}{*}{ViT-L \cite{dosovitskiy2020image} }&  & 42.6 &  6.8 & 25.8 & 0.1 & 0.0 & 15.1& &10.2 & 8.0 &79.8 & 0.2 &0.0 & 19.7 \\
\multirow{1}{*}{MTP \cite{wang2024mtp}}&\multirow{1}{*}{ViT-L \cite{dosovitskiy2020image}}&  & 68.3 &59.1&77.6&27.2&67.9&60.0 & &\textbf{32.7}&\textbf{75.9}&94.1&36.6&68.5&61.6 \\
\multirow{1}{*}{Rein (Baseline) \cite{Wei_2024_CVPR} }&\multirow{1}{*}{ViT-L \cite{dosovitskiy2020image}}&  & 62.9&\underline{68.2}&69.3& \underline{35.8}&62.2&59.7 & &31.1&{75.5}&93.6&\underline{41.3}&\underline{75.7}& \underline{63.4} \\
\multirow{1}{*}{CrossEarth (Ours)}&\multirow{1}{*}{ViT-L \cite{dosovitskiy2020image}}&  & \textbf{71.9}&\textbf{68.4}& \textbf{80.2}&\textbf{38.8}& {63.5}&\textbf{64.6 (+4.9)} & &\underline{32.4}&\underline{75.8}&\textbf{95.0}&\textbf{42.0}&\textbf{76.0}&\textbf{64.2 (+0.8)} \\\midrule

\multirow{1}{*}{DeepLabv2 \cite{chen2017deeplab}}&\multirow{1}{*}{ResNet-101 \cite{he2016deep}}& & 49.4&60.9&62.6&24.7&1.9&39.9 &  & 0.7&2.8&41.4&0.1&1.9&9.4 \\
\multirow{1}{*}{DAFormer \cite{hoyer2022daformer}}&\multirow{1}{*}{MiT-B5 \cite{xie2021segformer}}&  & 51.1&67.3&64.2&{39.8}&2.6&45.0 & & 59.5&\textbf{80.8}&80.6&\underline{36.9}&41.2&59.8 \\
\multirow{1}{*}{HRDA \cite{hoyer2022hrda}}&\multirow{1}{*}{MiT-B5 \cite{xie2021segformer}}&  & 52.6&65.3&\underline{65.3}&38.7&9.9&46.3 & & 54.1&79.4&82.8&34.9&62.6&62.7 \\

\multirow{1}{*}{S12-MoCo \cite{stewart2023ssl4eo}}&\multirow{1}{*}{ViT-S \cite{dosovitskiy2020image} }& & 36.0 & 30.0 & 28.1 & 0.0 & 0.0 & 18.3 & & 29.0 &41.2&37.3&0.0&0.0 & 21.5   \\

\multirow{1}{*}{S12-DINO \cite{stewart2023ssl4eo}}&\multirow{1}{*}{ViT-S \cite{dosovitskiy2020image} }&  & 37.5 & 29.5 & 44.9 & 0.3 & 0.0 & 22.4 & & 35.2 & 47.4 & 57.9 & 0.1 & 0.0  & 28.1  \\

\multirow{1}{*}{S12-MAE \cite{stewart2023ssl4eo}}&\multirow{1}{*}{ViT-S \cite{dosovitskiy2020image} }&  & 35.8 & 30.9 & 39.4 & 0.0 & 0.0 & 21.2 & & 31.7 & 32.7 & 62.6 & 0.0 & 0.0 & 25.4   \\

\multirow{1}{*}{DOFA \cite{xiong2024neural} }&\multirow{1}{*}{ViT-B \cite{dosovitskiy2020image} }&  & 36.5 & 31.2 & 38.2 & 0.0 & 0.0 & 21.2 & & 31.9 & 41.6 & 58.3 & 0.1 & 0.0 &26.4 \\

\multirow{1}{*}{SatMAE \cite{cong2022satmae}}&\multirow{1}{*}{ViT-L \cite{dosovitskiy2020image} }&  Tem (Source-Only)& 54.4 & 59.3 & 71.6 & 25.0 & 10.6 & 36.2 & Tem2Sub&50.1 & 74.8 & 80.4 & 19.1 & 34.6& 51.8   \\

\multirow{1}{*}{ScaleMAE \cite{reed2023scale}}&\multirow{1}{*}{ViT-L \cite{dosovitskiy2020image} }&  & 53.5 & 63.7 & 76.9 & 33.8 & 24.1 & 50.4 & &56.1 & 78.7 & 83.3 & 29.1 & 65.3 & 62.5 \\

\multirow{1}{*}{RemoteCLIP \cite{liu2024remoteclip}}&\multirow{1}{*}{ViT-L \cite{dosovitskiy2020image} }&  & 27.7 & 16.7  & 45.7& 0.0& 0.0 &18.0 & &25.6 & 32.0& 49.3& 0.0& 0.0& 54.3  \\
\multirow{1}{*}{MTP \cite{wang2024mtp}}&\multirow{1}{*}{ViT-L \cite{dosovitskiy2020image}}&  & 49.2&68.9&52.3&\underline{40.6}&3.4&42.9 & &{59.9}&\underline{80.6}&84.6&\textbf{41.0}&\textbf{65.0}&\textbf{66.2} \\
\multirow{1}{*}{Rein (Baseline) \cite{Wei_2024_CVPR}}&\multirow{1}{*}{ViT-L \cite{dosovitskiy2020image}}&  & \underline{54.0}& \textbf{70.3}&64.3&\textbf{40.9}& \textbf{27.2}&\underline{51.3} & &\textbf{61.8}&80.5&\textbf{84.9}&36.8&63.5& \underline{65.5} \\
\multirow{1}{*}{CrossEarth (Ours)}&\multirow{1}{*}{ViT-L \cite{dosovitskiy2020image}}&  &\textbf{56.7}&\underline{69.9}& \textbf{74.7}&{39.4}&\underline{26.5}&\textbf{53.5 (+2.2)} & &\underline{57.1}&77.8&\underline{84.3}&33.6&\underline{64.4}&\underline{63.5 (-2.0)} \\\midrule

\multirow{1}{*}{DeepLabv2 \cite{chen2017deeplab}}&\multirow{1}{*}{ResNet-101 \cite{he2016deep}}&  & 1.2&1.1&48.8&0.2&3.0&10.9 & & 0.4&0.6&58.7&0.1&1.2&12.2 \\
\multirow{1}{*}{DAFormer \cite{hoyer2022daformer}}&\multirow{1}{*}{MiT-B5 \cite{xie2021segformer}}&  & 73.0&63.7&84.0&23.4&74.7&\underline{63.8} & & \underline{20.5}&{67.1}&\textbf{91.8}&34.0&45.2&51.7 \\
\multirow{1}{*}{HRDA \cite{hoyer2022hrda}}&\multirow{1}{*}{MiT-B5 \cite{xie2021segformer}}&  & \underline{75.4}&60.6&{85.8}&19.2&\underline{75.7}&63.3 & & 19.2&66.6&89.3&28.2&21.3&44.9 \\

\multirow{1}{*}{S12-MoCo \cite{stewart2023ssl4eo}}&\multirow{1}{*}{ViT-S \cite{dosovitskiy2020image} }&  & 41.6 & 12.2 & 31.7 & 0.0 & 0.0 & 17.1 & & 4.8 & 28.1 & 27.9 & 0.0 & 0.0 &12.2    \\

\multirow{1}{*}{S12-DINO \cite{stewart2023ssl4eo}}&\multirow{1}{*}{ViT-S \cite{dosovitskiy2020image} }&  & 45.9 & 12.4 & 51.1 & 0.0 & 0.0 & 21.9 & & 10.2 & 33.6 & 79.4 & 0.0 & 0.0 & 24.7  \\

\multirow{1}{*}{S12-MAE \cite{stewart2023ssl4eo}}&\multirow{1}{*}{ViT-S \cite{dosovitskiy2020image} }&  & 46.7 & 10.1 & 55.3 & 0.0 & 0.0 & 22.4 & & 11.2 & 29.9 & 83.2 & 0.0 & 0.0 & 24.9   \\

\multirow{1}{*}{DOFA \cite{xiong2024neural} }&\multirow{1}{*}{ViT-B \cite{dosovitskiy2020image} }&  &46.8&11.5&56.3& 0.0 & 0.0 &22.9 &  & 7.6 & 25.4 & 71.0 & 0.0 &0.0 &20.8  \\

\multirow{1}{*}{SatMAE \cite{cong2022satmae}}&\multirow{1}{*}{ViT-L \cite{dosovitskiy2020image} }&  Tem2Tms&  64.4& 49.9 & 82.2&2.4&19.4&43.7 & Tem2Trf& 19.1& 51.5 & 90.9& 13.9& 6.2 &36.2\\

\multirow{1}{*}{ScaleMAE \cite{reed2023scale}}&\multirow{1}{*}{ViT-L \cite{dosovitskiy2020image} }&  & 72.1& 60.5 & 85.1 & 12.6 & 56.4& 57.3 & &21.9& 63.4& 93.4& 21.2& 29.8& 45.9 \\

\multirow{1}{*}{RemoteCLIP \cite{liu2024remoteclip}}&\multirow{1}{*}{ViT-L \cite{dosovitskiy2020image} }& &  33.5 & 11.3 & 50.7  &0.0 &0.0 &57.8 & &4.5 & 15.6& 63.4& 0.0& 0.0& 16.7 \\
\multirow{1}{*}{MTP \cite{wang2024mtp}}&\multirow{1}{*}{ViT-L \cite{dosovitskiy2020image}}&  & 74.3&\textbf{65.1}&84.9&20.0&\underline{79.1}&\textbf{64.7} & &19.3&{67.1}&90.8&\underline{34.4}&29.9&48.3 \\
\multirow{1}{*}{Rein (Baseline) \cite{Wei_2024_CVPR}}&\multirow{1}{*}{ViT-L \cite{dosovitskiy2020image}}&  & \textbf{75.5}&\underline{64.9}&\textbf{86.7}&\textbf{26.9}&60.4&62.9 & &20.4&\textbf{70.6}&89.0&33.1&\underline{66.2}&\underline{55.9} \\
\multirow{1}{*}{CrossEarth (Ours)}&\multirow{1}{*}{ViT-L \cite{dosovitskiy2020image}}&  &{75.3}& 59.9&\underline{86.0}&\underline{26.0}&61.7&61.8 (-1.1) & &\textbf{21.1}&\underline{67.8}&\underline{92.0}&\textbf{35.0}&\textbf{72.9}&\textbf{57.8 (+1.9)}  \\\midrule

\multirow{1}{*}{DeepLabv2 \cite{chen2017deeplab}}&\multirow{1}{*}{ResNet-101 \cite{he2016deep}}&  & \underline{68.9}&59.8&78.5&14.0&61.4&56.5 & & 1.5&17.2&43.2&0.0&2.2&12.8 \\
\multirow{1}{*}{DAFormer \cite{hoyer2022daformer}}&\multirow{1}{*}{MiT-B5 \cite{xie2021segformer}}&  & 65.8&68.0&75.1&35.2&\underline{76.5}&64.1 & & 57.2&79.8&77.9&39.5&54.4&61.8 \\
\multirow{1}{*}{HRDA \cite{hoyer2022hrda}}&\multirow{1}{*}{MiT-B5 \cite{xie2021segformer}}&  & 65.2&67.6&74.7&36.1&{69.2}&62.6 & & 55.7&81.6&78.1&42.6&61.0&63.8 \\

\multirow{1}{*}{S12-MoCo \cite{stewart2023ssl4eo}}&\multirow{1}{*}{ViT-S \cite{dosovitskiy2020image} }&  & 46.5& 22.2& 36.2 & 0.2 & 0.1 & 21.1 & & 35.3& 55.6 & 52.8 & 0.2 & 0.0 & 28.8 \\

\multirow{1}{*}{S12-DINO \cite{stewart2023ssl4eo}}&\multirow{1}{*}{ViT-S \cite{dosovitskiy2020image} }&  & 44.5 & 12.1 & 30.6&0.2&0.0 & 17.5 & & 30.7 & 35.8 & 48.0 & 0.1 & 0.0 & 22.9  \\

\multirow{1}{*}{S12-MAE \cite{stewart2023ssl4eo}}&\multirow{1}{*}{ViT-S \cite{dosovitskiy2020image} }&  & 45.6 & 8.2 & 33.4 & 0.2 & 0.0 & 17.5 & & 33.7 & 46.2 & 52.9 & 0.3 & 0.0 & 26.6  \\

\multirow{1}{*}{DOFA \cite{xiong2024neural} }&\multirow{1}{*}{ViT-B \cite{dosovitskiy2020image} }& & 50.7 & 21.9 & 46.0 & 1.1 & 0.8 &24.1& & 36.8 & 51.8& 57.5 & 2.1 & 0.6 & 29.8\\

\multirow{1}{*}{SatMAE \cite{cong2022satmae}}&\multirow{1}{*}{ViT-L \cite{dosovitskiy2020image} }&  Tms (Source-Only)& 67.1&56.1&76.5&20.6&66.5&57.4 & Tms2Sub&55.9&76.4&80.3&32.5&56.0&60.2 \\

\multirow{1}{*}{ScaleMAE \cite{reed2023scale}}&\multirow{1}{*}{ViT-L \cite{dosovitskiy2020image} }&  & 69.6& 63.4& 79.2& 23.0& 70.3&61.1 & & 58.8& 78.8& 80.8& 32.7& 59.5 &62.1\\

\multirow{1}{*}{RemoteCLIP \cite{liu2024remoteclip}}&\multirow{1}{*}{ViT-L \cite{dosovitskiy2020image} }&  & 44.4 & 14.9  & 33.9& 0.2&0.0  & 18.7& & 8.3& 19.2& 84.3& 0.6& 0.0& 22.5 \\
\multirow{1}{*}{MTP \cite{wang2024mtp}}&\multirow{1}{*}{ViT-L \cite{dosovitskiy2020image}}&  & 67.5&68.8& 75.3&38.0& 63.1&62.5 & &55.1&79.6&70.7&41.9&33.1&56.1 \\
\multirow{1}{*}{Rein (Baseline) \cite{Wei_2024_CVPR}}&\multirow{1}{*}{ViT-L \cite{dosovitskiy2020image}}&  & 68.2&\underline{69.6}&\underline{78.9}&\textbf{43.8}&64.5&\underline{65.0} & &\underline{62.7}&\underline{82.2}&\underline{83.4}&\underline{48.1}&\underline{66.8}& \underline{68.6} \\
\multirow{1}{*}{CrossEarth (Ours)}&\multirow{1}{*}{ViT-L \cite{dosovitskiy2020image}}&  &\textbf{71.2}& \textbf{69.9}&\textbf{81.7}& \underline{43.6}&\textbf{78.1}&\textbf{68.9 (+3.9)} & &\textbf{63.2}&\textbf{82.5}&\textbf{83.9}&\textbf{48.3}&\textbf{67.6}&\textbf{69.1 (+0.5)}
\\\midrule

\multirow{1}{*}{DeepLabv2 \cite{chen2017deeplab}}&\multirow{1}{*}{ResNet-101 \cite{he2016deep}}&  & 0.5&3.1&\textbf{56.6}&0.0&0.6&12.2 & & 0.7&0.4&41.2&0.0&2.9&9.1 \\
\multirow{1}{*}{DAFormer \cite{hoyer2022daformer}}&\multirow{1}{*}{MiT-B5 \cite{xie2021segformer}}&  & 37.5&66.0&14.8&37.3&1.6&31.4 & & 22.6&76.2&90.3&40.5&51.3&56.2 \\
\multirow{1}{*}{HRDA \cite{hoyer2022hrda}}&\multirow{1}{*}{MiT-B5 \cite{xie2021segformer}}&  & \underline{38.8}&66.6&19.6&40.2&1.5&33.3 & & \underline{24.3}&71.9&\underline{92.4}&36.4&59.7&56.9 \\

\multirow{1}{*}{S12-MoCo \cite{stewart2023ssl4eo}}&\multirow{1}{*}{ViT-S \cite{dosovitskiy2020image} }&  & 39.0 & 32.8 & 35.8 & 0.3 & 0.1 &21.6 & & 12.7 & 36.0 & 86.7 & 0.6 & 0.4 &27.3   \\

\multirow{1}{*}{S12-DINO \cite{stewart2023ssl4eo}}&\multirow{1}{*}{ViT-S \cite{dosovitskiy2020image} }&  & 36.5 & 28.2 & 24.9 & 0.1 & 0.0 & 18.0 & & 9.3 & 24.9 & 79.3 & 0.9 & 0.3 & 23.0   \\

\multirow{1}{*}{S12-MAE \cite{stewart2023ssl4eo}}&\multirow{1}{*}{ViT-S \cite{dosovitskiy2020image} }&  & 37.5 & 29.0 & 28.0 & 0.2 & 0.0 & 18.9 & & 13.9 & 33.5 & 89.4 & 1.1 & 0.1 & 27.6   \\

\multirow{1}{*}{DOFA \cite{xiong2024neural} }&\multirow{1}{*}{ViT-B \cite{dosovitskiy2020image} }&  & 37.0 &32.7 & 22.2 & 1.8 & 0.2 & 18.8 & & 14.6 & 35.0 & 85.5 & 9.2 & 1.3 & 29.1   \\

\multirow{1}{*}{SatMAE \cite{cong2022satmae}}&\multirow{1}{*}{ViT-L \cite{dosovitskiy2020image} }&  Tms2Tem& 40.9&55.2&31.6&32.5&2.2 & 32.5& Tms2Trf&26.0&63.5&94.1&31.2&50.2&53.0   \\

\multirow{1}{*}{ScaleMAE \cite{reed2023scale}}&\multirow{1}{*}{ViT-L \cite{dosovitskiy2020image} }&  & 41.7&63.0&35.3&33.5&2.6&35.2 & &27.9&71.2&94.2&31.2&52.4&55.4   \\

\multirow{1}{*}{RemoteCLIP \cite{liu2024remoteclip}}&\multirow{1}{*}{ViT-L \cite{dosovitskiy2020image} }&  &33.4  &14.3   & 13.8& 0.0 &0.0  &12.3 & &  8.3  & 19.2&84.3 &0.6 &0.0 & 22.5    \\
\multirow{1}{*}{MTP \cite{wang2024mtp}}&\multirow{1}{*}{ViT-L \cite{dosovitskiy2020image}}&  & 36.6&67.6&10.2&44.4&2.6&32.3 & &\textbf{31.9}&76.2&\textbf{93.4}&\textbf{46.0}&57.3&\underline{60.1} \\
\multirow{1}{*}{Rein (Baseline) \cite{Wei_2024_CVPR}}&\multirow{1}{*}{ViT-L \cite{dosovitskiy2020image}}&  & 36.8& \underline{73.5}&18.6&\textbf{47.1}&\textbf{10.0}&\underline{37.2} & &19.6&\textbf{77.8}&85.8&\underline{45.9}&\textbf{69.5}&59.7 \\
\multirow{1}{*}{CrossEarth (Ours)}&\multirow{1}{*}{ViT-L \cite{dosovitskiy2020image}}&  &\textbf{41.7}&\textbf{73.7}&\underline{35.0}&\underline{46.4}&\underline{5.7}&\textbf{40.5 (+3.3)} & &21.7&\underline{77.5}&89.4&45.5&\underline{69.4}&\textbf{60.7 (+1.0)} \\\midrule

\multirow{1}{*}{DeepLabv2 \cite{chen2017deeplab}}&\multirow{1}{*}{ResNet-101 \cite{he2016deep}}&  & 15.0&70.0&92.8&10.8&56.4&49.0 & & 1.8&5.5&41.1&0.0&2.7&10.2 \\
\multirow{1}{*}{DAFormer \cite{hoyer2022daformer}}&\multirow{1}{*}{MiT-B5 \cite{xie2021segformer}}&  & 25.6&\textbf{78.1}&\underline{95.5}&\underline{44.7}&\underline{79.1}&\underline{64.6} & & 54.0&79.6&78.2&42.7&60.0&62.9 \\
\multirow{1}{*}{HRDA \cite{hoyer2022hrda}}&\multirow{1}{*}{MiT-B5 \cite{xie2021segformer}}&  & 23.5&76.0&\textbf{95.6}&42.5&\textbf{80.1}&63.5 & & 47.8&79.7&77.6&42.3&\underline{69.3}&63.3 \\

\multirow{1}{*}{S12-MoCo \cite{stewart2023ssl4eo}}&\multirow{1}{*}{ViT-S \cite{dosovitskiy2020image} }&  & 16.6 & 32.5 & 92.5 & 2.0 & 0.9 & 28.9 & &40.7 & 53.0 & 71.3 & 0.1 & 0.2 & 33.0  \\

\multirow{1}{*}{S12-DINO \cite{stewart2023ssl4eo}}&\multirow{1}{*}{ViT-S \cite{dosovitskiy2020image} }&  & 17.8 & 47.0 & 91.4 & 0.0 & 1.2 & 31.5 & & 42.9 & 61.1 & 70.8 & 0.0 & 0.2 & 35.0  \\

\multirow{1}{*}{S12-MAE \cite{stewart2023ssl4eo}}&\multirow{1}{*}{ViT-S \cite{dosovitskiy2020image} }&  & 16.8 & 52.4 & 91.6 & 0.7 & 0.9 & 32.5 & & 39.3 & 62.1 & 67.4 & 0.0 & 0.9 & 34.0   \\

\multirow{1}{*}{DOFA \cite{xiong2024neural} }&\multirow{1}{*}{ViT-B \cite{dosovitskiy2020image} }&  &  18.5 & 43.9& 92.8 & 1.0 & 1.0& 31.2& & 44.9&63.5&70.7&0.0&1.3 &36.1  \\

\multirow{1}{*}{SatMAE \cite{cong2022satmae}}&\multirow{1}{*}{ViT-L \cite{dosovitskiy2020image} }&  Trf (Source-Only)& 26.5&69.7&95.3&34.7& 63.0 & 57.8 & Trf2Sub& 49.1 & 76.8& 79.2& 30.5& 59.9& 59.1   \\

    \multirow{1}{*}{ScaleMAE \cite{reed2023scale}}&\multirow{1}{*}{ViT-L \cite{dosovitskiy2020image} }&  & 27.7&72.7&95.2&35.4&66.3& 59.5 & & 52.8&78.9&78.4&35.3&57.8& 60.6   \\

\multirow{1}{*}{RemoteCLIP \cite{liu2024remoteclip}}&\multirow{1}{*}{ViT-L \cite{dosovitskiy2020image} }&  & 12.2 & 29.0  &  90.8& 0.1 & 0.1 & 26.4& &37.1  & 52.9 & 70.6 & 0.0 & 0.0& 32.1 \\
\multirow{1}{*}{MTP \cite{wang2024mtp}}&\multirow{1}{*}{ViT-L \cite{dosovitskiy2020image}}&  & \underline{27.5}&74.0&\underline{95.5}&42.4&74.6&62.8 & &48.7&80.4&74.1&42.4&32.3&55.6 \\
\multirow{1}{*}{Rein (Baseline) \cite{Wei_2024_CVPR}}&\multirow{1}{*}{ViT-L \cite{dosovitskiy2020image}}&  & 21.0&\underline{77.9}&88.1&\textbf{46.2}&71.9&61.0 & &\underline{59.5}&\textbf{82.2}&\underline{80.8}&\textbf{47.8}&67.6&\underline{67.6} \\
\multirow{1}{*}{CrossEarth (Ours)}&\multirow{1}{*}{ViT-L \cite{dosovitskiy2020image}}&  &\textbf{31.8}&76.3&95.3&44.0&{79.0}&\textbf{65.3 (+4.3)} & &\textbf{60.7}&\underline{81.5}&\textbf{83.0}&\underline{44.6}&\textbf{70.0}&\textbf{67.9 (+0.3)} \\ \midrule

\multirow{1}{*}{DeepLabv2 \cite{chen2017deeplab}}&\multirow{1}{*}{ResNet-101 \cite{he2016deep}}&  & 3.1&3.0&35.8&0.1&0.5&8.5 & & 2.4&1.8&43.9&0.1&6.1&10.9 \\
\multirow{1}{*}{DAFormer \cite{hoyer2022daformer}}&\multirow{1}{*}{MiT-B5 \cite{xie2021segformer}}&  & 43.0&63.4&41.6&39.6&10.3&39.6 & & 65.9&66.7&77.1&27.3&\underline{76.4}&62.7 \\
\multirow{1}{*}{HRDA \cite{hoyer2022hrda}}&\multirow{1}{*}{MiT-B5 \cite{xie2021segformer}}&  & \underline{43.4}&62.9&\textbf{56.8}&38.9&\underline{13.1}&\textbf{43.0} & & 65.7&67.2&78.0&26.5&\textbf{78.3}&63.1 \\

\multirow{1}{*}{S12-MoCo \cite{stewart2023ssl4eo}}&\multirow{1}{*}{ViT-S \cite{dosovitskiy2020image} }&  &  37.4 & 22.5 & 33.4 & 0.0 & 0.0 & 18.7 & & 49.9 & 21.1 & 60.0 & 0.1 & 0.1 & 26.3  \\

\multirow{1}{*}{S12-DINO \cite{stewart2023ssl4eo}}&\multirow{1}{*}{ViT-S \cite{dosovitskiy2020image} }&  &  32.3 & 14.8 & 39.2 & 0.0 & 0.0 & 17.3 & & 36.4 & 12.3 & 59.0 & 0.0 & 0.0 & 21.5 \\

\multirow{1}{*}{S12-MAE \cite{stewart2023ssl4eo}}&\multirow{1}{*}{ViT-S \cite{dosovitskiy2020image} }&  & 29.4 & 10.5 & 32.9 & 0.0 & 0.0 & 14.5 & & 31.9 & 10.7 & 55.0 & 0.0 & 0.1 & 19.5   \\

\multirow{1}{*}{DOFA \cite{xiong2024neural} }&\multirow{1}{*}{ViT-B \cite{dosovitskiy2020image} }&  & 37.3 & 17.8 &30.4&0.1&0.1 & 17.1 & & 49.9 & 21.4 & 56.9 &0.0 & 0.4 & 35.7 \\

\multirow{1}{*}{SatMAE \cite{cong2022satmae}}&\multirow{1}{*}{ViT-L \cite{dosovitskiy2020image} }&  Trf2Tem& 40.1&54.3&32.1&32.7&4.8 & 32.8 & Trf2Tms& 67.2&55.3&78.4&13.9&68.8 &56.7  \\

\multirow{1}{*}{ScaleMAE \cite{reed2023scale}}&\multirow{1}{*}{ViT-L \cite{dosovitskiy2020image} }&  & 36.5 & 53.5&14.7&36.2&13.7& 30.9& &65.3&63.8&76.5 &18.1&66.1 &58.0  \\

\multirow{1}{*}{RemoteCLIP \cite{liu2024remoteclip}}&\multirow{1}{*}{ViT-L \cite{dosovitskiy2020image} }&  & 34.6  & 17.3  & 27.9&  0.0& 0.0  &26.0 & &  43.2 & 14.7& 60.0& 0.0& 0.0& 23.6  \\
\multirow{1}{*}{MTP \cite{wang2024mtp}}&\multirow{1}{*}{ViT-L \cite{dosovitskiy2020image}}&  & 39.4&65.0&31.5&39.0&9.9&37.0 & &64.8&67.9&74.7&16.8&74.8&59.8 \\
\multirow{1}{*}{Rein (Baseline) \cite{Wei_2024_CVPR}}&\multirow{1}{*}{ViT-L \cite{dosovitskiy2020image}}&  & 37.8&\textbf{73.1}&20.3&\textbf{47.1}&{10.9}&37.8 & &\textbf{68.7}&\textbf{69.6}&\textbf{79.9}&\textbf{44.0}&62.4& \textbf{64.9} \\
\multirow{1}{*}{CrossEarth (Ours)}&\multirow{1}{*}{ViT-L \cite{dosovitskiy2020image}}&  &\textbf{43.8}&\underline{70.0}&\underline{41.1}&\underline{41.6}&  \textbf{15.1}&\underline{42.3 (+4.5)} & &\underline{68.1}&\underline{68.2}&\underline{79.0}&\underline{34.3}&{72.5}&\underline{64.4 (-0.5)} \\
\bottomrule
    \end{tabular}}
    
    \label{casid}
\end{table}

\twocolumn
\clearpage